%% file: main.tex
\documentclass[lettersize,journal]{IEEEtran}
\usepackage{amsmath,amsfonts}
\usepackage{array}
\usepackage[font=normalsize,labelfont=sf,textfont=sf]{subfig}
\usepackage{textcomp}
\usepackage{stfloats}
\usepackage{url}
\usepackage{verbatim}
\usepackage{graphicx}
\usepackage{cite}
\usepackage{bm}
\usepackage{adjustbox}
\usepackage{booktabs}
\usepackage{multirow}
\usepackage[table]{xcolor}
\usepackage{pifont}
\usepackage{makecell}
\usepackage{enumitem}
\usepackage{titletoc}

\def\BibTeX{{\rm B\kern-.05em{\sc i\kern-.025em b}\kern-.08em
    T\kern-.1667em\lower.7ex\hbox{E}\kern-.125emX}}

\newcommand{\mypara}[1]{\smallskip\noindent\textbf{#1}}
\newcommand{\revise}[1]{#1}

\newcommand{\rev}[1]{#1}

\makeatletter
\renewcommand{\thesection}{\arabic{section}}

\makeatother

\definecolor{cvprblue}{rgb}{0.21,0.49,0.74}
\usepackage[colorlinks=true, linkcolor=cvprblue, citecolor=cvprblue, urlcolor=cvprblue, filecolor=cvprblue]{hyperref}
\begin{document}

\title{Generalized Decoupled Learning for Enhancing Open-Vocabulary Dense Perception}
\author{
    Junjie Wang,
    Keyu Chen,
    Yulin Li,
    Bin Chen,
    Hengshuang Zhao,
    Xiaojuan Qi,
    Zhuotao Tian\textsuperscript{*}
}

\twocolumn[{%
\renewcommand\twocolumn[1][]{#1}%
\maketitle
\vspace{-8mm}
\centering
\includegraphics[width=.9\textwidth]{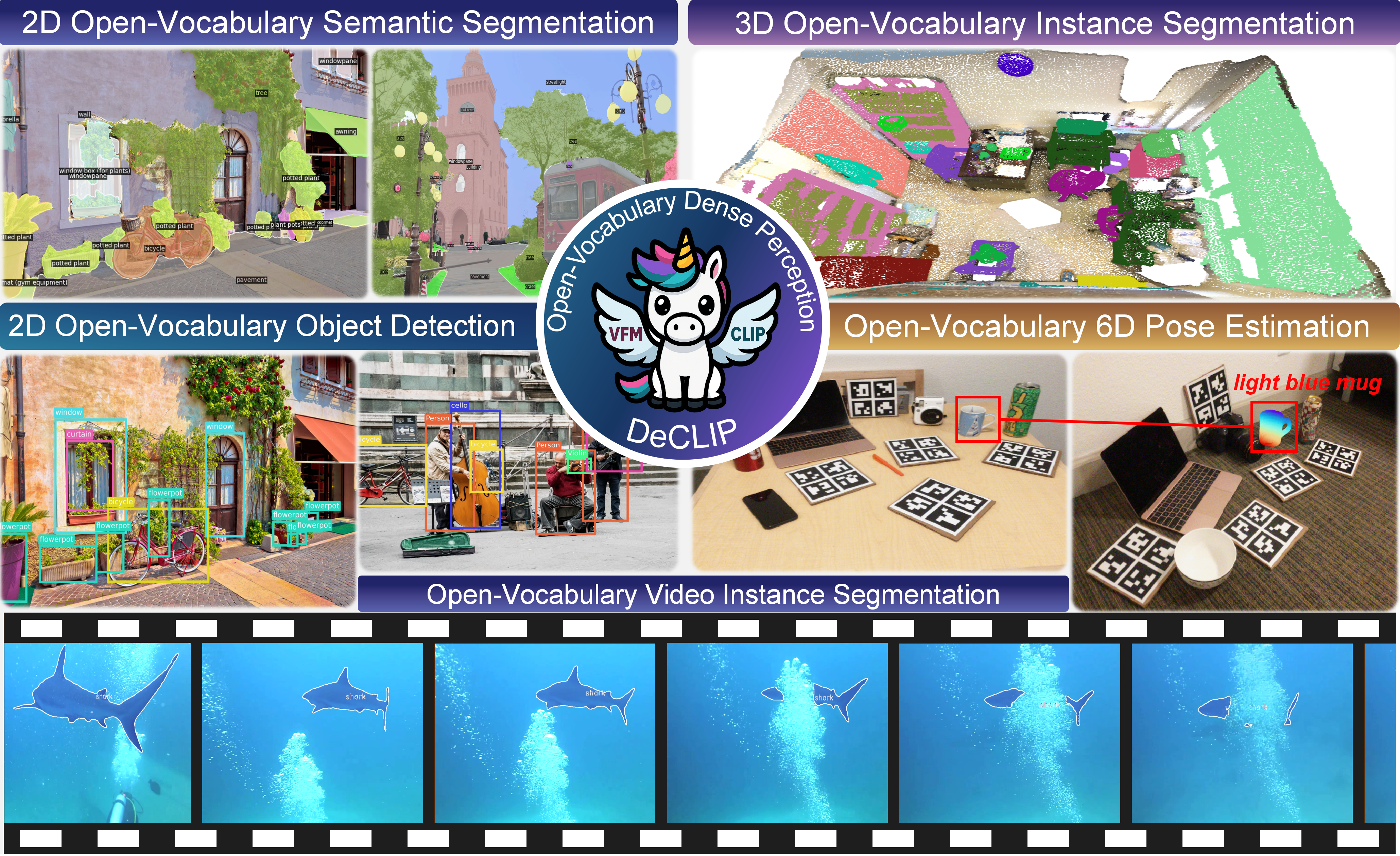}
\vspace{-2mm}
\captionof{figure}{\textbf{Overview of the proposed method.} DeCLIP adopts a decoupled learning strategy to enhance pixel-level open-vocabulary representations, which can serve as a general foundation model for OV dense perception tasks such as 2D detection and segmentation, 3D instance segmentation, video instance segmentation, and 6D object pose estimation.}
\label{teaser_1}
\vspace{0.4cm}
}]      

% 作者详细信息放此处，teaser不兼容\thanks
\begingroup
\renewcommand{\thefootnote}{}%
\footnotetext[1]{
Junjie Wang, Keyu Chen, Yulin Li, Bin Chen, and Zhuotao Tian are with the Harbin Institute of Technology (Shenzhen).\\
Zhuotao Tian is also with the Shenzhen Loop Area Institute.\\
Hengshuang Zhao and Xiaojuan Qi are with the University of Hong Kong.\\
\textsuperscript{*}Corresponding Author: Zhuotao Tian.\\
}
\endgroup
% 结束作者详细信息

\input{sec/0_abstract}
\input{sec/1_intro}
\input{sec/2_related_works}

\input{sec/3_background_motivation}
\input{sec/4_method}
\input{sec/5_experiments}
\input{sec/6_conclusion}

% Include supplement-only references in the shared bibliography.
\nocite{fasterrcnn,unet,mask2former,wu2023aligning,wu2023cora,catseg,clip,clim,wu2023clipself,regionclip,ovdquo,vild,fvlm,OVSeg,stablediffusion,ovdsurvey,groupvit,sasdet,detic,ovdetr,sam3d,OVIR,oryon,open3dis,openmask3d,clip_vis,OVFormer,MAFT,pacl,samclip,ovsam,frozenseg,sam,repa,convnext,ViT,transformer,dino,dinov2,sam2,grad,sclip,CLIPtrase,clearclip,clipdino,DiffSegmenter,DPL,cliper,silc,ZeroSeg,maskrcnn,rkd,isbnet,mask3d,openscene,dbscan,OpenIns3D,mscoco,adamw,evaclip,register,scannet200,MaskTrack_rcnn,IDOL,SORT,OWTB,XMem,LV-VIS,lvis,burst,OVIS,ShapeNet6D,ShapeNetSem,REAL275,Toyota,bop,swin,detr,kim2023region,CFM,ZegFormer,ZSseg,SAN,ODISE,fcclip,maskclip,reco,tcl,naclip,resclip,proxyclip,cocostuff,ade20k,voc,context,cityscape,scd,coco_panoptic,am_radio,ijepa,sd+dino,cc3m,tinyclip,robust,segearth,openearthmap,loveda,isaid,uavid,udd5,vdd,gem}
\bibliographystyle{IEEEtran}
\bibliography{main}

\input{sec/7_appendix}

\end{document}

%% file: sec/0_abstract.tex
\begin{abstract}
Dense visual perception tasks have been constrained by their reliance on predefined categories, limiting their applicability in real-world scenarios where visual concepts are unbounded. While Vision-Language Models (VLMs) like CLIP have shown promise in open-vocabulary tasks, their direct application to dense perception often leads to suboptimal performance due to limitations in local feature representation. In this work, we present our observation that CLIP's image tokens struggle to effectively aggregate information from spatially or semantically related regions, resulting in features that lack local discriminability and spatial consistency. To address this issue, we propose DeCLIP, a novel framework that enhances CLIP by decoupling the self-attention module to obtain ``content'' and ``context'' features respectively. The context features are enhanced by jointly distilling semantic correlations from Vision Foundation Models (VFMs) and object integrity cues from diffusion models, thereby enhancing spatial consistency. In parallel, the content features are aligned with image crop representations and constrained by region correlations from VFMs to improve local discriminability. Extensive experiments demonstrate that DeCLIP establishes a solid foundation for open-vocabulary dense perception, consistently achieving state-of-the-art performance across a broad spectrum of tasks, including 2D detection and segmentation, 3D instance segmentation, video instance segmentation, and 6D object pose estimation. Code is available at https://github.com/xiaomoguhz/DeCLIP
\end{abstract}

%% file: sec/1_intro.tex
\section{Introduction}
\IEEEPARstart{I}{n} the era of deep learning, dense perception tasks like object detection \cite{fasterrcnn} and image segmentation \cite{unet,mask2former} have rapidly advanced and are widely used. However, traditional methods \cite{fasterrcnn,mask2former} recognize only a fixed set of predefined categories. This restriction hinders the practical application of these methods in real-world settings, where the range of visual concepts is virtually boundless. Consequently, increasing attention has been drawn to OV (Open-Vocabulary) methods \cite{wu2023aligning,wu2023cora,catseg}, which aim to detect and segment objects from any category using textual descriptions.

\begin{figure*}[htbp]
\centering
\includegraphics[width=0.9\linewidth]{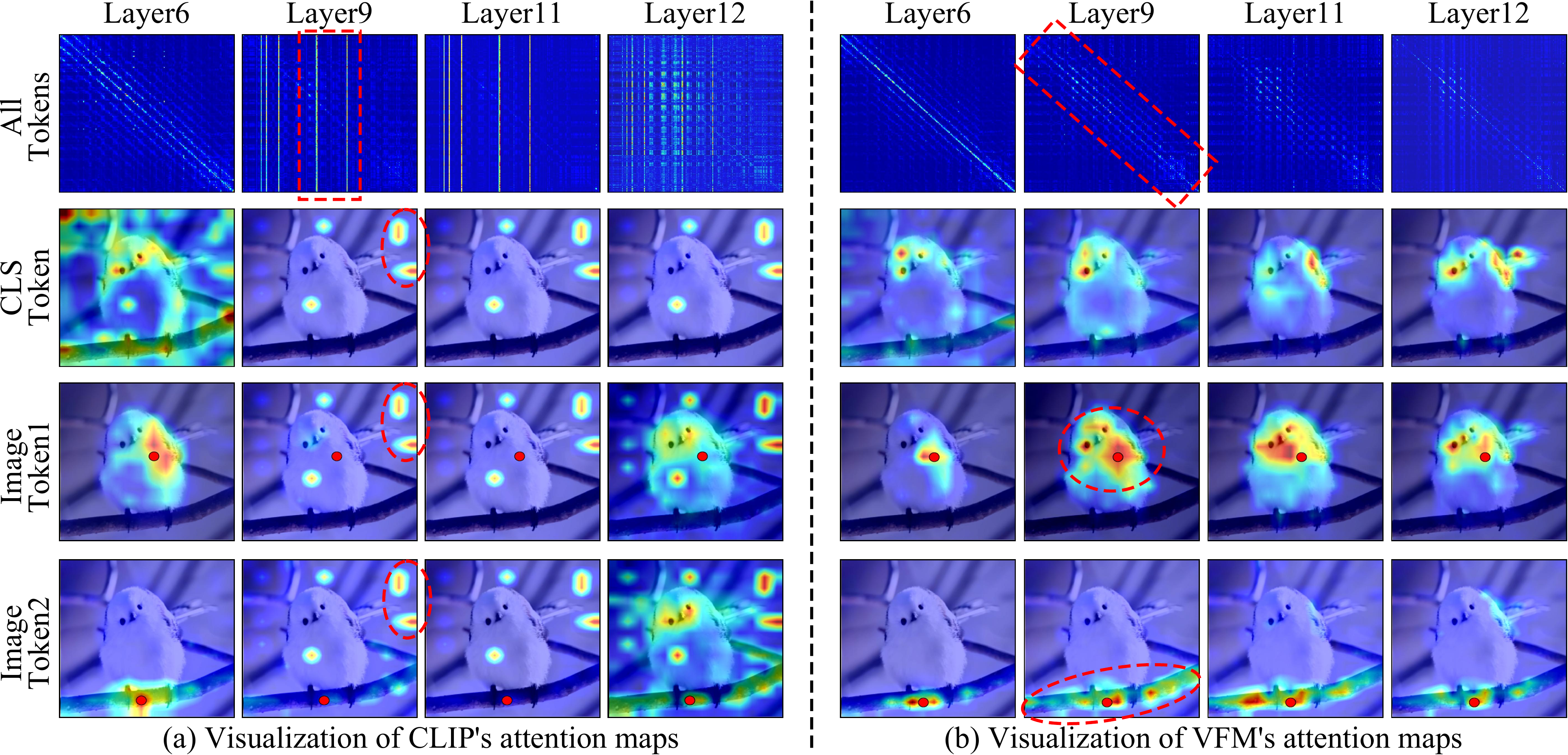}
\caption{\textbf{Comparison of the attention patterns of CLIP and VFM across different layers.} \revise{The attention weights of CLIP and VFM are collected at different layers, averaged over heads, and upsampled to the original image resolution. The first row visualizes attention weights for all tokens, while the remaining rows use the [CLS] token and selected image tokens as queries. Red dots indicate selected query image tokens. The figure highlights CLIP's deep-layer proxy-token behavior and VFM's consistent dense correlations, which are discussed in Sec.~\ref{key_observations}.}}
\label{motivation}
\end{figure*}

% \par Building on the success of Vision-Language Models (VLMs) \cite{clip,evaclip,openclip,flip} pre-trained on image-text pairs, such as CLIP \cite{clip}, researchers have started leveraging these models for OV dense perception tasks. Among these \cite{wu2023aligning,clim,wu2023clipself,sasdet,ovdquo,rtgen,declip}, transfer-learning approaches \cite{frozenseg,ovdquo,fvlm,fcclip,wu2023clipself,MAFT} have shown outstanding performance. These methods utilize the image encoder of VLM as a feature extractor and exclusively train lightweight task-specific components. Whereas using VLMs as feature extractors offers significant advantages due to their comprehensive pre-training, directly applying these image-level models to dense perception often leads to domain shift issues \cite{wu2023cora,wu2023clipself}.

\par Building on the success of Vision-Language Models (VLMs)~\cite{clip} pre-trained on image-text pairs, such as CLIP~\cite{clip}, researchers have extended these models to OV dense perception tasks~\cite{clim,wu2023clipself}. To apply CLIP to such tasks, various method paradigms have been explored, including pseudo-labeling~\cite{regionclip,ovdquo}, knowledge distillation~\cite{vild}, and transfer learning~\cite{fvlm,OVSeg}. Leveraging CLIP as the foundation provides significant advantages due to its comprehensive pre-training. However, directly applying these image-level models to dense prediction tasks often leads to domain shift issues, which degrades performance~\cite{wu2023cora}.

\mypara{What Hinders CLIP in Dense Perception? }
To assess CLIP's constraints in dense perception, we first analyze the discrepancies in attention maps across multiple layers between CLIP and VFM (Vision Foundation Model). The latter is known to excel in dense perception tasks. As shown in Fig.~\ref{motivation}, our experiments reveal that CLIP's [CLS] token may interfere with the correlations among other image tokens, leading to suboptimal performance in dense perception tasks. 

\mypara{Emerging Differences in Deep-Layer Attention.} We observe significant differences in attention maps between CLIP and VFMs begin to emerge from the 9th layer onward. Specifically, in deeper layers, CLIP's [CLS] token shifts focus away from primary objects within the image and attends highly to certain specific tokens, as highlighted by the red dashed circle in the 2nd row of Fig.~\ref{motivation}(a). Moreover, CLIP's image tokens (rows 3 and 4, Fig.~\ref{motivation}(a)) exhibit similar behavior to the [CLS] token, showing high attention to several specific tokens rather than semantically related regions, regardless of their positions. In contrast, VFM's image tokens consistently focus on semantically relevant regions from shallow to deep layers.
\par This observation sheds light on why CLIP struggles in dense perception tasks: its image tokens fail to aggregate information from semantically related regions, resulting in dense features that lack local discriminability and spatial consistency\footnote{Local discriminability refers to a model's ability to distinguish semantics between objects, while spatial consistency means semantics within an object remain coherent, similar, and clearly bounded.}. As shown in Fig.~\ref{teaser2}(b), directly using CLIP features on the COCO dataset yields inferior performance in object detection and semantic segmentation. An intuitive approach is to enhance CLIP's local representations through fine-tuning. However, balancing the optimizations of both dense feature spatial correlations and vision-language semantic alignment within a unified architecture becomes a new challenge. Therefore, \textit{can we disentangle CLIP's features and apply separate constraints to obtain diverse features in a unified architecture?}

\par \mypara{Our Solution. }
To address these challenges, we propose DeCLIP, a general unsupervised fine-tuning method aimed at enhancing both the discriminability and spatial consistency of CLIP’s local features. The core idea is to decouple CLIP’s self-attention and learn from different teachers separately.

Specifically, DeCLIP decouples the features in the self-attention module into ``\textit{content}'' and ``\textit{context}'' components. The ``content'' features, responsible for local discriminability, are fine-tuned by aligning pooled region features with their corresponding image crop [CLS] representations. Meanwhile, the “context” features, responsible for spatial consistency, are learned from the feature correlations generated by VFMs. This decoupled distillation design effectively mitigates optimization conflicts, improving the generalization ability when applying CLIP to downstream OV dense perception tasks. 
\par 
In addition, we observe that the semantic affinity maps of VFM (i.e., the teacher signals for context features) lack object integrity, such as unclear object boundaries and internal holes. To address this, we leverage self-attention maps from the Stable Diffusion (SD)~\cite{stablediffusion} to enhance semantic completeness. Additionally, we find that aligning content features with [CLS] representations weakens CLIP's dense correlations, prompting us to use regional correlations from VFM as a constraint. These enhancements further improve the fine-grained perceptual capability. As shown in Fig.~\ref{teaser2}, DeCLIP significantly outperforms CLIP in dense perception tasks. 

To summarize, our contributions are as follows:

\begin{itemize}
\item Through a comparative analysis of the attention maps of CLIP and VFM, we identified that CLIP's inferior performance in dense perception tasks stems from its image tokens failing to attend to semantically relevant regions in the deep layers.
\item To address this issue, we propose DeCLIP, an effective unsupervised fine-tuning framework, to enhance the discriminability and spatial consistency of CLIP's dense features via a decoupled feature enhancement strategy.
\item As illustrated in Fig.~\ref{teaser2}(a), extensive experiments demonstrate that DeCLIP can be applied to various tasks, including 2D detection/segmentation, 3D instance segmentation, and video instance segmentation. These results highlight its potential to serve as a foundation model for OV dense perception tasks.
\end{itemize}

%% file: sec/2_related_works.tex
\begin{figure*}[tbp]
\centering
\includegraphics[width=0.9\linewidth]{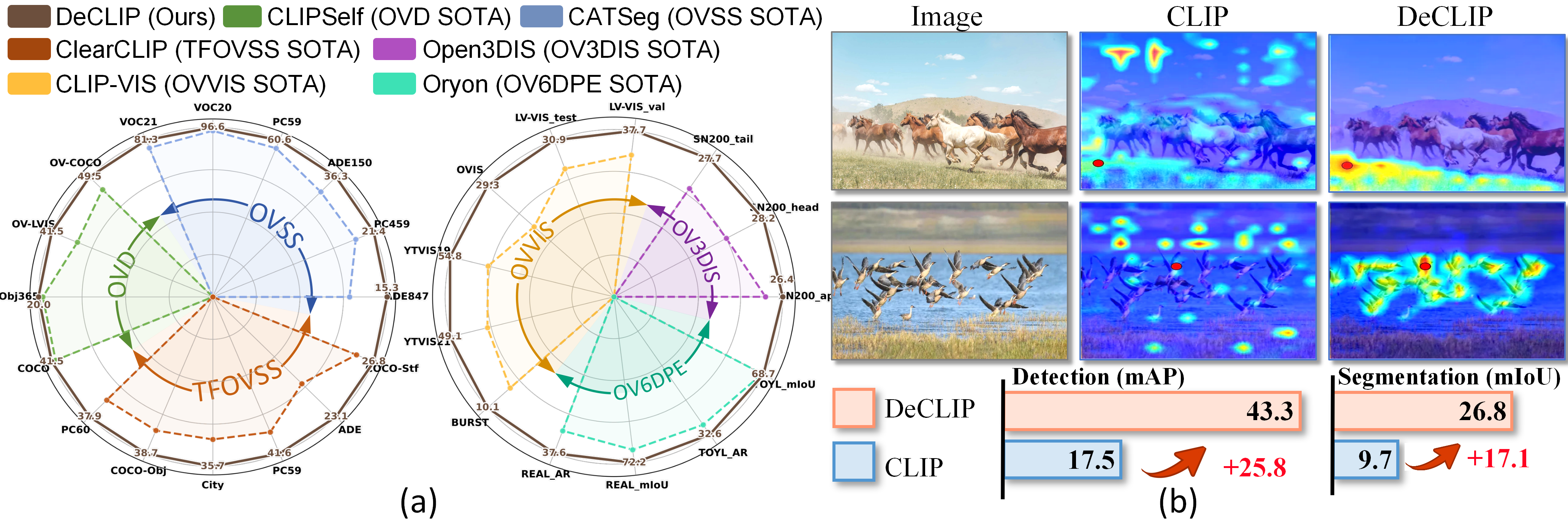}
\caption{\textbf{(a) Comparison between DeCLIP and state-of-the-art methods across six OV dense perception tasks.} These tasks include OV detection and semantic segmentation (OVD, OVSS), training-free OV semantic segmentation (TFOVSS), OV 3D and video instance segmentation (OV3DIS, OVVIS), and OV 6D pose estimation (OV6DPE). Experiments demonstrate the effectiveness of DeCLIP as a foundation model for OV dense perception. \textbf{(b) Quantitative and qualitative comparisons between DeCLIP and CLIP.} Compared to CLIP, DeCLIP's image tokens attend to semantically relevant regions of query tokens, significantly improving dense perception performance. The query tokens are highlighted by red dots. }
\label{teaser2}
\end{figure*}

\begin{figure*}[th]
  \centering
  \includegraphics[width=0.9\linewidth]{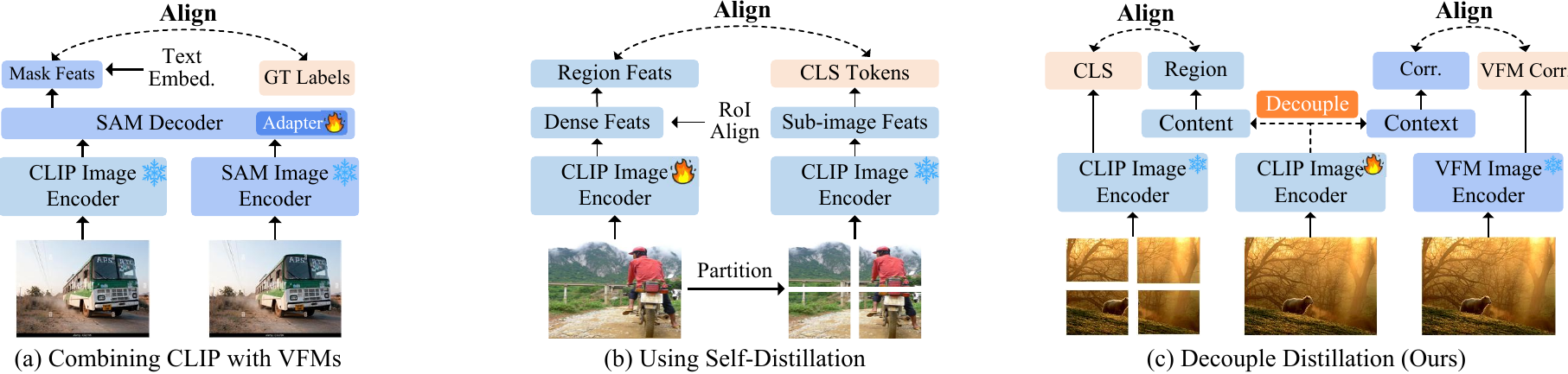}
  \caption{\textbf{Comparison between DeCLIP and existing methods.} \textbf{(a)} Methods that combine CLIP with VFMs typically integrate CLIP into SAM and transform it into an OV segmenter, which often requires costly dense annotations to train adapters. \textbf{(b)} Pre-fine-tuning methods adapt CLIP by aligning CLIP's local features with the CLS tokens of corresponding sub-images. However, these methods still fail to apply to image segmentation tasks. \textbf{(c)} Our DeCLIP enhances dense features' local discriminability and spatial consistency via the content-context decoupled learning, enabling a unified OV dense perception foundation model.}
\label{related_work_fig}
\end{figure*}
\section{Related Works}
\subsection{Open-Vocabulary Dense Perception}
\label{RW_1}
OV dense perception aims to achieve fine-grained classification of arbitrary visual instances by leveraging textual descriptions~\cite{ovdsurvey}. According to the structure and dimensionality of the input data, tasks can be categorized into three scenarios:

\mypara{2D Dense Perception. }This scenario takes RGB images as input, with typical tasks such as object detection~\cite{wu2023aligning} and image segmentation~\cite{groupvit,fvlm}. 2D OV dense perception methods based on pretrained CLIP can generally be categorized into pseudo-labeling~\cite{regionclip,sasdet,detic}, knowledge distillation~\cite{ovdetr}, and transfer learning~\cite{fvlm,ovdquo,wu2023clipself}. Regardless of the approach, these methods heavily rely on CLIP's dense perception capabilities. For instance, \cite{regionclip} produces region-text pseudo-labels using CLIP's regional features. \cite{wu2023clipself,fvlm} employ CLIP's image encoder as the backbone and train task-specific detection or segmentation components upon it. Clearly, these methods are limited by the quality of CLIP's dense features.

\mypara{3D Dense Perception. }Beyond 2D images, OV fine-grained perception can be extended to 3D scenes, enabling instance segmentation~\cite{sam3d,OVIR} or object pose estimation~\cite{oryon} on point cloud and RGB-D data. The challenges faced in 3D scenes are similar to those in 2D. Since CLIP itself lacks the ability to process 3D data, existing methods typically project 3D data into 2D RGB images from different viewpoints for OV instance recognition~\cite{open3dis,openmask3d}, then fuse the results from multiple viewpoints to achieve 3D instance segmentation.

\mypara{Video Dense Perception. }In addition to static 2D and 3D scenarios, OV dense perception can also be applied to video sequences, facilitating video instance segmentation~\cite{clip_vis,OVFormer}. In video scenarios, CLIP is often used to extract features for single-frame images, which are then pooled with masks to construct memory banks for multi-frame consistency tracking~\cite{clip_vis}. The quality of CLIP’s dense features directly affects the accuracy of single-frame instance segmentation and the effectiveness of the memory bank.

\par Therefore, whether for 2D, 3D, or video scenarios, CLIP's dense perception capability is critical. To address this, this paper proposes a decoupled feature enhancement strategy to improve the spatial consistency and discriminability of CLIP's dense features, and comprehensively evaluates the effectiveness of the proposed method across the aforementioned tasks.

\subsection{Adapting CLIP to Dense Perception Tasks.}
\label{RW_2}
Since VLMs~\cite{clip} are initially trained on image-text pairs, directly applying these image-level models to dense perception tasks, which require region-level or pixel-level semantic understanding, often results in significant performance degradation. Several studies have attempted to mitigate the domain shift problem encountered when applying CLIP to such tasks. These approaches can be broadly categorized into three main groups:

\mypara{Joint Fine-Tuning.} These methods fine-tune CLIP while training task-specific components~\cite{MAFT,catseg,OVSeg}. For instance, OV semantic segmentation method CAT-Seg~\cite{catseg} proposes an attention fine-tuning strategy based on ViT CLIP, which generalizes well to unseen categories. MAFT~\cite{MAFT} proposes a mask-aware fine-tuning strategy, which enhances CLIP's sensitivity to mask proposals while maintaining its transferability to unseen categories.

\mypara{Pre-Fine-Tuning.} These methods directly fine-tune CLIP using cost-efficient techniques~\cite{wu2023clipself,pacl,regionclip}. For instance, CLIM~\cite{clim} employs a mosaic augmentation technique to stitch multiple images into a single image, enabling each sub-image to serve as a pseudo-region for region-text contrastive learning. As illustrated in Fig.~\ref{related_work_fig}(b), CLIPSelf~\cite{wu2023clipself} enhances CLIP's region classification accuracy by maximizing cosine similarity between its region representations and the corresponding image crop representations.

\mypara{Combining CLIP with VFMs.} These studies~\cite{samclip,ovsam,frozenseg} investigate the integration of CLIP with VFMs. For instance, SAM-CLIP~\cite{samclip}, OV-SAM~\cite{ovsam}, and FrozenSeg~\cite{frozenseg} seek to combine the advanced image segmentation capabilities of SAM~\cite{sam} with the zero-shot semantic understanding capabilities of CLIP, as illustrated in Fig.~\ref{related_work_fig}(a).

\par Despite the promising results of the three categories of methods, they continue to exhibit certain limitations. Joint fine-tuning methods are typically specific to tasks or models and heavily rely on labor-intensive annotations of dense perception tasks. Pre-fine-tuning methods demonstrate broader applicability. However, their region-level fine-tuning technique remains limited in image segmentation tasks that require pixel-level details. Methods that combine CLIP with VFMs mainly focus on integrating CLIP into SAM or distilling them into a multi-task model, rather than enhancing CLIP itself as DeCLIP does. Notably, recent advances~\cite{repa} in image generation have also explored using VFM to enhance local representations. In contrast, DeCLIP introduces an innovative decoupled distillation strategy, as shown in Fig.~\ref{related_work_fig}(c).

%% file: sec/3_background_motivation.tex
\section{Background and Motivation}
\subsection{Preliminaries}
\label{preliminary}
\mypara{Contrastive Language-Image Pre-training} (CLIP) \cite{clip} is built upon two encoders, one for images and one for text. The visual encoder can be a CNN series \cite{convnext} or ViT \cite{ViT}, and the text encoder is a Transformer \cite{transformer}. This paper focuses on CLIP with the ViT architecture, which adopts the [CLS] token to represent the overall features of an image. CLIP learns vision-language alignment by maximizing the cosine similarity between the [CLS] token and text features of matched image-text pairs, and minimizing the similarity for unmatched pairs. 

\mypara{Dense Feature Extraction with CLIP.} ViT-based CLIP consists of a series of stacked attention blocks. For example, the ViT-B version of CLIP includes 12 attention block layers. Let $\mathbf{X}=\{\bm{x}_0, \bm{x}_1, \cdots, \bm{x}_{h \times w}\}$ denote the input to the last attention block, where $\bm{x}_i \in \mathbb{R}^{1 \times D}$. The computation within this attention block can be expressed as:
\begin{align}
\mathbf{Q} &= \text{Proj}_q(\mathbf{X}), \, \mathbf{K} = \text{Proj}_k(\mathbf{X}), \, \mathbf{V} = \text{Proj}_v(\mathbf{X}), \\
\mathbf{Y} &= \mathbf{X} + \text{Proj}\left(\text{Attn}_{qk} \cdot \mathbf{V}\right), \\
\mathbf{Z} &= \mathbf{Y} + \text{FFN}(\mathbf{Y}),
\end{align}
where $\mathbf{Q}$, $\mathbf{K}$, and $\mathbf{V}$ represent the query, key, and value embeddings, respectively; $\text{Proj}$ denotes projection layers; $\text{Attn}_{qk}=\text{SoftMax}\left(\mathbf{Q}\mathbf{K}^\top / \sqrt{d}\right)$ represents the self-attention process, with $d$ denoting the dimension of each attention head. $\text{FFN}$ denotes a feed-forward network. For simplicity, normalization operations are omitted. 
After passing through the final attention block, $\mathbf{Z}[0]$ represents the global [CLS] token. The remaining image patch embeddings $\mathbf{Z}[1:h \times w]$ can be reshaped to obtain dense feature representations $\mathbf{X}_{\text{dense}} \in \mathbb{R}^{C \times H \times W}$\footnote{The final vision-language projection layer is omitted here for brevity.}.
\begin{figure}[t]
  \centering
  \includegraphics[width=\linewidth]{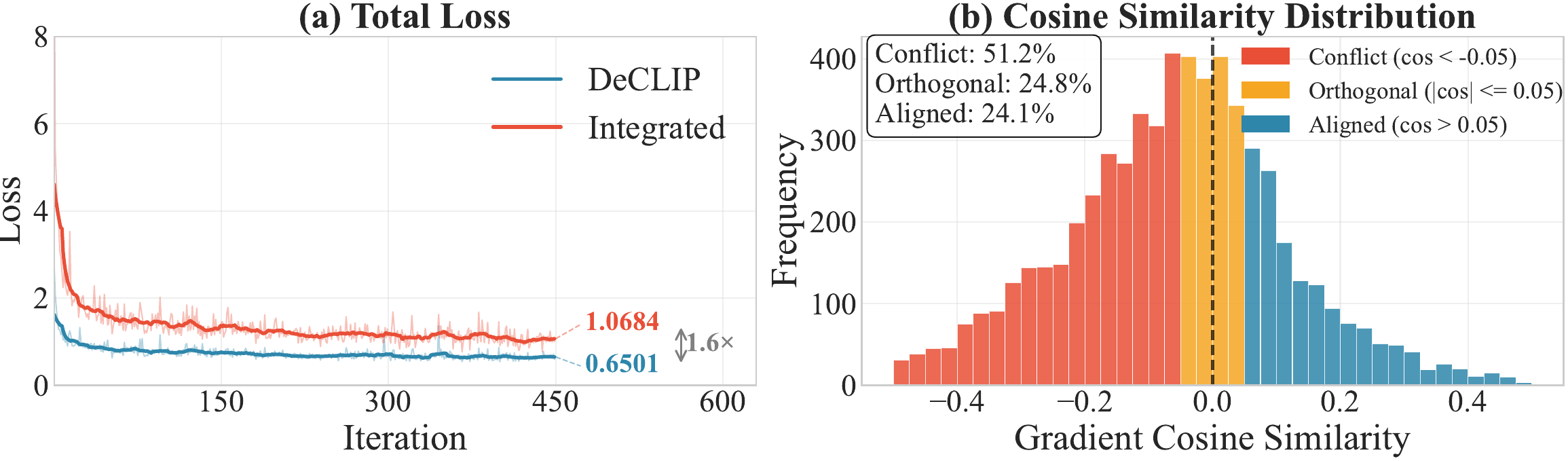}
  \caption{\textbf{Loss and gradient analysis.} \textbf{(a)} Training loss curves indicating superior stability and convergence for decoupled learning compared to the integrated approach. \textbf{(b)} Gradient cosine similarity between regional vision-language alignment and VFM distillation losses, which exhibit conflicting gradients for approximately 51.2\% of the training duration.}
\label{loss_grad}
\end{figure}

\subsection{Key Observations}
\label{key_observations}
To understand CLIP's limitations in dense perception, we first analyze the differences in attention maps across different layers of CLIP and VFM.

\mypara{Dense Correlations in CLIP's Early Layers.} As shown in the first column of Fig.~\ref{motivation}(a), we observe that in the shallow layers of CLIP (e.g., layer 6), the [CLS] token's attention is broadly distributed across the image. From the viewpoint of image tokens, the attention weights of each image token primarily focus on semantically relevant regions. The overall attention pattern at this stage exhibits locality, with the highlighted attention regions mainly distributed along the diagonal.

\mypara{Vanishing Correlations in CLIP's Deep Layers.} As illustrated in the second column of Fig.~\ref{motivation}(a), a notable shift in the attention patterns is observed in the deeper layers of CLIP. Specifically, in the deeper layers (e.g., layer 9), the [CLS] token shifts its focus away from the primary objects in the image and attends to several specific tokens, as indicated by the red dashed circles. For clarity, we refer to these special tokens as ``proxy tokens''. These specific tokens may serve as ``proxies'' for the [CLS] token. This suggests that these tokens aggregate essential information from other image tokens, enabling the [CLS] token to form an approximate ``global view'' by summarizing their content, thereby facilitating image classification. However, these ``proxy tokens'' may negatively affect the attention patterns of CLIP's image tokens.

\par In particular, when the [CLS] token focuses on the proxy tokens, CLIP's image tokens similarly shift their attention to these special tokens instead of semantically or spatially relevant regions. Furthermore, when the visualization position of the query image token is altered (from the bird to the branch, fourth row of Fig.~\ref{motivation}(a)), we observe that the new query image token continues to assign high attention to the ``proxy tokens.'' Notably, we observe that all image tokens at this stage exhibit this global attention pattern, with the highlighted attention regions manifesting as several prominent vertical lines (first row, Fig.~\ref{motivation}(a)).
\revise{Appendix~\ref{supp:subsec:proxy_token_behavior} provides large-scale statistics and proxy-as-query analyses.}

 \begin{table}[tbp]
    \centering
    \caption{\textbf{Quantitative comparison between decoupled and integrated learning.} Integrated learning with VFM distillation improves segmentation but compromises region classification accuracy. In contrast, our decoupled learning strategy simultaneously enhances performance in both tasks.}
    \begin{adjustbox}{width=\linewidth,center,valign=t}
    \begin{tabular}{l|cccc}
        \toprule
        \multirow{2.5}{*}{Distillation Type} & \multicolumn{2}{c}{Region Classification (mAcc)} & \multicolumn{2}{c}{Semantic Segmentation (mIoU)} \\
        \cmidrule(lr){2-3} \cmidrule(lr){4-5}
        & COCO (Thing) & COCO (Stuff) & Context59 & Cityscapes \\
        \midrule
        Baseline \cite{wu2023clipself} & 69.5 & 44.6 & 29.4 & 25.6 \\
         Integrated Learning & 65.6 \textcolor[HTML]{ff0000}{(-3.9)} & 41.3 \textcolor[HTML]{ff0000}{(-3.3)} & 32.4 \textcolor[HTML]{159838}{(+3.0)} & 28.7 \textcolor[HTML]{159838}{(+3.1)} \\
        Decoupled Learning & 75.0 \textcolor[HTML]{159838}{(+5.5)} & 51.8 \textcolor[HTML]{159838}{(+7.2)} & 35.3 \textcolor[HTML]{159838}{(+5.9)} & 32.3 \textcolor[HTML]{159838}{(+6.7)} \\
        \bottomrule
    \end{tabular}
    \end{adjustbox}
\label{sanity}
\end{table}

\mypara{VFM Exhibits Consistent Dense Correlations. }Considering the limitations of CLIP in dense perception, we examine VFMs such as DINO~\cite{dino,dinov2} and SAM~\cite{sam,sam2}. As shown in Fig.~\ref{motivation}(b), VFM attention maps do not exhibit the ``proxy token'' phenomenon observed in CLIP. Their image tokens consistently attend to semantically relevant regions from shallow to deep layers. In contrast, CLIP loses such dense correlations in deeper layers. This motivates us to use VFM correlations to improve CLIP dense features.

However, directly integrating VFM distillation with regional V-L alignment is suboptimal. We adopt an existing region Vision-Language (V-L) alignment method~\cite{wu2023clipself} as the baseline (Fig.~\ref{related_work_fig}(b)) and add VFM distillation\footnote{VFM distillation refers to aligning the feature self-correlations between CLIP's $\mathbf{X}_{\text{dense}}$ and those of the VFM.}. As shown in Tab.~\ref{sanity}, this integrated strategy improves segmentation but degrades regional classification. Following~\cite{grad}, we compute the gradient cosine similarity between the two losses. Fig.~\ref{loss_grad}(b) shows that their gradients are frequently conflicting, which explains the unstable optimization and higher loss in Fig.~\ref{loss_grad}(a). These observations motivate our decoupled learning strategy (Fig.~\ref{related_work_fig}(c)), which separates the two objectives into different feature branches. Compared with integrated learning, DeCLIP improves optimization stability and enhances both regional classification and segmentation performance (Tab.~\ref{sanity}). \revise{\mbox{Extended} comparisons before and after decoupling are provided in Appendix~\ref{supp:sec:analysis}, including quantitative comparisons, individual loss curves, training stability statistics, gradient statistics, attention comparisons, and feature visualizations.}

\begin{figure*}[htbp]
\centering
\includegraphics[width=0.9\linewidth]{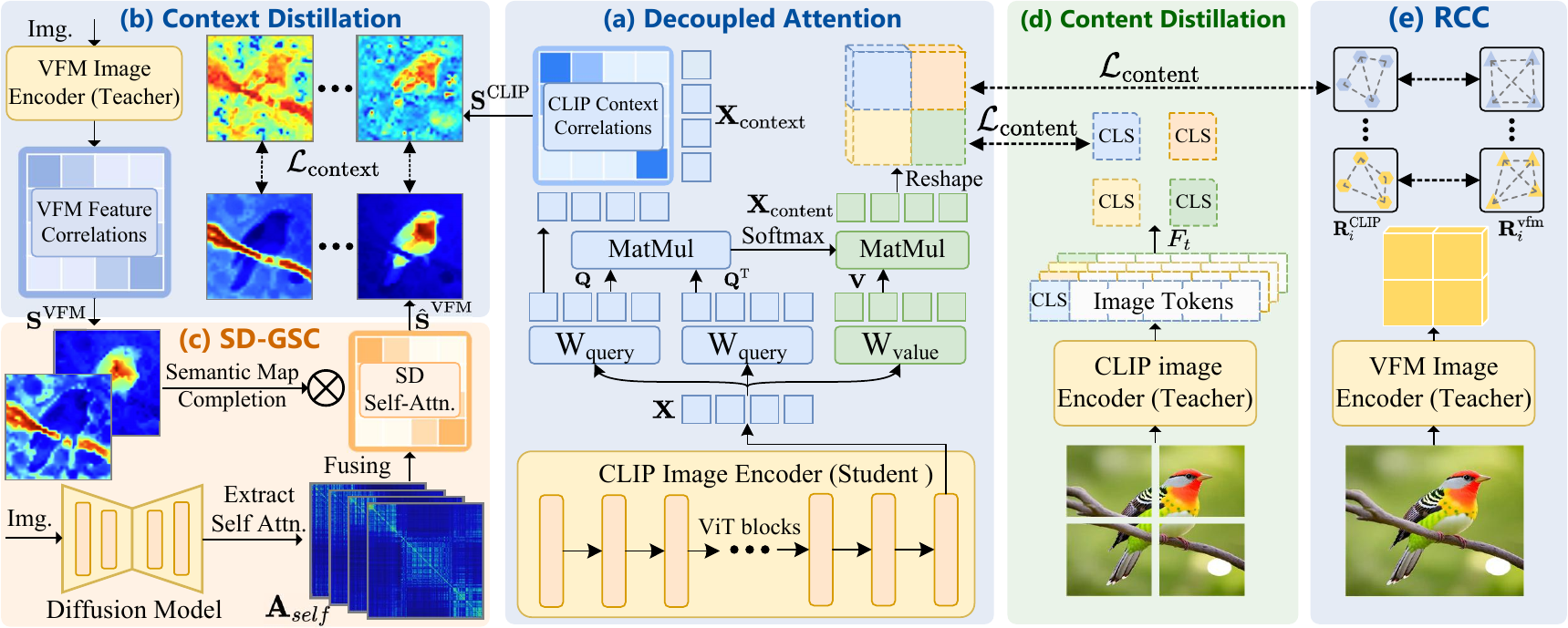}
\caption{\textbf{Illustration of the DeCLIP framework.} \textbf{(a) Decoupled Attention:} The final attention module of CLIP is decoupled into context features (to enhance spatial consistency) and content features (to enhance local vision-language alignment). \textbf{(b) Context Distillation:} Enhancing context features by leveraging semantic affinity maps from the VFM. \textbf{(c) SD-Guided Semantic Completion (SD-GSC):} The self-attention maps of SD are leveraged to enhance the semantic affinity map of VFMs and further optimize the distillation signals for the context features. \textbf{(d) Content Distillation:} Enhancing content features by aligning region representations with the corresponding [CLS] tokens. \textbf{(e) Region Correlation Constraint (RCC):} Using VFM's region correlations to prevent the collapse of CLIP's dense correlations during content distillation.}
\label{fig_method}
\end{figure*}

%% file: sec/4_method.tex
\section{Method}
\par In this section, we present DeCLIP, an unsupervised fine-tuning framework for adapting CLIP to dense perception tasks. We first explain how to decouple CLIP's self-attention mechanism into ``content'' and ``context'' components (Sec.~\ref{sec3.1}). Then, in Sec.~\ref{declip}, we illustrate how to enhance context features by distilling from VFM's semantic affinity maps, and how to further refine these affinity maps using the self-attention maps of SD. Subsequently, we elaborate on how to enhance the content features using the [CLS] representation from CLIP and the region correlations from VFM.

\subsection{Decoupled Attention}
\label{sec3.1}

\mypara{Rethinking the Self-Attention.} As described in Sec.~\ref{preliminary}, in CLIP's last attention block, the $\mathbf{V}$ features are weighted and summed under the guidance of the attention map ($\text{Attn}_{qk}$) derived from $\mathbf{Q}$ and $\mathbf{K}$, which define spatial or semantic correlations among image tokens. Studies \cite{sclip,CLIPtrase,clearclip,clipdino} have shown that CLIP's dense features $\mathbf{X}_{\text{dense}}$ can be directly used for semantic segmentation by per-pixel classification, indicating that each pixel of $\mathbf{X}_{\text{dense}}$ contains independent semantic information. Inspired by this, we regard $\mathbf{Q}$ and $\mathbf{K}$ as anchors for improving spatial consistency, and $\mathbf{X}_{\text{dense}}$ as an anchor for enhancing local discriminability.

\par Additionally, recent training-free OVSS studies \cite{sclip,clearclip} have further promoted us to decouple CLIP’s self-attention followed by distillation. They modify CLIP’s attention block from $\text{Attn}_{qk}$ to $\text{Attn}_{qq}$ and remove the residual connections, simplifying the optimization of local feature consistency by focusing on $\mathbf{Q}$ alone. Based on our rethinking of CLIP's self-attention and inspired by these methods, we propose decoupling CLIP's last attention block to obtain “content” and “context” features for distillation as follows:
\begin{align}
&\mathbf{X}_{\text{context}} = \text{Proj}_q(\mathbf{X}), \,  \mathbf{V} = \text{Proj}_v(\mathbf{X}), \\
&\mathbf{X}_{\text{content}} =\text{Proj}\left(\text{Attn}_\text{context} \cdot \mathbf{V}\right),
\\
&\text{Attn}_\text{{context}}=\text{SoftMax}\left(\mathbf{X}_{\text{context}}\mathbf{X}_{\text{context}}^\top / \sqrt{d}\right).
\label{eq5}
\end{align}
\par Specifically, $\mathbf{V}$ is aggregated based on the attention map ($\text{Attn}_\text{context}$) generated from $\mathbf{X}_{\text{context}}$. $\mathbf{X}_{\text{context}}$ determines which image tokens are semantically or spatially related. $\mathbf{X}_{\text{content}}$ carries the semantic information of each image token in the V-L space. By decoupling the features in this manner, we can apply different guidance constraints to $\mathbf{X}_{\text{context}}$ and $\mathbf{X}_{\text{content}}$ to obtain diverse feature representations in a unified architecture without interference. As observed in Sec.~\ref{key_observations}, VFM demonstrates a strong correlation among image tokens with the same semantics. Therefore, we leverage it as guidance for $\mathbf{X}_{\text{context}}$ to enhance the spatial consistency of CLIP's dense features. Meanwhile, we utilize CLIP's [CLS] representation for sub-images~\cite{wu2023clipself} as guidance for $\mathbf{X}_{\text{content}}$ to improve the local V-L alignment of CLIP's dense features. 
\par As shown in Tab.~\ref{sanity} (row 3), this decoupled optimization significantly enhances the local discriminability and spatial consistency of CLIP's features, resulting in simultaneous improvements in both region classification accuracy and semantic segmentation performance.

\begin{figure}[tbp]
\centering
\includegraphics[width=0.99\linewidth]{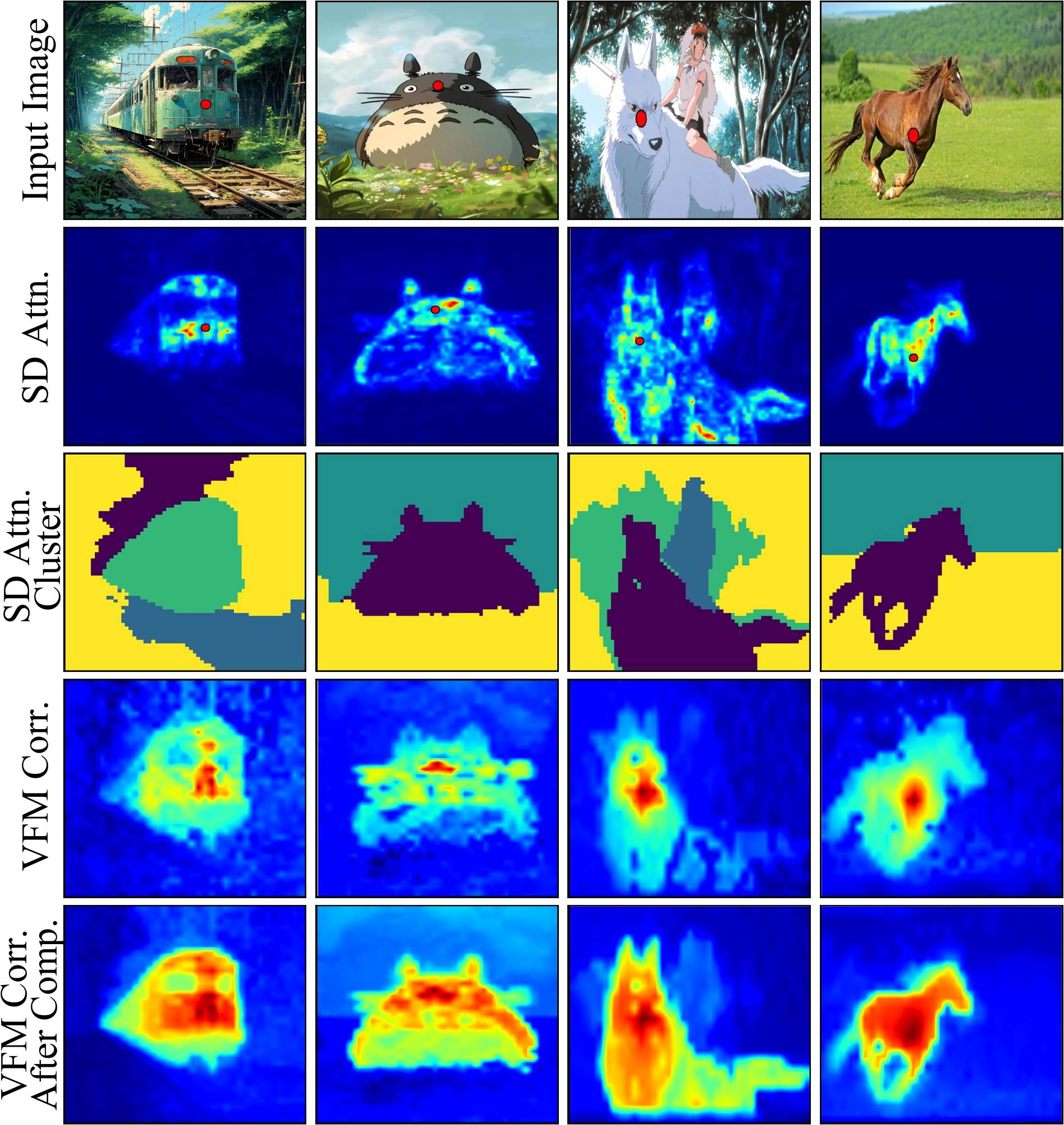}
\caption{\textbf{Visualization of VFM semantic affinity maps before and after semantic completion.} The cosine similarity and attention between the query token (red dot) and the remaining image tokens are visualized. SD attention maps are highly sensitive to high-frequency features, especially object contours (second row). K-means clustering of SD attention maps demonstrates notable object integrity (third row). Before completion (fourth row), the VFM semantic affinity map is blurred at boundaries and contains internal gaps. After completion (last row), the affinity map exhibits improved object integrity.}
\label{motivation2}
\end{figure}

\subsection{DeCLIP}
\label{declip}
In this section, we elaborate on how the decoupled features $\mathbf{X}_{\text{content}}$ and $\mathbf{X}_{\text{context}}$ learn from their respective teacher models to enhance CLIP's performance on OV dense perception tasks.

\par \mypara{Context Feature Distillation.} As discussed in Sec.~\ref{key_observations}, VFMs do not exhibit CLIP’s “proxy” token issue and better correlate semantically related image tokens, which may be conducive to the fine-grained local perception. Therefore, we distill these correlations into CLIP’s $\mathbf{X}_{\text{context}}$ features.

\par As illustrated in Fig.~\ref{fig_method}(b), given an input image $\mathbf{I}$, the VFM processes it and obtains feature representations $\mathbf{X}_{\text{dense}}^{\text{VFM}} \in \mathbb{R}^{\revise{D} \times HW}$. Here, $D$ represents the channel dimension of the VFM. Meanwhile, the student CLIP model takes the image $\mathbf{I}$ as input and outputs the content feature $\mathbf{X}_{\text{content}} \in \mathbb{R}^{C \times H \times W}$ and the context feature $\mathbf{X}_{\text{context}} \in \mathbb{R}^{C \times HW}$, as mentioned in Eq.\eqref{eq5}. Here, $C$ represents the dimension of the CLIP visual encoder. To ensure consistency in the number of image tokens after patch embedding, different input resolutions are typically used for the VFM and the student CLIP. 
\par To facilitate the transfer of semantic affinities among VFM features to CLIP's dense features, an intermediary is necessary to represent the correlation between pairs of image tokens. In our method, cosine similarity is employed as follows\footnote{For simplicity, we use $\cos(\cdot, \cdot)$ to denote the computation of cosine similarity between two matrices.}:
\begin{equation}
\mathbf{S}^{\text{VFM}} = \cos((\mathbf{X}_{\text{dense}}^{\text{VFM}})^{\mathsf{T}},\mathbf{X}_{\text{dense}}^{\text{VFM}})
\end{equation}
where $\mathbf{S}^{\text{VFM}} \in \mathbb{R}^{HW \times HW}$ denotes the similarity matrix of the entire VFM feature, where each entry $\mathbf{S}_{ij}^{\text{VFM}}$ represents the cosine similarity between patch tokens $\bm{x}_i$ and $\bm{x}_j$. Based on the analysis in Sec.~\ref{key_observations}, it is possible to improve CLIP dense features by leveraging the semantic affinity capability of VFM. A straightforward method involves aligning the correlations between $\mathbf{S}^{\text{VFM}}$ and $\mathbf{X}_{\text{context}}$ using either KL divergence or L2 loss. However, as shown in the fourth row of Fig.~\ref{motivation2}, VFM's affinity map $\mathbf{S}^{\text{VFM}}$ suffers from a lack of object integrity, exhibiting unclear boundaries and internal holes. 

\mypara{SD-Guided Semantic Completion (SD-GSC).} OV dense perception tasks require continuous and consistent predictions. However, the limited object integrity in VFM semantic affinity maps may lead to suboptimal distillation signals for context features. To address this limitation, we focus on another foundation model, the SD models~\cite{stablediffusion}, which are capable of generating high-quality images from textual descriptions.

\par We hypothesize that the self-attention modules~\cite{transformer,ViT} in the SD model~\cite{unet} effectively capture object boundaries and layout details~\cite{DiffSegmenter,DPL}, which are crucial for generating high-resolution and realistic images. As illustrated in Fig.~\ref{motivation2} (row 2), we visualize the 8$\times$ downsampled self-attention maps of SD. They clearly delineate the main object layout and contour details within the image. The K-means clustering results further demonstrate the effectiveness of SD self-attention in capturing object integrity (row 3, Fig.~\ref{motivation2}). However, these maps are less effective than VFM's semantic affinity maps in capturing semantic discrimination. As shown in Fig.~\ref{motivation2} (column 3), SD's attention incorrectly attends to the girl in the image (with the query token on the wolf), whereas the VFM's semantic affinity map does not. The inferior results of directly distilling SD's attention map (Tab.~\ref{ablation_vfm_sd}) further support this observation. On the other hand, this property complements VFM, whose semantic affinity maps are relatively accurate but lack object integrity. Therefore, we leverage SD's self-attention maps to further enhance the distillation signals from VFM for context features.

\par Specifically, the same image $\mathbf{I}$, along with an empty text prompt, is fed into the SD model to extract the self-attention maps from the U-Net. The self-attention maps are denoted as $\mathbf{A}_{self} \in \mathbb{R}^{L\times HW \times HW}$, where $L$ represents the total number of attention heads across all layers. Following~\cite{DiffSegmenter,cliper}, matrix chain multiplication is employed to fuse attention maps across different layers and heads, as shown below: 
\begin{equation}
\hat{\mathbf{A}}_{self} = \prod_{i=1}^{L} \mathbf{A}_{self}[i].
\end{equation}
Where $\hat{\mathbf{A}}_{self} \in \mathbb{R}^{HW \times HW}$ represents the fused attention map. Subsequently, the fused attention map $\hat{\mathbf{A}}_{self}$ is multiplied with the semantic affinity map $\mathbf{S}^{\text{VFM}}$ generated by VFM to incorporate object boundary information. The results are presented in the last row of Fig.~\ref{motivation2}. 
\begin{equation}
\hat{\mathbf{S}}^{\text{VFM}} = \hat{\mathbf{A}}_{self} \times \mathbf{S}^{\text{VFM}}.
\end{equation}
Subsequently, we use the KL divergence loss to align the similarity matrix $\mathbf{S}^{\text{CLIP}}\in \mathbb{R}^{HW \times HW}$ produced from $\mathbf{X}_{\text{context}}$ with the completed $\hat{\mathbf{S}}^{\text{VFM}} \in \mathbb{R}^{HW \times HW}$, as illustrated below:
\begin{equation}
\mathcal{L}_{\mathrm{context}} = \frac{1}{HW} \sum_{i=1}^{HW} \text{KL} \left( \hat{\mathbf{S}}_{i,:}^{\text{VFM}}, \mathbf{S}_{i,:}^{\text{CLIP}} \right),
\end{equation}
Notably, even with the introduction of the SD model, our DeCLIP fine-tuning framework still maintains efficacy and does not require any additional textual queries or annotations. 

\begin{figure}[tbp]
\centering
\includegraphics[width=0.99\linewidth]{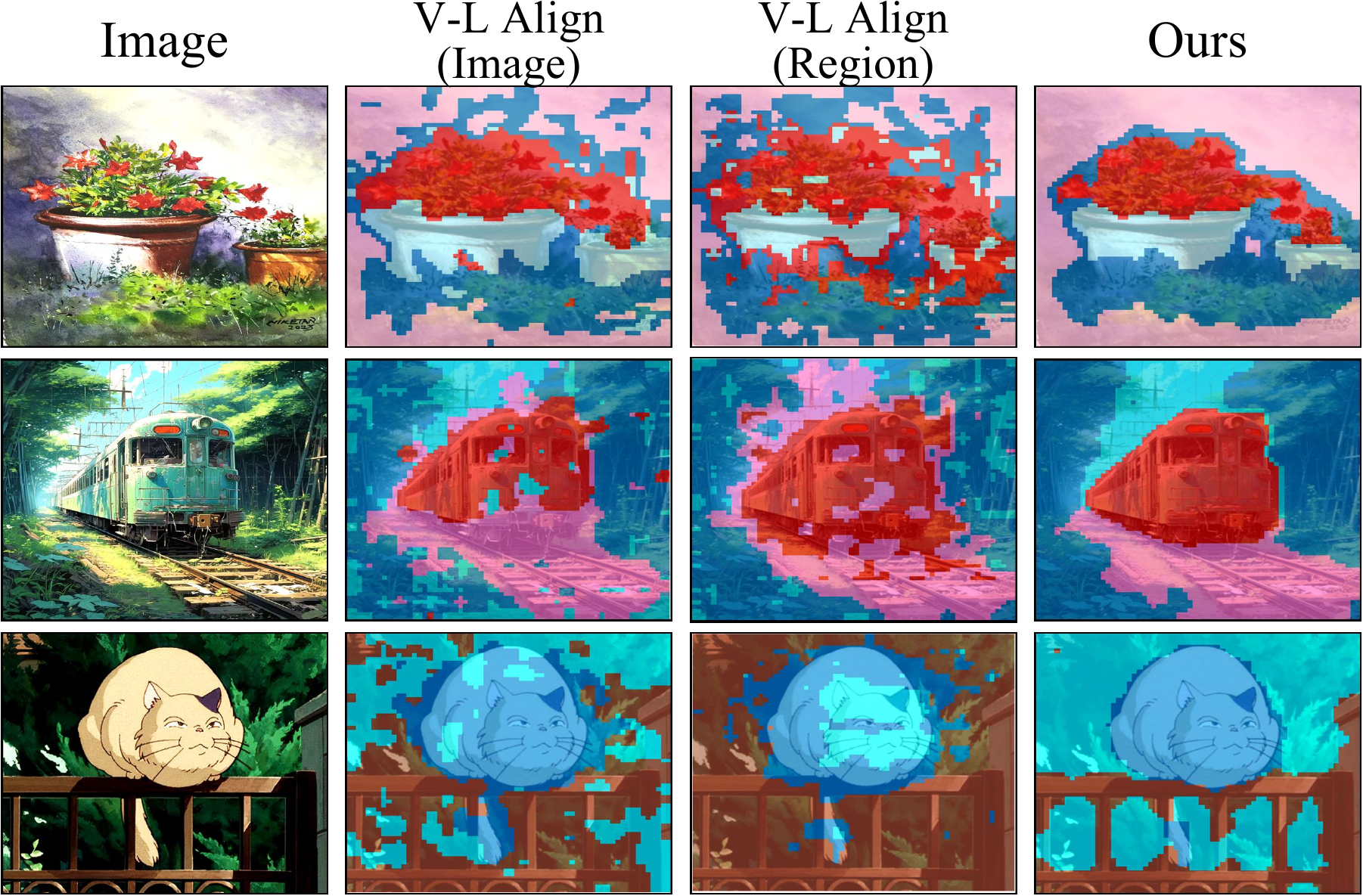}
\caption{\textbf{Comparison of feature semantic consistency.} ``V-L align (Image)'' refers to the vanilla CLIP, while ``V-L align (Region)'' denotes existing fine-tuning approaches~\cite{wu2023clipself,regionclip}. By performing K-means clustering on the cosine similarity maps of CLIP features, we observe that region-level fine-tuning may further weaken the pixel-level perception capability of CLIP.}
\label{motivation3}
\end{figure}

\par \mypara{Content Feature Distillation.} As shown in Fig.~\ref{fig_method}(d), the second teacher model in DeCLIP is itself, which is known as self-distillation \cite{silc,wu2023clipself,ZeroSeg,pacl}. We employ image patching to align the region representations of the student model’s feature map $\mathbf{X}_{\text{content}}$ with the corresponding image crop representations (i.e., [CLS] token) of the teacher model. 

\par Specifically, the input image $\mathbf{I}$ is initially partitioned into $k$ sub-regions. Subsequently, these sub-regions are cropped from the original image, resulting in a set of sub-images $S = \left\{\mathbf{I}_1^{\prime}, \mathbf{I}_2^{\prime}, \dots, \mathbf{I}_k^{\prime} \right\}$. Then, the student model applies RoI Align \cite{maskrcnn} to obtain pooled region features from $\mathbf{X}_{\text{content}}$ using the cropping coordinates of $S$, resulting in a region feature set $F_s = \left\{\bm{f}_1^s, \bm{f}_2^s, \dots, \bm{f}_k^s \right\}$, where $\bm{f}_i^s \in \mathbb{R}^{ N^2 \times C}$, and $N$ denotes the size of each RoI region.

\par Meanwhile, the teacher model takes the sub-image set $S$ as input and outputs a series of [CLS] tokens corresponding to the cropped sub-images, resulting in [CLS] token set $F_t = \left\{\bm{f}_1^t, \bm{f}_2^t, \dots, \bm{f}_k^t \right\}$, where $\bm{f}_i^t \in \mathbb{R}^{ C \times 1}$. For each region feature $\bm{f}_i^s$ generated by the student model, we compute a weighted summation according to its similarity with the teacher's [CLS] token $\bm{f}_i^t$ to obtain $\bar{\bm{f}}_i^s$, as defined below:
\begin{equation}
\bar{\bm{f}}_i^s = \text{softmax} \left( \cos(\bm{f}_i^s, \bm{f}_i^t) \right)^{\mathsf{T}} \cdot \bm{f}_i^s
\end{equation}
where $\bar{\bm{f}}_i^s \in \mathbb{R}^{1 \times C}$. We employ a cosine similarity loss to align the teacher's [CLS] representations from $F_t$ with the region features obtained by weighted summation from $F_s$, as follows:
\begin{equation}
\mathcal{L}_{\mathrm{content}}=\frac{1}{k} \sum_{i=1}^k1 - \cos\left(\bar{\bm{f}}_i^s, \bm{f}_i^t\right).
\end{equation}
The intuition of this distillation branch is that the [CLS] token in CLIP acquires strong representational capabilities through contrastive learning on large-scale image-text pairs, as described in Sec.~\ref{preliminary}. Therefore, aligning the region features in CLIP, \textit{i.e.}, $F_s$, with the corresponding [CLS] representations, \textit{i.e.}, $F_t$, enhances the local discriminative capability of its dense features.

\mypara{Region Correlation Constraint (RCC).} As shown in the third column of Fig.~\ref{motivation3}, we empirically find that directly aligning regional features and their corresponding [CLS] representations further weakens the correlations of CLIP's dense features (detailed analysis is provided in Appendix~\ref{supp:sec:rcc_vis}). Therefore, we utilize an additional constraint term during the content distillation process to prevent the degradation of CLIP's dense correlations by leveraging those of the VFM.
\par Specifically, we first restore the spatial dimensions of the VFM features to obtain $\mathbf{X}_{\text{dense}}^{\text{VFM}} \in \mathbb{R}^{\revise{D} \times H \times W}$. Then, we use RoI Align~\cite{maskrcnn} with the cropping coordinates to extract the regional representations of VFM, resulting in $F_{\text{vfm}} = \left\{\bm{f}_1^{\text{vfm}}, \bm{f}_2^{\text{vfm}}, \dots, \bm{f}_k^{\text{vfm}} \right\}$, where $\bm{f}_i^{\text{vfm}} \in \mathbb{R}^{N^2 \times D}$. Subsequently, we employ the KL divergence to align the internal correspondences of regional features between CLIP and VFM~\cite{rkd}. The new content loss is defined as follows:
\begin{equation}
\mathcal{L}_{\mathrm{content}} = \frac{1}{k} \sum_{i=1}^k  1 - \cos\left(\bar{\bm{f}}_i^s, \bm{f}_i^t\right) + \text{KL} \left( \mathbf{R}_i^{\text{vfm}}, \mathbf{R}_i^{\text{CLIP}} \right)
\end{equation}
\begin{equation}
\mathbf{R}_i^{\text{vfm}}=\cos(\bm{f}_i^{\text{vfm}},(\bm{f}_i^{\text{vfm}})^{\mathsf{T}}), \quad \mathbf{R}_i^{\text{CLIP}}=\cos(\bm{f}_i^{s},(\bm{f}_i^{s})^{\mathsf{T}})
\end{equation} 
Finally, the entire distillation learning process of DeCLIP is summarized as follows:
\begin{equation}
\mathcal{L}_{\mathrm{total}}=\mathcal{L}_{\mathrm{content}}+\lambda\mathcal{L}_{\mathrm{context}}.
\end{equation}

%% file: sec/5_experiments.tex
\begin{table*}[tbp]
\centering
\caption{\textbf{Quantitative results of OV 3D instance segmentation evaluated on the ScanNet200 dataset.} DeCLIP significantly enhances the performance of existing methods and achieves the highest results. $^{\dagger}$ indicates that training and evaluation are conducted on the same dataset, representing the reference results of conventional closed-set 3D instance segmentation.  }
\label{3D_seg}
\adjustbox{width=.92\textwidth}{
\begin{tabular}{lccccccc}
\toprule
Method & Setting & 3D Proposal & AP & AP$_{50}$ & AP$_{25}$ & AP$_{\mathrm{head}}$ & AP$_{\mathrm{tail}}$ \\ 
\midrule
 ISBNNet~\cite{isbnet}$^{\dagger}$ & Fully-sup & None & 24.5 & 32.7 & 37.6 & 38.6 & 12.5 \\ 
 Mask3D~\cite{mask3d}$^{\dagger}$ & Fully-sup & None & 26.9 & 36.2 & 41.4 & 39.8 & 17.9 \\ 
\midrule
OpenScene~\cite{openscene} + DBScan~\cite{dbscan} & Open-vocab & None & 2.8 & 7.8 & 18.6 & 2.7 & 2.6 \\ 
OpenScene~\cite{openscene} + Mask3D~\cite{mask3d} & Open-vocab & Mask3D~\cite{mask3d} & 11.7 & 15.2 & 17.8 & 13.4 & 9.9 \\ 
SAM3D~\cite{sam3d} & Open-vocab & None & 6.1 & 14.2 & 21.3 & 7.0 & 4.6 \\ 
OVIR-3D~\cite{OVIR} & Open-vocab & None & 13.0 & 24.9 & 32.3 & 14.4 & 11.7 \\ 
OpenIns3D~\cite{OpenIns3D} & Open-vocab & Mask3D~\cite{mask3d} & 8.8 & 10.3 & 14.4 & 16.0 & 4.2 \\ 
OpenMask3D~\cite{openmask3d} & Open-vocab & Mask3D~\cite{mask3d} & 15.4 & 19.9 & 23.1 & 17.1 & 14.9 \\ 
Open3DIS (2D)~\cite{open3dis} & Open-vocab & None & 18.2 & 26.1 & 31.4 & 18.9 & 19.2 \\ 
Open3DIS (3D)~\cite{open3dis} & Open-vocab & ISBNNet~\cite{isbnet} & 18.6 & 23.1 & 27.3 & 24.7 & 13.3 \\ 
Open3DIS (2D+3D)~\cite{open3dis} & Open-vocab & ISBNNet~\cite{isbnet} & 23.7 & 29.4 & 32.8 & 21.2 & 21.8 \\ 

\midrule
    \rowcolor[HTML]{d4effb} OpenMask3D + DeCLIP & Open-vocab & Mask3D~\cite{mask3d} &18.3 \textcolor[HTML]{2ECC71}{(+2.9)}  &23.4 \textcolor[HTML]{2ECC71}{(+3.5)} &26.8 \textcolor[HTML]{2ECC71}{(+3.7)} &18.4 \textcolor[HTML]{2ECC71}{(+1.3)} &18.9 \textcolor[HTML]{2ECC71}{(+4.0)} \\ 
    \rowcolor[HTML]{d4effb} Open3DIS (2D+3D) + DeCLIP & Open-vocab & ISBNNet~\cite{isbnet} &\textbf{26.4} \textcolor[HTML]{2ECC71}{(+2.7)}  &\textbf{32.8} \textcolor[HTML]{2ECC71}{(+3.4)} &\textbf{36.2} \textcolor[HTML]{2ECC71}{(+3.4)} &\textbf{28.2}  \textcolor[HTML]{2ECC71}{(+7.0)} &\textbf{27.7}  \textcolor[HTML]{2ECC71}{(+5.9)} \\ 
\bottomrule
\end{tabular}
}
\end{table*}

\section{Experiments}
This section evaluates DeCLIP across diverse OV dense perception benchmarks. We first outline the experimental setup and implementation details. Subsequently, we present performance improvements across various OV tasks, including detection, semantic segmentation (and its training-free variant), 3D instance segmentation, video instance segmentation, and 6D object pose estimation. For all baselines integrated with DeCLIP, we strictly adopt a ``drop-in replacement" strategy that replaces the image encoder while maintaining the original training and inference settings, including V-L prompt templates. All experiments also follow the standard OV paradigm without task-specific training on novel categories. For the training-free tasks, we utilize standard ImageNet prompts to ensure alignment with existing methods~\cite{clearclip}. Beyond standard benchmarks, we extend our evaluation to the Supplementary Material, providing detailed analyses on robustness against input corruptions, cross-domain generalization (Appendix~\ref{supp:sec:robustness}), and computational efficiency (Appendix~\ref{supp:sec:efficiency}).

\subsection{Experiment Details}
\mypara{Training Settings.} DeCLIP is trained on the training set of COCO2017~\cite{mscoco} using 8 GPUs, each with a batch size of 2, for 6 epochs (about 20 min/epoch). The AdamW~\cite{adamw} optimizer with a learning rate of $1\mathrm{e}{-5}$ and a weight decay of 0.1 is employed during the training process. Unless otherwise specified, we use EVA-CLIP~\cite{evaclip} as the VLM baseline in our experiments and fine-tune all of its encoder layers by default. We did not apply any augmentation to the input images because such operations are found to degrade performance.

\mypara{Hyperparameter Settings.} By default, we perform context feature distillation for DeCLIP's ViT-B and ViT-L using the DINOv2~\cite{dinov2} ViT-B and ViT-L models with registers~\cite{register}. We use input images with different resolutions for CLIP and VFM models to maintain a consistent number of tokens after patch embedding, since CLIP uses a patch size of 16 while DINOv2 uses 14. For example, we set the input resolution for CLIP to 560 and for DINOv2 to 490, ensuring both models possess 1225 image tokens. Following~\cite{cliper,DiffSegmenter}, we extract self-attention maps from the 45th time-step (out of 50 steps) of Stable Diffusion V2.1~\cite{stablediffusion} for the SD-guided semantic completion module. Following~\cite{wu2023clipself}, for the content feature distillation branch, we partition the image into $k$ blocks, where $k = m \times n$ and $m$ and $n$ are randomly sampled from the range [1, 6]. The weight $\lambda$ for context feature distillation is set to 0.25 by default. Distinct image normalization hyperparameters are set for the VFM, CLIP, and SD models, consistent with the settings applied during their respective pre-training.

\begin{figure}[tbp]
  \centering
  \includegraphics[width=.82\linewidth]{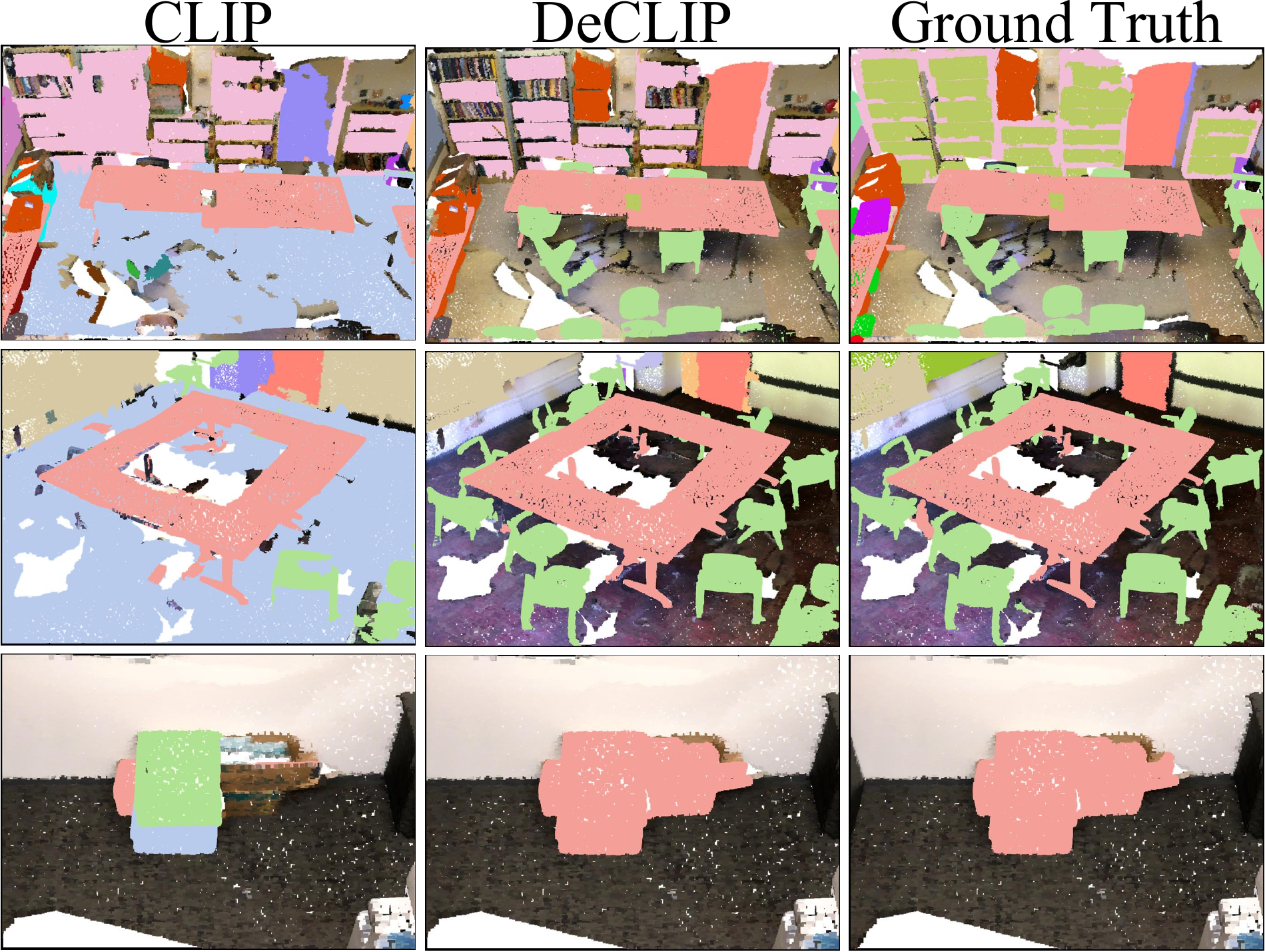}
  \caption{Qualitative comparison of OV 3D instance segmentation results between CLIP and DeCLIP on the ScanNet200~\cite{scannet200} dataset. The baseline method is Open3DIS~\cite{open3dis}.}
  \label{fig_3d}
\end{figure}

\begin{table*}[htbp]
\centering
\caption{\textbf{Quantitative results on the OV video instance segmentation task.} `/' indicates that training and evaluation are conducted on the same dataset, representing the reference results of conventional closed-set video instance segmentation. The combination of CLIP-VIS and DeCLIP achieves new state-of-the-art results on this task. }
\label{video_seg}
\adjustbox{width=.9\textwidth}{
\setlength{\tabcolsep}{4pt}
\begin{tabular}{l|cc|
                *{2}{c}
                *{2}{c}
                c
                c
                c
                *{2}{c}}
\toprule
\multirow{2.5}{*}{Method} & \multirow{2.5}{*}{Training Data} & \multirow{2.5}{*}{Backbone} &
\multicolumn{2}{c}{LV-VIS val} &
\multicolumn{2}{c}{LV-VIS test} &
OVIS & 
YTVIS19  &
YTVIS21 &
\multicolumn{2}{c}{BURST} \\ \cmidrule(lr){4-5}\cmidrule(lr){6-7}\cmidrule(lr){8-8}
\cmidrule(lr){9-9}\cmidrule(lr){10-10}\cmidrule(lr){11-12}
 &  &  & AP & AP$_\text{N}$ & AP & AP$_\text{N}$ & AP & AP & AP & AP & AP$_\text{N}$ \\
\midrule
\multicolumn{12}{c}{\em Classical video instance segmentation} \\
\midrule
MaskTrack R-CNN~\cite{MaskTrack_rcnn} & / & R50 & -- & -- & -- & -- & 10.8 & 30.3 & 28.6 & -- & -- \\
Mask2Former~\cite{mask2former}    & / & R50 & -- & -- & -- & -- & 17.3 & 46.4 & 40.6 & -- & -- \\
IDOL~\cite{IDOL}            & / & R50 & -- & -- & -- & -- & 30.2 & 49.5 & 43.9 & -- & -- \\
\midrule
\multicolumn{12}{c}{\em OV video instance segmentation} \\
\midrule
Detic\cite{detic}+SORT~\cite{SORT}   & LVIS & SwinB & 12.8 & 6.6 & 9.4 & 4.7 & 11.7 & 23.8 & 21.6 & 2.5 & 1.0 \\
Detic\cite{detic}+OWTB~\cite{OWTB}   & LVIS & SwinB & 14.5 & 11.8 & 13.6 & 5.5 & 30.0 & 9.7 & 11.4 & 3.9 & 2.4 \\
Detic\cite{detic}+XMem~\cite{XMem}   & LVIS & SwinB & 16.3 & 10.6 & 13.1 & 7.7 & -- & -- & -- & -- & -- \\
OV2Seg~\cite{LV-VIS}                & LVIS & SwinB & 21.1 & 16.3 & 16.4 & 11.5 & 17.5 & 37.6 & 33.9 & 4.9 & 3.0 \\
OVFormer~\cite{OVFormer}            & LVIS & SwinB+ViT-B & -- & -- & -- & -- & 21.3 & 44.3 & 37.6 & -- & -- \\
CLIP-VIS~\cite{clip_vis}   & LVIS & ConvNeXt-B & 32.2 & 40.2 & 25.3 & 30.6 & 18.5 & 42.1 & 37.9 & 8.3 & 12.7 \\
\midrule
\rowcolor[HTML]{d4effb} CLIP-VIS+DeCLIP   & LVIS & ViT-B/16 &34.8  &44.3  &28.4  &34.0  &22.2  &50.6  &43.3  &8.9  &15.7  \\
\rowcolor[HTML]{d4effb} CLIP-VIS+DeCLIP   & LVIS & ViT-L/14 & \textbf{37.7}  & \textbf{45.9} &\textbf{30.9}  &\textbf{36.9}  &\textbf{29.3}  &\textbf{54.8}  &\textbf{49.1}  &\textbf{10.1}  & \textbf{16} \\
\bottomrule
\end{tabular}}
\label{ovvis}
\end{table*}

\begin{figure*}[tb]
  \centering
  \includegraphics[width=.92\linewidth]{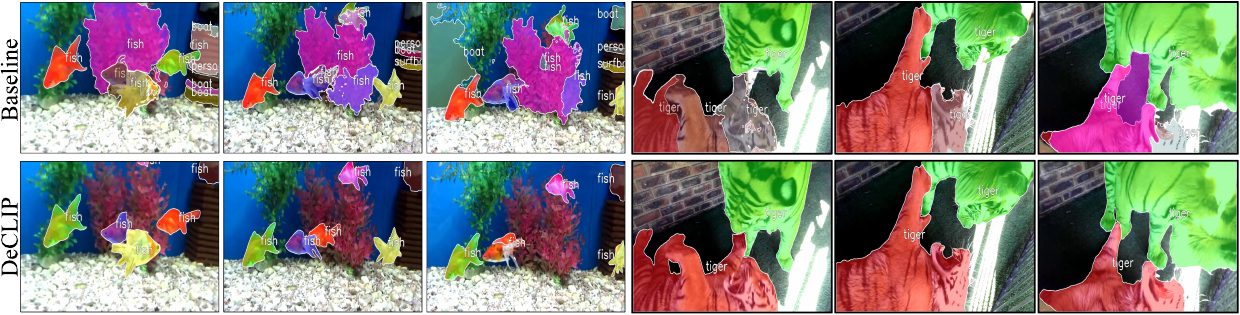}
  \caption{Qualitative comparison of OV video instance segmentation results between CLIP and DeCLIP on the YTVIS21~\cite{MaskTrack_rcnn} dataset. The baseline method is CLIP-VIS~\cite{clip_vis}.}
  \label{fig_video}
\end{figure*}

\subsection{OV 3D Instance Segmentation}
In this task, we use OpenMask3D~\cite{openmask3d} and Open3DIS~\cite{open3dis} as baselines to evaluate the effectiveness of DeCLIP in the 3D OV segmentation task. Apart from replacing the CLIP model in these two methods with DeCLIP, all other settings remain consistent with the original papers.

\mypara{Datasets and Evaluation Metrics.} We evaluate on ScanNet200~\cite{scannet200}, a large-scale dataset featuring a long-tailed distribution with 1,201 training and 312 validation scenes. Detailed multi-modal data preprocessing steps are provided in Appendix~\ref{supp:sec:3d_preprocessing}. We evaluate using the standard AP metrics at IoU thresholds of 0.5 (AP$_{50}$) and 0.25 (AP$_{25}$), as well as the mAP across IoU thresholds from 0.5 to 0.95 in increments of 0.05 (AP). Furthermore, we report the AP for specific category groups, including $\mathrm{AP}_{\mathrm{head}}$ and $\mathrm{AP}_{\mathrm{tail}}$.

\mypara{Results.} The quantitative evaluation of
ScanNet200 is summarized in Tab.~\ref{3D_seg}. By replacing the vanilla CLIP with our proposed DeCLIP, the performance of both OpenMask3D (14.9 vs.\ 18.9 on the $\mathrm{AP}_{\mathrm{tail}}$ metric) and Open3DIS (21.8 vs.\ 27.7 on the $\mathrm{AP}_{\mathrm{tail}}$ metric) is significantly improved. Moreover, our DeCLIP surpasses existing 3D OV instance segmentation methods. Fig.~\ref{fig_3d} presents a qualitative comparison between CLIP and DeCLIP on ScanNet200.

\subsection{OV Video Instance Segmentation}
\label{OV_Video_Instance_Segmentation}

We employ CLIP-VIS~\cite{clip_vis} as the baseline to evaluate the effectiveness of DeCLIP in the video dense perception task. Only the backbone network of CLIP-VIS is replaced with DeCLIP, which remains frozen during training, while all other settings are kept consistent with the original paper. Additionally, since CLIP-VIS extracts multi-scale features from CLIP, to remain consistent with the original paper, we extract features from layers [3, 5, 7, 11] of DeCLIP-B/16 and layers [6, 10, 14, 23] of DeCLIP-L/14, and obtain multi-scale features through downsampling and transposed convolution.

\mypara{Datasets and Evaluation Metrics.} We train CLIP-VIS on the LVIS training set~\cite{lvis}. The evaluation covers the validation and test sets of LV-VIS~\cite{LV-VIS}, the validation sets of YTVIS2019/2021~\cite{MaskTrack_rcnn}, and BURST~\cite{burst}. We specifically employ the OVIS~\cite{OVIS} validation set to assess model robustness against severe occlusions, with detailed occlusion definitions provided in Appendix~\ref{supp:subsec:occlusion}. We adopt the standard mean AP based on Spatiotemporal Intersection-over-Union (IoU) as the evaluation metric. This metric inherently penalizes temporal drift and reflects long-term consistency (definitions in Appendix~\ref{supp:subsec:video_consistency}). Additionally, since OV tasks primarily focus on unseen objects, we separately report the AP for novel categories (AP$_\text{N}$).

\mypara{Experimental Results.}  Tab.~\ref{video_seg} summarizes the quantitative results of applying DeCLIP to the video instance segmentation task. DeCLIP leads to significant performance improvements across all evaluation datasets. The performance gains in spatiotemporal AP indicate that DeCLIP enhances the temporal consistency of video instance segmentation. Notably, the substantial improvement on OVIS demonstrates the robustness of our method against occlusions. We provide a detailed performance analysis across different occlusion levels in Appendix~\ref{supp:subsec:occlusion}. Fig.~\ref{fig_video} presents a qualitative comparison between DeCLIP and vanilla CLIP-VIS on the YTVIS21 dataset.

\subsection{OV Object 6D Pose Estimation}
% \mypara{OV Object 6D Pose Estimation.} 
Oryon~\cite{oryon} defines this task, which targets the inference of the position and orientation of arbitrary objects in three-dimensional space from images or point clouds based on language descriptions. We employ Oryon as the baseline framework. We maintain the default training and inference configurations of the original Oryon model while substituting the image encoder with DeCLIP.

\mypara{Datasets and Evaluation Metrics.} Following Oryon's open-vocabulary setting~\cite{oryon}, we train our model on the ShapeNet6D~\cite{ShapeNet6D,ShapeNetSem} dataset and evaluate it on the REAL275~\cite{REAL275} and Toyota-Light (TOYL)~\cite{Toyota} datasets. Similar to Oryon, we evaluate the pose estimation results using the metrics proposed by the BOP benchmark~\cite{bop}, including AR, VSD, MSSD, and MSPD. For the specific meanings of these metrics, please refer to~\cite{bop}. Additionally, the Oryon model includes a decoding head for predicting object masks. Therefore, we also report the improvement in the object mask prediction quality metric (mIoU) after replacing the backbone network with DeCLIP. 
\begin{table}[tbp]
\centering
\caption{\textbf{Results on OV object 6D pose estimation. }$\dagger$ denotes ``oracle'' performance using GT object masks for pose estimation. $^\star$ indicates our ViT-B results based on the official code, as Oryon uses the ViT-L version of CLIP by default.}

\setlength{\tabcolsep}{2.5pt}

\begin{tabular}{lccccccc}
\toprule
\textbf{Method} & \textbf{Backbone} & \textbf{Dataset} & \textbf{AR$\uparrow$} & \textbf{VSD$\uparrow$} & \textbf{MSSD$\uparrow$} & \textbf{MSPD$\uparrow$} & \textbf{mIoU$\uparrow$} \\
\midrule
Oryon$\dagger$ & ViT-L/14 & \multirow{5}{*}{REAL275} & 46.5 & 32.1 & 50.9 & 56.7 & 100.0 \\
Oryon$^\star$ & ViT-B/16 &  &28.8  &18.9  &32.4  &35.1 &57.4  \\
Oryon & ViT-L/14 &  & 32.2 & 23.6 & 36.6 & 36.4 & 66.5 \\
DeCLIP & ViT-B/16 & &33.7   &23.2   &37.6   &40.2 &65.1 \\
DeCLIP & ViT-L/14 &  & 37.6 & 27.5 & 42.2 & 43.2 & 72.2 \\
\midrule
\rowcolor[HTML]{d4effb} $\Delta$ Score & ViT-B/16 & REAL275 & 
\textcolor[HTML]{2ECC71}{(+4.9)} & 
\textcolor[HTML]{2ECC71}{(+4.3)} & 
\textcolor[HTML]{2ECC71}{(+5.2)} & 
\textcolor[HTML]{2ECC71}{(+5.1)} & 
\textcolor[HTML]{2ECC71}{(+7.7)} \\
\rowcolor[HTML]{d4effb} $\Delta$ Score & ViT-L/14 & REAL275 & 
\textcolor[HTML]{2ECC71}{(+5.4)} & 
\textcolor[HTML]{2ECC71}{(+3.9)} & 
\textcolor[HTML]{2ECC71}{(+5.6)} & 
\textcolor[HTML]{2ECC71}{(+6.8)} & 
\textcolor[HTML]{2ECC71}{(+5.7)} \\
\midrule
Oryon$\dagger$ & ViT-L/14 & \multirow{5}{*}{TOYL} & 34.1 & 13.9 & 42.9 & 45.5 & 100.0 \\
Oryon$^\star$ & ViT-B/16 &  &29.8  &11.1  &38.0  &40.1  &64.3  \\
Oryon & ViT-L/14 &  & 30.3 & 12.1 & 37.5 & 41.4 & 68.1 \\
DeCLIP & ViT-B/16 &  &31.9  &11.8  &40.3  &43.5  &71.4  \\
DeCLIP & ViT-L/14 &  & 32.6 & 12.7 & 41.2 & 43.8 & 68.7 \\
\midrule
\rowcolor[HTML]{d4effb} $\Delta$ Score & ViT-B/16 & TOYL & 
\textcolor[HTML]{2ECC71}{(+2.1)} & 
\textcolor[HTML]{2ECC71}{(+0.7)} & 
\textcolor[HTML]{2ECC71}{(+2.3)} & 
\textcolor[HTML]{2ECC71}{(+3.4)} & 
\textcolor[HTML]{2ECC71}{(+7.1)} \\
\rowcolor[HTML]{d4effb} $\Delta$ Score & ViT-L/14 & TOYL & 
\textcolor[HTML]{2ECC71}{(+2.3)} & 
\textcolor[HTML]{2ECC71}{(+0.6)} & 
\textcolor[HTML]{2ECC71}{(+3.7)} & 
\textcolor[HTML]{2ECC71}{(+2.4)} & 
\textcolor[HTML]{2ECC71}{(+0.6)} \\
\bottomrule
\end{tabular}

\label{tab_6d}
\end{table}

\mypara{Quantitative Results.} As shown in Tab.~\ref{tab_6d}, for the most important metric AR (Average Recall, average of VSD, MSSD, and MSPD), DeCLIP achieves performance improvements of 5.4 on REAL275 and 2.3 on TOYL compared to Oryon. Additionally, DeCLIP outperforms Oryon in object mask prediction quality, achieving improvements of 5.7 on the REAL275 dataset and 0.6 on the TOYL dataset. Notably, Oryon utilizes a dual-encoder architecture (CLIP~\cite{clip} and Swin Transformer~\cite{swin}), with the Swin Transformer compensating for CLIP’s limitations in local representation. However, replacing CLIP with DeCLIP in Oryon still leads to a significant improvement in OV pose estimation performance. This further demonstrates the superiority of DeCLIP, which exhibits pixel-level vision-language alignment capabilities.

\begin{figure}[tbp]
  \centering
  \includegraphics[width=.99\linewidth]{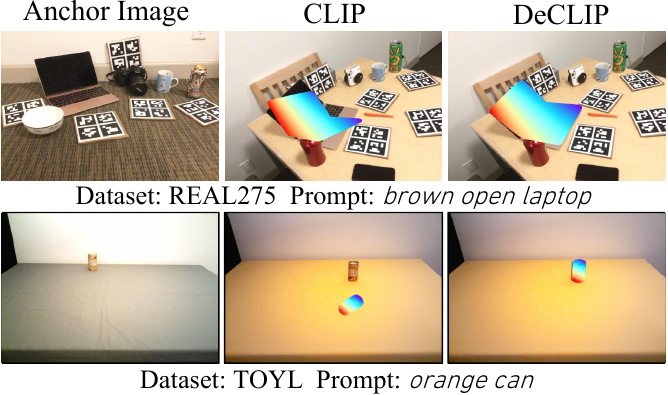}
  \caption{Qualitative comparison of OV pose estimation results between CLIP and DeCLIP on REAL275~\cite{REAL275} and TOYL~\cite{Toyota}. Following Oryon~\cite{oryon}, we color the object model by mapping its 3D coordinates to the RGB space for visualization. }
  \label{fig_6d}
\end{figure}

\mypara{Qualitative Results.} As shown in Fig.~\ref{fig_6d}, for the ``brown open laptop'' object in the REAL275 dataset, both Oryon (CLIP) and Oryon (DeCLIP) are able to estimate the object pose, but DeCLIP demonstrates greater precision in terms of detail. In contrast, for the ``orange can'' object in TOYL (under varying illumination conditions), CLIP fails in pose estimation, whereas DeCLIP still exhibits strong robustness.

\begin{table*}[t]
\caption{\textbf{Comparison with state-of-the-art OV object detection methods.} Caption supervision indicates learning from extra image-text pairs, while CLIP supervision refers to transferring knowledge from CLIP. $^\dagger$: DETR-based detectors \cite{detr}.}
\label{tab1}
\centering
\begin{minipage}[t]{0.47\textwidth}
\centering
\begin{adjustbox}{width=\textwidth,center}
  \begin{tabular}{l|l|l|c}
    \toprule
    Method & Supervision & Backbone & $\text{AP}_{50}^{\text{Novel}}$ \\
    \midrule
    \multicolumn{4}{c}{\textbf{(a) OV-COCO}}\\
    \midrule
    ViLD \cite{vild} & CLIP & RN50  & 27.6 \\
    Detic \cite{detic} & Caption   & RN50 & 27.8 \\
    OV-DETR$^\dagger$ \cite{ovdetr} & CLIP     & RN50     & 29.4 \\
    BARON-KD \cite{wu2023aligning} & CLIP   & RN50     & 34.0 \\
    SAS-Det \cite{sasdet} & CLIP    & RN50 & 37.4 \\
    OV-DQUO$^\dagger$ \cite{ovdquo} & CLIP     & RN50       & 39.2 \\
    RegionCLIP \cite{regionclip} & Captions  & RN50x4       & 39.3 \\
    CORA$^\dagger$ \cite{wu2023cora} & CLIP   & RN50x4    & 41.7 \\
    OV-DQUO$^\dagger$ \cite{ovdquo} & CLIP     & RN50x4       & 45.6 \\
    \midrule
    RO-ViT \cite{kim2023region} &  CLIP   & ViT-L/16     & 33.0 \\
    CFM-ViT \cite{CFM} & CLIP   & ViT-L/16     & 34.1 \\
    F-ViT+CLIPSelf \cite{wu2023clipself} & CLIP  & ViT-B/16  & 37.6 \\
    F-ViT+CLIPSelf \cite{wu2023clipself} & CLIP  & ViT-L/14  & 44.3 \\
    \midrule
    \rowcolor[HTML]{d4effb}F-ViT \cite{wu2023clipself}+DeCLIP & CLIP     & ViT-B/16       &43.3   \textcolor[HTML]{2ECC71}{(+5.7)}\\
     \rowcolor[HTML]{d4effb}F-ViT \cite{wu2023clipself}+DeCLIP & CLIP     & ViT-L/14       &50.2  \textcolor[HTML]{2ECC71}{(+5.9)}  \\
    \rowcolor[HTML]{d4effb}OV-DQUO+DeCLIP$^\dagger$ & CLIP     & ViT-B/16       & \textbf{47.3} \textcolor[HTML]{808080}{(+8.1)}  \\
     \rowcolor[HTML]{d4effb}OV-DQUO+DeCLIP$^\dagger$ & CLIP     & ViT-L/14       & \textbf{49.5} \textcolor[HTML]{808080}{(+3.9)} \\
    \bottomrule
  \end{tabular}
\end{adjustbox}
\end{minipage}
\hfill
\begin{minipage}[t]{0.47\textwidth}
\centering
\begin{adjustbox}{width=\textwidth,center}
  \begin{tabular}{l|l|l|c}
    \toprule
    Method & Supervision & Backbone & $\text{mAP}_{r}$ \\
    \midrule
    \multicolumn{4}{c}{\textbf{(b) OV-LVIS}}\\
    \midrule
    ViLD \cite{vild} &CLIP & RN50 & 16.3 \\
    OV-DETR$^\dagger$ \cite{ovdetr} &CLIP & RN50 & 17.4 \\
    BARON-KD \cite{wu2023aligning} &CLIP & RN50 & 22.6 \\
    RegionCLIP \cite{regionclip} &Caption & RN50x4 & 22.0 \\
    OV-SAM \cite{ovsam} &CLIP & RN50x16 & 24.0 \\
    CORA$^{+}$$^\dagger$ \cite{wu2023cora} &Caption & RN50x4 & 28.1 \\
    F-VLM \cite{fvlm} &CLIP & RN50x64 & 32.8 \\
    \midrule
    CLIPSelf \cite{wu2023clipself} &CLIP & ViT-B/16 & 25.3 \\
    OV-DQUO$^\dagger$ \cite{ovdquo} &CLIP & ViT-B/16 & 29.7 \\
    Detic \cite{detic} & Caption   & Swin-B & 33.8 \\
    RO-ViT \cite{kim2023region} &CLIP & ViT-H/16 & 34.1 \\
    CLIPSelf \cite{wu2023clipself} &CLIP & ViT-L/14 & 34.9 \\
    OV-DQUO$^\dagger$ \cite{ovdquo} &CLIP & ViT-L/14 & 39.3 \\
    \midrule
    \rowcolor[HTML]{d4effb}F-ViT \cite{wu2023clipself}+DeCLIP & CLIP     & ViT-B/16       & 27.1 \textcolor[HTML]{2ECC71}{(+1.8)} \\
    \rowcolor[HTML]{d4effb}F-ViT \cite{wu2023clipself}+DeCLIP & CLIP     & ViT-L/14       & 37.8 \textcolor[HTML]{2ECC71}{(+2.9)}  \\
    \rowcolor[HTML]{d4effb}OV-DQUO$^\dagger$+DeCLIP & CLIP     & ViT-B/16       &\textbf{31.5} \textcolor[HTML]{2ECC71}{(+1.8)}  \\
    \rowcolor[HTML]{d4effb}OV-DQUO$^\dagger$+DeCLIP & CLIP     & ViT-L/14       & \textbf{41.5} \textcolor[HTML]{2ECC71}{(+2.2)}  \\
    \bottomrule
  \end{tabular}
\end{adjustbox}

\end{minipage}
\end{table*}

\subsection{OV Object Detection}

In this task, DeCLIP is evaluated using two baseline models: F-ViT~\cite{wu2023clipself} and OV-DQUO~\cite{ovdquo}. F-ViT is based on the traditional Faster R-CNN~\cite{fasterrcnn} architecture, while OV-DQUO adopts the modern Detection Transformer~\cite{detr}. We replace only the backbone networks of F-ViT and OV-DQUO with DeCLIP, while keeping all other settings consistent with those in the original papers.

\mypara{Datasets and Evaluation Metrics.} Following established settings~\cite{wu2023clipself,ovdquo}, we evaluate our model on the OV-COCO~\cite{mscoco} and OV-LVIS~\cite{lvis} datasets. For OV-COCO, we train on 48 base categories and evaluate on 17 novel categories using the mAP at an IoU threshold of 0.5 ($\text{AP}_{50}^{\text{Novel}}$). For OV-LVIS, we train on Common and Frequent categories while evaluating on the unseen Rare categories. Since Rare categories contain fewer than 10 images in the original dataset, this protocol assesses the generalization capability to unseen concepts within extreme long-tailed distributions. We adopt the mAP on these rare categories at IoU thresholds ranging from 0.5 to 0.95 ($\text{mAP}_{r}$) as the metric. 

\mypara{Quantitative Results.} Tab.~\ref{tab1} presents DeCLIP's performance on OV-COCO and OV-LVIS benchmarks. On OV-COCO, DeCLIP improves the F-ViT \cite{wu2023clipself} baseline by 5.7 and 5.9 mAP, and the OV-DQUO \cite{ovdquo} baseline by 8.1 and 3.9 mAP on novel categories. On OV-LVIS, it achieves gains of 1.8 and 2.9 mAP with F-ViT, as well as 1.8 and 2.2 mAP with OV-DQUO on rare classes. \revise{Additional quantitative analysis by object scale is provided in Appendix~\ref{supp:subsec:object_scale_detection}.} These results validate the superiority of DeCLIP in detecting unseen categories in extreme long-tailed scenarios.

\begin{table*}[tbp]
    \centering
    \caption{Results on OV semantic segmentation. $^\dagger$ indicates results re-experimented by CAT-Seg \cite{catseg}.}
    \begin{adjustbox}{width=.99\textwidth}
    \begin{tabular}{l|ll|ccccccc}
    \toprule
    Method & Backbone & Training Set & ADE847 & Context459 & ADE150 & Context59 & VOC20 & VOC21\\
    \midrule
    ZegFormer$^\dagger$ \cite{ZegFormer} &ViT-B/16  & COCO-Stuff   &5.6   &10.4   & 18.0   &45.5   &89.5   &65.5   \\
    ZSseg \cite{ZSseg}    &ViT-B/16  & COCO-Stuff   &7.0   &-   &20.5   &47.7   & 88.4   &-   \\
    OVSeg \cite{OVSeg}     &ViT-L/14  & COCO-Stuff   &9.0   &12.4   & 29.6   & 55.7   &  94.5   &-  \\
    SAN \cite{SAN}     &ViT-L/14  & COCO-Stuff   &13.7   &17.1   &33.3   &60.2   &95.5   &-   \\
    ODISE \cite{ODISE}     &ViT-L/14  & COCO-Panoptic &11.1   &14.5   & 29.9   & 57.3   &-   &84.6   \\
    MAFT \cite{MAFT}& ConvNeXt-L  &COCO-Stuff  &13.1 &17.0  &34.4  & 57.5   &93.0   &-  \\
    FC-CLIP \cite{fcclip}  & ConvNeXt-L & COCO-Panoptic & 14.8   &18.2  &34.1   & 58.4   & 95.4   & 81.8   \\
    FrozenSeg \cite{frozenseg}& ConvNeXt-L & COCO-Panoptic &14.8 &19.7 &34.4 &- &-&82.5\\
    CAT-Seg \cite{catseg} &ViT-B/16  &COCO-Stuff  &12.0  &19.0  &31.8  &57.5   &94.6   &77.3   \\
    CAT-Seg \cite{catseg}  &ViT-L/14  &COCO-Stuff  &16.0  &23.8  &37.9  &63.3   &97.0   &82.5   \\    \midrule
     \rowcolor[HTML]{d4effb}  
    CAT-Seg+DeCLIP & ViT-B/16  & COCO-Stuff  & 15.3 \textcolor[HTML]{2ECC71}{(+3.3)}  & 21.4 \textcolor[HTML]{2ECC71}{(+2.4)} & 36.3 \textcolor[HTML]{2ECC71}{(+4.5)}  & 60.6 \textcolor[HTML]{2ECC71}{(+3.1)} & 96.6 \textcolor[HTML]{2ECC71}{(+2.0)} & 81.3 \textcolor[HTML]{2ECC71}{(+4.0)} \\
    \rowcolor[HTML]{d4effb}  CAT-Seg+DeCLIP & ViT-L/14  & COCO-Stuff  & \textbf{17.6} \textcolor[HTML]{2ECC71}{(+1.6)}  & \textbf{25.9} \textcolor[HTML]{2ECC71}{(+2.1)} & \textbf{40.7} \textcolor[HTML]{2ECC71}{(+2.8)}  & \textbf{63.9} \textcolor[HTML]{2ECC71}{(+0.6)} & \textbf{97.7} \textcolor[HTML]{2ECC71}{(+0.7)}  & \textbf{83.9} \textcolor[HTML]{2ECC71}{(+1.4)} \\
    \bottomrule
    \end{tabular}
    \end{adjustbox}
\label{tab4}
\end{table*}

\begin{table*}[tbp]
    \centering
    \caption{Comparison with state-of-the-art training-free OV semantic segmentation methods.}
    \begin{adjustbox}{width=.99\textwidth}
    \begin{tabular}{lccc|ccccccc|c}
        \toprule
        \multirow{2.5}{*}{Method} & \multicolumn{3}{c}{\textit{With a background category}} & \multicolumn{5}{c}{\textit{Without background category}} & \multirow{2.5}{*}{Avg.} \\
        \cmidrule(lr){2-4} \cmidrule(lr){5-9}
        & VOC21  & Context60 & COCO-Object & VOC20 & Cityscapes & Context59 & ADE & COCO-Stuff &  \\
        \midrule
        CLIP \cite{clip} & 18.8 & 9.9  & 8.1  & 49.4 & 6.5  & 11.1 & 3.1  & 5.7  & 14.1 \\
        \midrule
        MaskCLIP \cite{maskclip} & 43.4 & 23.2 & 20.6 & 74.9 & 24.9 & 26.4 & 11.9 & 16.7 & 30.3 \\
        GroupViT \cite{groupvit} & 52.3 & 18.7 & 27.5 & 79.7 & 18.5 & 23.4 & 10.4 & 15.3 & 30.7 \\
        ReCo \cite{reco} & 25.1 & 19.9 & 15.7 & 57.7 & 21.6 & 22.3 & 11.2 & 14.8 & 23.5 \\
        TCL \cite{tcl}& 51.2 & 24.3 & 30.4 & 77.5 & 23.5 & 30.3 & 14.9 & 19.6 & 33.9 \\
        SCLIP \cite{sclip}& 59.1 & 30.4 & 30.5 & 80.4 & 32.2 & 34.2 & 16.1 & 22.4 & 38.2 \\
        ClearCLIP \cite{clearclip} &51.8  &32.6  &33.0 &80.9  &30.0  &35.9  &16.7  &23.9  &38.1  \\
        NACLIP~\cite{naclip} &58.9  &32.2  &33.2 &79.7  &35.5  &35.2  &17.4  &23.3  &39.4\\
        CLIPDINOiser~\cite{clipdino}&62.1  &32.4  &34.8 &80.9  &31.7  &35.9  &20.0  &24.6  &40.3 \\ 
        ResCLIP~\cite{resclip}&61.1  &33.5 &35.0 &\textbf{86.0}  &35.9  &36.8  &18.0  &24.7  &41.4 \\ 
        ProxyCLIP~\cite{proxyclip} &61.3  &35.3  &37.5 &80.3  &\textbf{38.1}  &39.1  &20.2  &26.5 &42.3 \\ 
        \midrule
         \rowcolor[HTML]{d4effb} 
        DeCLIP (Ours) &\textbf{64.1}   &\textbf{37.9}  &\textbf{38.7}  &\underline{85.3}  &\underline{35.7}  &\textbf{41.6} &\textbf{23.1}  &\textbf{26.8} &\textbf{44.1} \\
        \bottomrule
    \end{tabular}
    \end{adjustbox}
\label{tab5}
\end{table*}

\subsection{OV Semantic Segmentation}
\label{sec:ov_seg}
We integrate DeCLIP into the CAT-Seg \cite{catseg} baseline. For all experiments, we follow the default training and inference settings of vanilla CAT-Seg, with the only modification being the replacement of the image encoder with DeCLIP.

\mypara{Dataset and Evaluation Metrics.} We train models on COCO-Stuff~\cite{cocostuff} following the original CAT-Seg~\cite{catseg} method, and evaluate them on ADE20K~\cite{ade20k}, PASCAL VOC~\cite{voc}, and PASCAL-Context~\cite{context} datasets. The ADE20K, PASCAL-Context, and PASCAL VOC datasets each contain two sets of categories: A-150 (150 classes) and A-847 (847 classes)~\cite{ZegFormer} for ADE20K, PC-59 (59 classes) and PC-459 (459 classes) for PASCAL-Context, and PAS-20 (20 object classes) and PAS-21 (20 classes plus background) for PASCAL VOC. We use mIoU as the evaluation metric for all experiments.

\mypara{Quantitative Results.} Tab.~\ref{tab4} presents the performance of CAT-Seg~\cite{catseg} with DeCLIP as the backbone on OV semantic segmentation benchmarks. The results show that DeCLIP significantly enhances segmentation performance on all datasets. Notably, even with ViT-B/16, CAT-Seg nearly surpasses existing SOTA methods using substantially larger encoders like ConvNeXt-L~\cite{convnext}. With ViT-L/14, the model achieves new SOTA results in OV semantic segmentation. \revise{We attribute the marginal gain under the CAT-Seg protocol to the strong backbone provided by the conference-version DeCLIP and to supervised fine-tuning on COCO-Stuff, which can attenuate the incremental gains from RCC and SD-GSC.}

\subsection{Training-Free OV Semantic Segmentation}
\label{sec:training_free_ov_seg}
Unlike the OV semantic segmentation task, which requires training task-specific segmentation components using the base-category dataset, this task directly leverages CLIP's dense feature map $\mathbf{X}_{\text{dense}}$ for OV semantic segmentation. Following existing methods~\cite{CLIPtrase,sclip,clearclip}, we directly compute the cosine similarity between each pixel in the DeCLIP feature map and all category texts to achieve zero-shot semantic segmentation.
For all datasets, we generate textual descriptions by utilizing the standard ImageNet prompts~\cite{clip} in conjunction with their respective class names. No post-processing steps are applied. The low-resolution prediction result is up-sampled to the original resolution to obtain the final segmentation map. 

\mypara{Dataset and Evaluation Metrics.} The evaluation of this task does not require training and thus does not include a training set. The validation set comprises six widely used semantic segmentation benchmark datasets: PASCAL VOC 2012~\cite{voc}, PASCAL Context~\cite{context}, Cityscapes~\cite{cityscape}, ADE20K~\cite{ade20k}, COCO Stuff~\cite{mscoco}, and COCO Object~\cite{cocostuff}. For datasets that include a background category, we denote them as VOC21 and Context60, while those without a background category are denoted as VOC20 and Context59. We employ mIoU as the evaluation metric for these benchmarks.

\mypara{Quantitative Results.} As shown in Tab.~\ref{tab5}, DeCLIP outperforms all existing methods in terms of average mIoU across eight benchmarks, achieving an average mIoU of 44.1. This result highlights the effectiveness of our approach in enhancing the discriminability and spatial consistency of VLM features.

\begin{figure}[tbp]
    \centering
    \begin{minipage}{0.49\columnwidth}
        \centering
        \includegraphics[width=\linewidth]{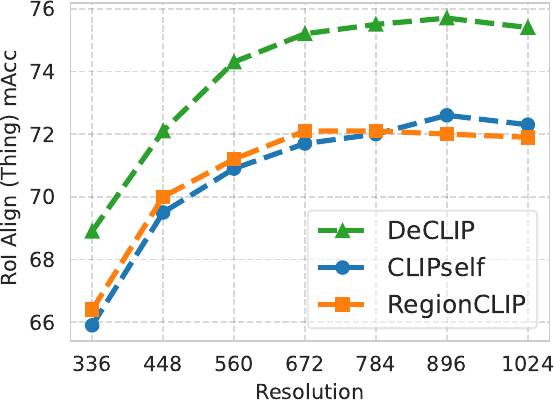}
    \end{minipage}
    \begin{minipage}{0.49\columnwidth}
        \centering
        \includegraphics[width=\linewidth]{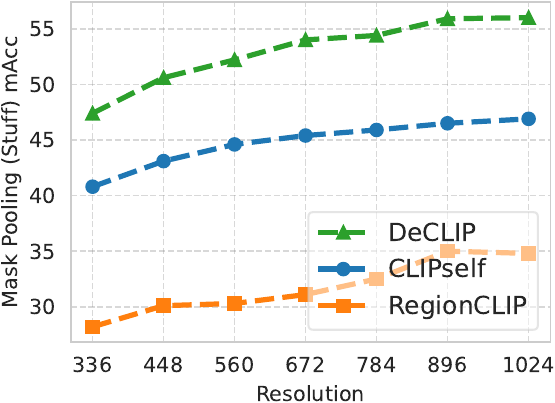}
    \end{minipage}
    \caption{\textbf{Comparison with existing regional V-L alignment methods on OV region classification.} DeCLIP consistently outperforms existing methods across various resolutions.}
    
    \label{fig6}
\end{figure}

\begin{table}[tbp]
    \centering
    \setlength{\tabcolsep}{2.5pt}
    \caption{\textbf{Component analysis of DeCLIP.} \revise{We report OV detection performance on OV-COCO using mAP and training-free OVSS performance on PASCAL VOC and COCO-Obj using mIoU. Content+Context corresponds to the conference version. RCC and SD-GSC are newly added components.}}
    \begin{tabular}{cccc|cccc}
        \toprule
        \multirow{2}{*}{\makecell[c]{Content\\Distill.}} &
        \multirow{2}{*}{\makecell[c]{Context\\Distill.}} &
        \multirow{2}{*}{\makecell[c]{RCC}}  &
        \multirow{2}{*}{\makecell[c]{SD-GSC}}         &
        \multirow{2}{*}{\makecell[c]{OV-COCO}} &
        \multirow{2}{*}{\makecell[c]{VOC21}} &
        \multirow{2}{*}{\makecell[c]{COCO-Obj}} &
        \multirow{2}{*}{\makecell[c]{\textbf{Avg.}}} \\
        &&&&&&& \\
        \midrule
        & & & &17.5  &43.4  &20.6  &27.2 \\
        \ding{51}& & & &37.6  &42.3  &23.2  &34.4 \\
        &\ding{51} & & &37.4  & 59.7 & 36.4   &44.5 \\
        \ding{51}&\ding{51} & & &41.1  &59.9  &36.5  &45.8 \\
        \ding{51}&\ding{51} &\ding{51} & &42.5  &60.5
  &36.9  &46.6 \\
        \ding{51}&\ding{51} & &\ding{51} &41.9  &63.7  &38.5  &48.0 \\
        \rowcolor[HTML]{d4effb} \ding{51}&\ding{51} &\ding{51} &\ding{51}     &\textbf{43.3}  &\textbf{64.1}   &\textbf{38.7}  &\textbf{48.7} \\
        \bottomrule
    \end{tabular}
\label{module_ablation}
\end{table}

\subsection{OV Region Classification}
Following existing methods~\cite{wu2023clipself,scd,regionclip}, we extract region features using ground-truth boxes or masks via RoI Align~\cite{maskrcnn} or mask pooling. Each region is classified by cosine similarity with category text embeddings. Using ground-truth regions isolates classification from localization errors, directly evaluating the V-L alignment of dense features.

\mypara{Datasets and Evaluation Metric.} This evaluation is training-free and requires no training set. We utilize Top-1 mean accuracy (mAcc) to assess classification performance on boxes and masks within the COCO Panoptic dataset~\cite{coco_panoptic}. This metric quantifies the quality of V-L alignment, where higher accuracy indicates superior correspondence between dense visual features and text embeddings.

\mypara{Quantitative Results.} As illustrated in Fig.~\ref{fig6}, DeCLIP consistently outperforms existing methods in region recognition across all resolutions in terms of Top-1 mAcc. More extensive analyses on V-L alignment are provided in Appendix~\ref{supp:subsec:vl-align-analy}.

\begin{figure}[tbp]
\centering
\includegraphics[width=0.99\linewidth]{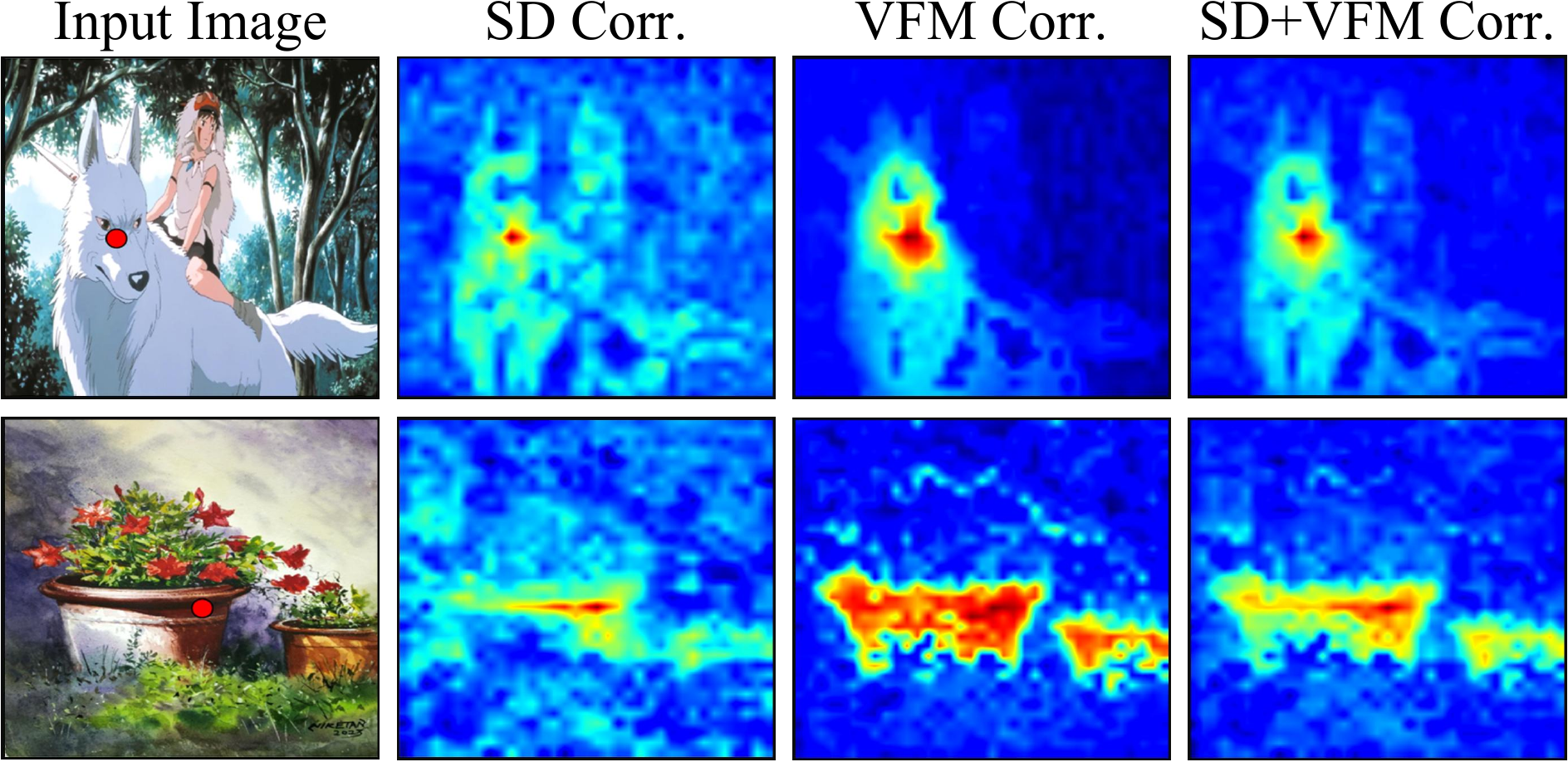}
\caption{\textbf{Comparison of semantic affinity maps between SD and VFM features.} Both SD features and the channel concatenation of SD and VFM features exhibit lower semantic affinity capability than using VFM features alone.}
\label{ablation_vfm_sd_vis}
\end{figure}

\subsection{Ablation Study}
\label{ablation_study_sec}

\begin{table}[tbp]
\centering
\caption{\textbf{Ablation Study on the collaboration strategies between VFM and SD.} ``Concat" combines VFM and SD features along the channel dimension. $^\star$ represents the weighted feature concatenation strategy. ``DirectAlign" refers to the direct distillation of SD self-attention maps. }
\setlength{\tabcolsep}{6pt} 
\begin{tabular}{lccccc}
\toprule
Method & VOC20 & Context59 & Cityscapes & ADE & COCO-Stf \\
\midrule
Concat &75.2 &30.6 &24.4 &19.2 &19.9 \\
Concat$^\star$ &84.4 &39.1 &33.7 &21.7 &25.1 \\
DirectAlign &79.6 & 32.0 &28.5 &18.0 &21.3 \\
\rowcolor[HTML]{d4effb} SD-GSC &\textbf{85.3} &\textbf{41.6} &\textbf{35.7} &\textbf{23.1} & \textbf{26.8} \\
\bottomrule
\end{tabular}
\label{ablation_vfm_sd}
\end{table}

\mypara{Module Ablation.} \revise{We conduct the ablation study on the F-ViT baseline for OV detection and training-free OVSS. Both settings freeze the visual encoder and thus more directly evaluate dense representation quality. As shown in Tab.~\ref{module_ablation}, the introduction of content distillation significantly boosts detection performance on OV-COCO but provides marginal benefits for segmentation. In contrast, context distillation enhances both tasks effectively. The combination of these two strategies establishes a strong foundation and achieves an AP of 41.1 on OV-COCO. Furthermore, imposing the RCC constraint refines feature alignment and leads to a consistent average improvement. Notably, the introduction of SD-GSC supplements detailed semantic information and results in a substantial average increase of 2.2. Since SD-GSC is used only during training, it introduces no additional inference cost. Ultimately, the full model raises the average score to 48.7 and demonstrates a comprehensive gain of 2.9.}

\begin{figure*}[tbp]
  \centering
  \includegraphics[width=.9\linewidth]{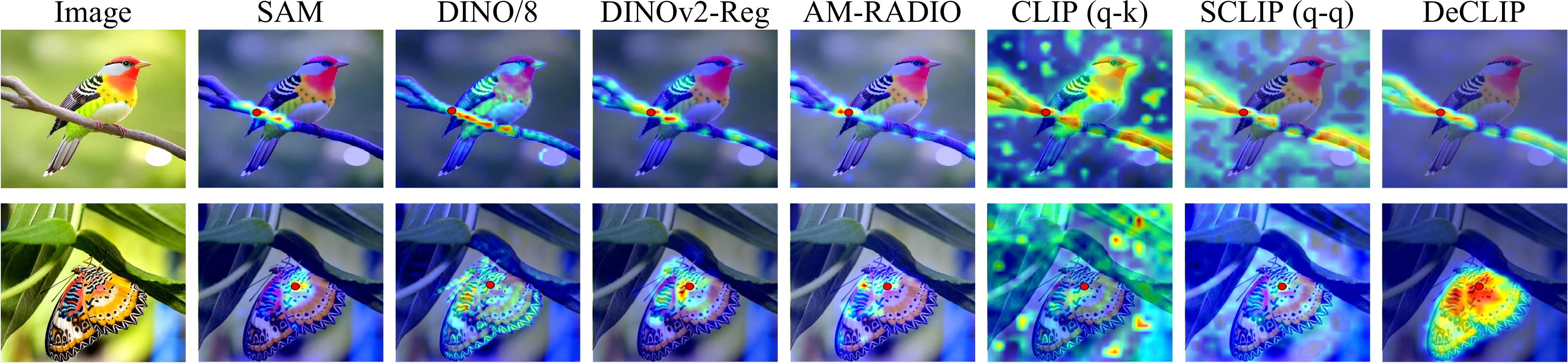}
  \caption{\textbf{Qualitative comparison of attention maps.} The comparison methods include: SAM~\cite{sam,sam2}, trained on large-scale mask annotations; DINO~\cite{dino} and DINOv2~\cite{dinov2} with registers~\cite{register}, trained using a self-supervised approach; AM-RADIO~\cite{am_radio}, utilizing multi-teacher distillation from CLIP, SAM, and DINOv2 models; and SCLIP~\cite{sclip}, incorporating correlative self-attention mechanisms. The query image token is marked in red. }
  \label{vfm_vis_ablation}
\end{figure*}

\mypara{Analysis of Semantic Completion Strategies.} We investigate semantic completion sources and integration strategies. We first evaluate the impact of replacing Stable Diffusion with discriminative or reconstruction-based alternatives, specifically SAM~\cite{sam} and I-JEPA~\cite{ijepa}. Quantitative comparisons in Appendix~\ref{supp:subsec:sd_gsc_quan} demonstrate that our SD-based approach consistently yields superior performance across OV segmentation and detection benchmarks. Furthermore, qualitative visualizations in Appendix~\ref{supp:subsec:sd_gsc_qual} confirm that the generative priors of SD provide better global structural integrity than the boundary-focused constraints of SAM or the masked prediction features of I-JEPA. \revise{The effects of the Stable Diffusion layer and timestep are further analyzed in Appendix~\ref{supp:subsec:sd_layer_timestep}.} Regarding the integration strategy, we compare our method (Sec.~\ref{declip}) against feature concatenation~\cite{sd+dino} and direct attention distillation. As reported in Tab.~\ref{ablation_vfm_sd}, our proposed strategy achieves the highest performance while simple concatenation provides only marginal gains. We attribute this to the weak semantic discriminability of SD features (Fig.~\ref{ablation_vfm_sd_vis}). This supports SD as a structural prior rather than a direct VFM feature source. 

\begin{table}[tb]
\centering
\caption{Ablation studies on different VLM baselines. $\dagger$ indicates re-implementation based on the official code.}
\setlength{\tabcolsep}{1pt}
\begin{tabular}{lcccccc}
\toprule
VLM Model & VOC21 & Context60 & COCO-Obj & Cityscapes & ADE & COCO-Stf \\
\midrule
OpenAI-CLIP~\cite{clip} &18.8 &9.9 &8.1 &6.5 &3.1 &5.7 \\
ClearCLIP~\cite{clearclip} &51.8 &32.6 &33.0 &30.0 &16.7 &23.9 \\
\rowcolor[HTML]{d4effb} DeCLIP &\textbf{64.9} &\textbf{36.3} &\textbf{39.1} &\textbf{35.5} &\textbf{19.8} &\textbf{24.6} \\
\midrule
EVA-CLIP~\cite{evaclip} &23.4 &12.8 &15.3 &12.8 &7.7 &9.7 \\
EVA-ClearCLIP$^\dagger$ &47.0 &29.7 &30.2 &26.3 &16.7 &20.4 \\
\rowcolor[HTML]{d4effb}  DeCLIP &\textbf{64.1} &\textbf{37.9} &\textbf{38.7} &\textbf{35.7} &\textbf{23.1} &\textbf{26.8} \\
\bottomrule
\end{tabular}
\label{vlm_ablation}
\end{table}

\begin{figure*}[htbp]
  \centering
  \includegraphics[width=.9\linewidth]{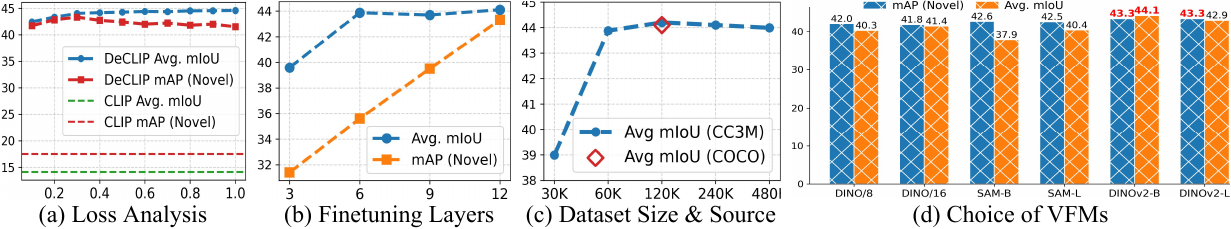}
  \caption{\textbf{Ablation studies of DeCLIP method.} (a), (b), (c), and (d) denote the weight of the context loss, the number of fine-tuned layers in CLIP, the domain and scale of the distillation dataset, and the impact of VFM selection on OV dense perception performance (detection and segmentation), respectively. The segmentation results are averaged over multiple datasets. All experiments are performed using the ViT-B version of DeCLIP.}
  \label{four_ablation_studies}
\end{figure*}

\mypara{Sensitivity Analysis of $\lambda$.} We perform an ablation study to examine the relationship between the hyperparameter $\lambda$ and dense perception performance, as shown in Fig.~\ref{four_ablation_studies}(a). Experimental results demonstrate that our method exhibits strong robustness, and the dense perception performance of DeCLIP does not fluctuate drastically with changes in $\lambda$. Furthermore, the results indicate that $\lambda=0.25$ strikes a good balance between OV detection and segmentation. \revise{Crop sampling sensitivity is analyzed in Appendix~\ref{supp:subsec:crop_sampling_sensitivity}.}

\mypara{Sensitivity to Number of Fine-Tuning Layers.} We examine the relationship between the number of fine-tuning attention blocks and dense perception performance in Fig.~\ref{four_ablation_studies}(b). We observe that as the number of fine-tuning layers increases, the performance of OV detection and segmentation continuously improves. Therefore, we chose to fine-tune all attention blocks in the implementation of DeCLIP.

\mypara{Analysis on VLM Baselines.} DeCLIP utilizes EVA-CLIP as the default VLM baseline~\cite{wu2023clipself}. We assess the generalization of our method across different VLMs and investigate whether EVA-CLIP introduces additional performance advantages. As shown in Tab.~\ref{vlm_ablation}, DeCLIP maintains robust performance with both EVA-CLIP and OpenAI-CLIP, consistently outperforming state-of-the-art methods. Moreover, we conduct sanity checks in Appendix~\ref{supp:subsec:sanity} to exclude potential performance gains attributed solely to the EVA-CLIP baseline. Experimental results demonstrate that it introduces no observable biases.

\mypara{Generalization and Robustness Analysis.} We evaluate DeCLIP's robustness across diverse data sources, application domains, and visual conditions. DeCLIP adopts the widely used COCO~\cite{mscoco} dataset with 118k images as the default for distillation, following mainstream approaches~\cite{scd,regionclip,wu2023clipself}. This ensures fair comparisons and consistent experimental settings with existing methods. 
Additionally, we investigate DeCLIP's sensitivity to both the source and size of the distillation dataset. Specifically, we employ the CC3M~\cite{cc3m} dataset to assess the impact of data scale and domain on performance. CC3M is a dataset widely utilized for CLIP training in the literature, and its source differs from COCO. Fig.~\ref{four_ablation_studies}(c) demonstrates that DeCLIP remains robust to these variations. Beyond training data, we assess generalization under significant domain shifts by applying the COCO-distilled DeCLIP directly to remote sensing scenarios (details in Appendix~\ref{supp:subsec:cross_domain}). The results indicate strong transferability even without domain-specific fine-tuning. Furthermore, we analyze model reliability under visual degradation (details in Appendix~\ref{supp:subsec:low_quality}). Comprehensive tests on 15 corruption types reveal that DeCLIP maintains superior stability compared to baselines.

\mypara{Sensitivity to VFM Type and Size.} We further investigate the effect of different VFM choices on OV dense perception performance. This analysis encompasses various architectures, including SAM~\cite{sam}, DINO~\cite{dino}, and DINOv2~\cite{dinov2}, across different scales (e.g., ViT-B and ViT-L).  As shown in Fig.~\ref{four_ablation_studies}(d), variations in VFM type and size exert minimal influence on OV object detection. DINOv2 yields the most favorable results, marginally outperforming SAM and DINO. In contrast, VFM selection significantly affects semantic segmentation tasks that require finer perceptual granularity. DINOv2-B achieves the highest performance in OV semantic segmentation, followed by DINOv2-L, the DINO series, and the SAM series. This can be attributed to the stronger semantic correspondence capability of DINOv2. Overall, DINOv2-B provides the best comprehensive performance. \revise{Fig.~\ref{vfm_vis_ablation} compares DeCLIP with existing visual backbones. Experiments show that DeCLIP more effectively focuses on regions spatially or semantically associated with the query image token.}

\mypara{Inference and Efficiency Analysis.} We quantitatively evaluate the efficiency of DeCLIP. Appendix~\ref{supp:subsec:inference_efficiency} verifies that DeCLIP serves as a representation learning framework without altering the original model architecture. It preserves the inference latency, parameter count, and memory footprint of vanilla CLIP. This efficiency advantage extends to downstream task-specific models. As demonstrated in Appendix~\ref{supp:subsec:realtime-vis}, integrating DeCLIP into a video segmentation framework enhances downstream performance while preserving the inference frame rate of the baseline. Furthermore, we validate the scalability of DeCLIP in resource-constrained scenarios. Appendix~\ref{supp:subsec:lightweight_backbone} details the application of decoupled learning to the lightweight TinyCLIP architecture~\cite{tinyclip} (39M parameters). Results indicate that DeCLIP generalizes effectively to these compact models. Finally, Appendix~\ref{supp:subsec:inference_optimization} demonstrates that DeCLIP is orthogonal to existing inference optimization methods, enabling an 86\% acceleration with FP8 precision while incurring negligible accuracy degradation.

\subsection{Limitations and Failure Analysis}
\label{failure_case}
Despite significant improvements in OV dense perception, DeCLIP exhibits limitations in certain extreme scenarios. As detailed in Appendix~\ref{supp:sec:failure_case}, the model effectively rectifies baseline classification errors for small objects and refines segmentation boundaries. However, challenges persist in identifying extremely tiny objects in rare categories, segmenting instances under severe occlusion, and handling adverse lighting conditions. These findings suggest future directions for enhancing fine-grained visual-language alignment.

%% file: sec/6_conclusion.tex
\section{Conclusion}
In this paper, we systematically analyze the attention maps of CLIP and VFM, revealing that CLIP’s core limitation in dense perception tasks is that its image tokens cannot effectively aggregate information from semantically relevant regions. To address this issue, we propose DeCLIP, a novel unsupervised fine-tuning framework that enhances dense representations of CLIP via a decoupled feature enhancement strategy. Extensive experiments on more than six diverse OV dense perception benchmarks, spanning 2D, 3D, and video data, demonstrate that DeCLIP achieves significant improvements over the original CLIP and other state-of-the-art methods. These results highlight the potential of DeCLIP as a foundational model for OV dense perception tasks. 

%% file: sec/7_appendix.tex
\newpage
\ifdefined\maketitlesupplementary
    \maketitlesupplementary
\else
    \twocolumn[{
        \centering
        \Large \textbf{Supplementary Material} \\
        \vspace{1.0em}
    }]
\fi

\section*{Overview}
\noindent This material provides supplementary details to the main paper, including the following sections:

\setcounter{section}{0}
\renewcommand{\thesection}{\Alph{section}}  
\renewcommand{\theHsection}{appendix.\arabic{section}} 

{
\hypersetup{colorlinks=true, linkcolor=black} 

\titlecontents{section}
  [2.5em]  
  {\vspace{.5em}}  
  {\bfseries\contentslabel[\textbullet\ (\thecontentslabel)]{2.5em}}  
  {}  
  {\bfseries\titlerule*[6pt]{.}\contentspage}  
  [\vspace{.5em}] 

\titlecontents{subsection}
  [4.3em]  
  {}  
  {\contentslabel[\textbf{\textendash}\ (\thecontentslabel)]{3.3em}}  
  {}  
  {\titlerule*[6pt]{.}\contentspage}  

\startcontents[appendix]  
\printcontents[appendix]{}{1}{}
}

\begin{figure}[hbt]
    \centering
    \includegraphics[width=0.99\linewidth]{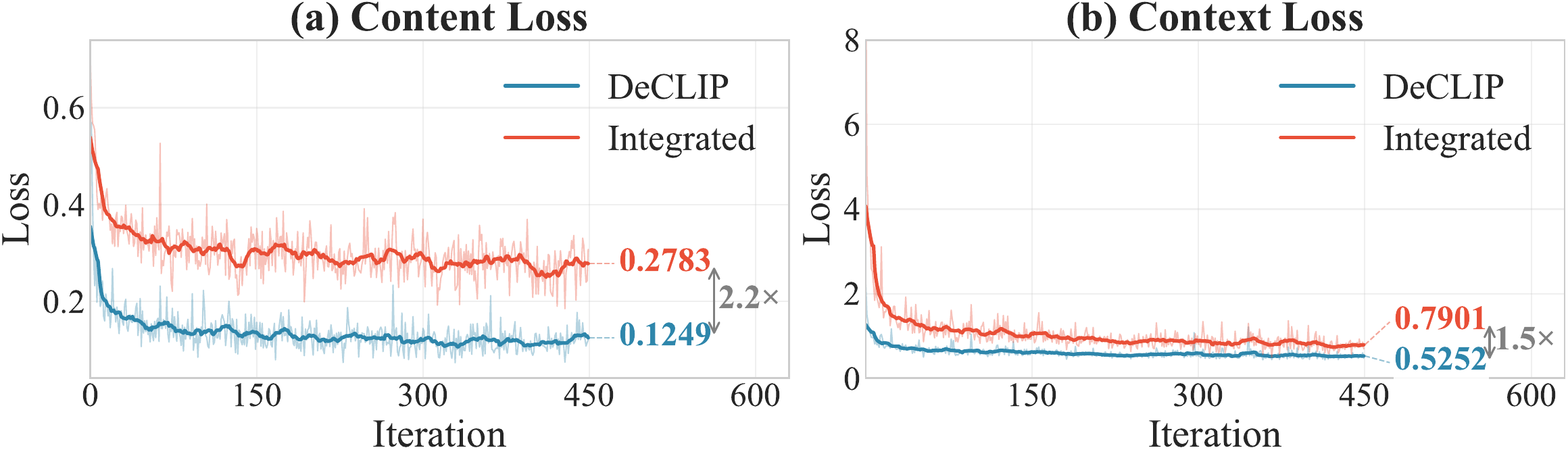}
    \caption{\textbf{Analysis of decoupled learning and integrated distillation.}  Training loss curves for \textbf{(a)} content losses and \textbf{(b)} context losses. The proposed decoupled learning method (blue) demonstrates faster convergence, reduced final loss, and enhanced training stability.}
    \label{fig:loss_2}
\end{figure}

\section{Extended Analysis of Decoupled Learning}
\label{supp:sec:analysis}
In this section, we provide an extended analysis comparing the efficacy of the proposed decoupled learning strategy with the vanilla integrated learning method.

\subsection{Analysis of Individual Training Dynamics}
\label{supp:subsec:individual_training}

\begin{figure}[hbt]
    \centering
    \includegraphics[width=0.99\linewidth]{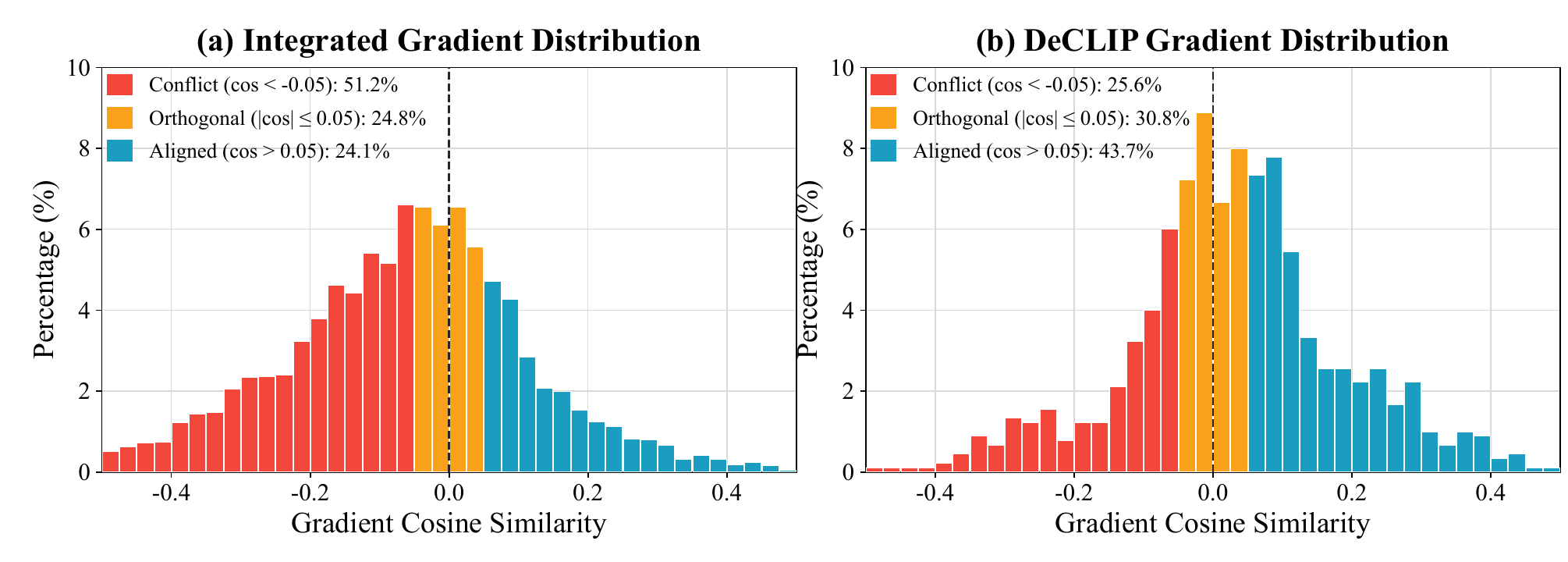}
    \caption{\revise{\textbf{Gradient conflict analysis for content/context decoupling.} We compare the gradient cosine similarity distributions of the content and context losses for \textbf{(a)} the integrated baseline and \textbf{(b)} DeCLIP. Red, orange, and blue bars denote conflicting, orthogonal, and aligned gradients, respectively.}}
    \label{fig:supp_loss_grad_distribution}
\end{figure}

To complement the main text analysis, we examine the individual training dynamics in Fig.~\ref{fig:loss_2}. The decoupled strategy (blue curve) exhibits superior stability and faster convergence compared to the integrated baseline (red curve). Quantitatively, the final content and context losses of the integrated approach are approximately $2.2\times$ and $1.5\times$ higher than those of DeCLIP. These results confirm that separating optimization objectives mitigates interference and leads to a better local minimum. This improved convergence justifies the performance superiority observed in dense perception tasks.
\revise{Fig.~\ref{fig:supp_loss_grad_distribution} shows that DeCLIP yields more positive gradient cosine similarities than integrated learning, indicating fewer conflicts between the content and context objectives.}

\subsection{\revise{Training Stability Statistics}}
\label{supp:subsec:training_stability_statistics}
\revise{We further quantify the training stability of integrated distillation and DeCLIP using the logged total loss over the final epoch. As shown in Tab.~\ref{tab:training_stability_statistics}, DeCLIP obtains a lower final loss and smaller loss fluctuation than integrated distillation. This trend is consistent with the reduced gradient conflict ratio, indicating that decoupling improves optimization compatibility and training stability.}

\begin{table}[t]
\centering
\setlength{\tabcolsep}{5pt}
\caption{\revise{\textbf{Training stability comparison between integrated distillation and DeCLIP.} Mean and standard deviation (Std.) are computed from the logged total loss over the final epoch. The coefficient of variation (CV) is computed as Std./Mean and measures relative loss fluctuation. Lower Std. and CV indicate more stable training.}}
\label{tab:training_stability_statistics}
\resizebox{\linewidth}{!}{%
\begin{tabular}{lcccc}
\toprule
Method & Final loss mean & Std. & CV & Gradient conflict \\
\midrule
Integrated Distillation & 1.0798 & 0.1598 & 14.8\% & 51.2\% \\
DeCLIP & \textbf{0.6500} & \textbf{0.0679} & \textbf{10.5\%} & \textbf{25.6\%} \\
\bottomrule
\end{tabular}
}
\end{table}

\subsection{\revise{Proxy-Token Behavior Analysis}}
\label{supp:subsec:proxy_token_behavior}

\begin{figure*}[tbh]
    \centering
    \includegraphics[width=0.99\linewidth]{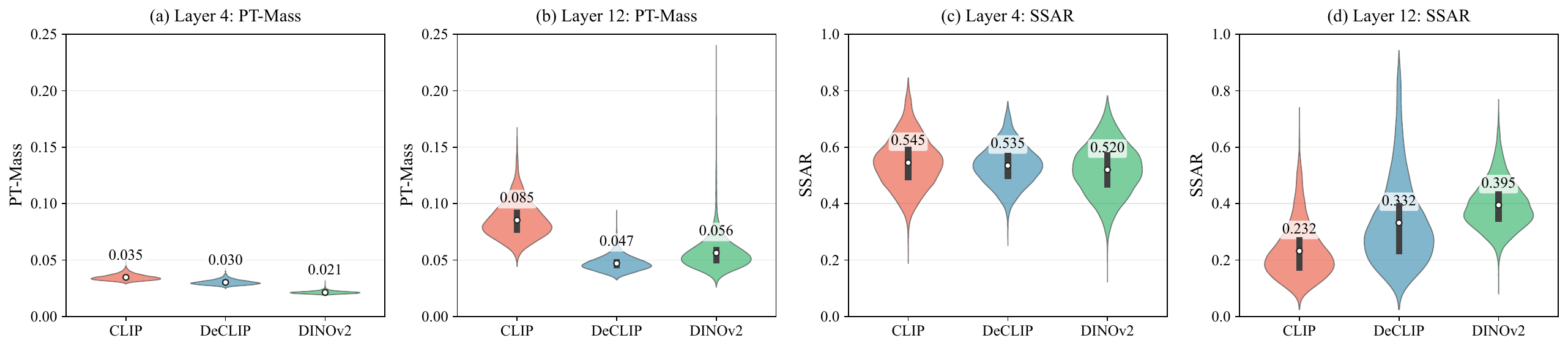}
    \caption{\revise{\textbf{Large-scale proxy-token statistics.} We measure Proxy-Token Mass (PT-Mass) and Same-Semantic-Region Attention Ratio (SSAR) on 5{,}000 COCO-Panoptic validation images. PT-Mass is the attention mass on the top-$k$ proxy tokens, and SSAR is the attention mass within the same semantic region. Numbers are means.}}
    \label{fig:proxy_token_statistics}
\end{figure*}

\mypara{\revise{Large-Scale Statistical Analysis.}} \revise{We further examine whether the proxy-token phenomenon in CLIP is a general deep-layer pattern rather than a single-sample observation. Fig.~\ref{fig:proxy_token_statistics} reports proxy-token concentration and same-region attention statistics. CLIP shows stronger proxy concentration and weaker same-region attention in deep layers, while DeCLIP reduces this pattern and improves semantic attention consistency.}

\begin{figure}[tbh]
    \centering
    \includegraphics[width=0.99\linewidth]{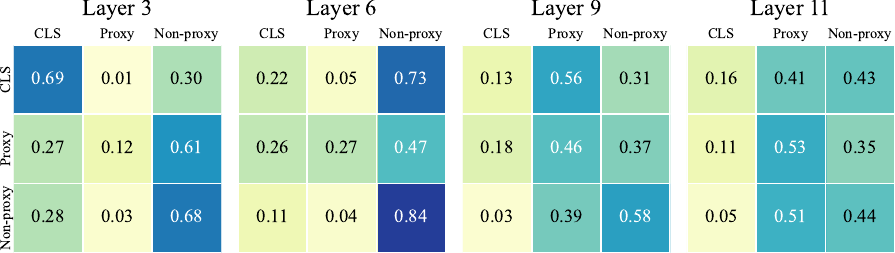}
    \caption{\revise{\textbf{Attention interaction in CLIP.} We select the top-10 proxy tokens at Layer~11 and trace them across layers over 500 COCO images. Rows and columns denote query and key groups, respectively. Values are mean attention mass.}}
    \label{fig:proxy_token_interaction}
\end{figure}

\begin{figure*}[tbh]
    \centering
    \includegraphics[width=0.98\linewidth]{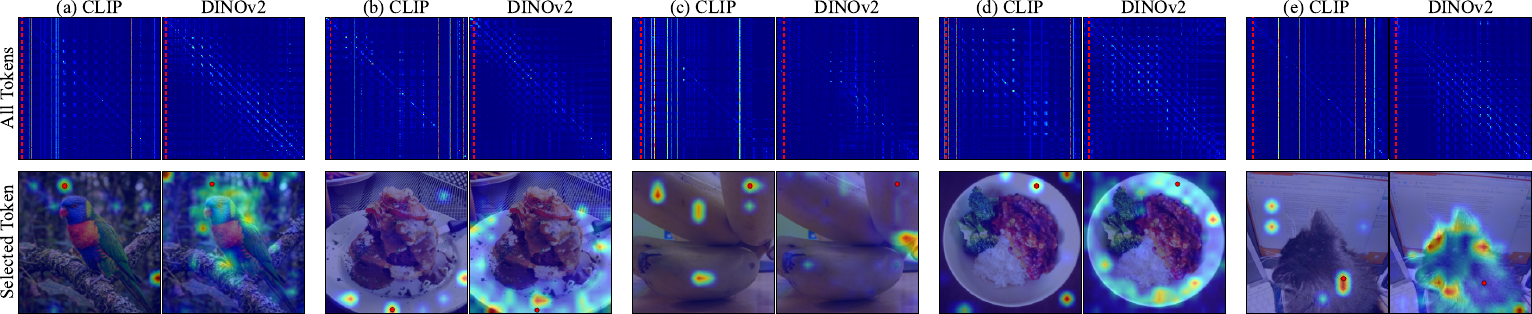}
    \caption{\revise{\textbf{Proxy-token query attention.} We compare CLIP and DINOv2 at Layer~11. CLIP shows vertical proxy-token stripes and concentrated proxy-query responses, while DINOv2 shows spatially distributed attention.}}
    \label{fig:proxy_token_as_query}
\end{figure*}

\mypara{\revise{Proxy Token as Query.}} \revise{We also analyze the attention pattern when selected proxy tokens are used as query tokens. As shown in Fig.~\ref{fig:proxy_token_interaction}, shallow-layer queries mainly attend to non-proxy tokens. In deep layers, both non-proxy and proxy-token queries assign substantially more attention to the proxy set. This confirms that selected proxy tokens act as attention sinks in CLIP's deep layers. Fig.~\ref{fig:proxy_token_as_query} provides qualitative examples.}

\subsection{Extended Quantitative Analysis}
\label{supp:subsec:quantitative_analysis}
To supplement the main results, we further provide a quantitative comparison between decoupled and integrated learning on dense perception tasks. As shown in Tab.~\ref{tab:decouple_detection}, the OV detection results indicate that decoupled learning surpasses the integrated baseline in both $\text{AP}_{50}^{\text{Base}}$ and $\text{AP}_{50}^{\text{Novel}}$ metrics. Additionally, Tab.~\ref{tab:decouple_seg} details the performance on OV semantic segmentation. DeCLIP achieves an mIoU of 44.1 and exceeds the integrated baseline by 5.4 points. These experimental outcomes demonstrate the superiority of the proposed decoupled learning method.

\begin{table}[t]
    \centering
    \caption{\textbf{Comparison of decoupled and integrated learning in OV detection.} Results are reported as $\text{AP}_{50}$ on base, novel, and all categories. Decoupled learning yields consistent gains over integrated distillation, with notably larger improvements on novel categories.}
    \setlength{\tabcolsep}{2.3pt}
    \begin{tabular}{lccccc}
    \toprule
        Method & Backbone & Detector & $\text{AP}_{50}^{\text{Base}}$ & $\text{AP}_{50}^{\text{Novel}}$ & $\text{AP}_{50}^{\text{All}}$ \\
    \midrule
        Integrated Learning & \multirow{2}{*}{ViT-B/16~\cite{evaclip}} & \multirow{2}{*}{F-ViT~\cite{wu2023clipself}} & 53.6 & 36.5 & 49.1 \\
        DeCLIP & & & \textbf{56.7} & \textbf{43.3} & \textbf{53.2} \\
    \midrule
        Integrated Learning & \multirow{2}{*}{ViT-L/14} & \multirow{2}{*}{F-ViT} & 62.7 & 43.2 & 57.6 \\
        DeCLIP & & & \textbf{65.8} & \textbf{50.2} & \textbf{61.7} \\
    \bottomrule
    \end{tabular}
    \label{tab:decouple_detection}
\end{table}

\begin{figure*}[tbh]
    \centering
    \includegraphics[width=0.99\linewidth]{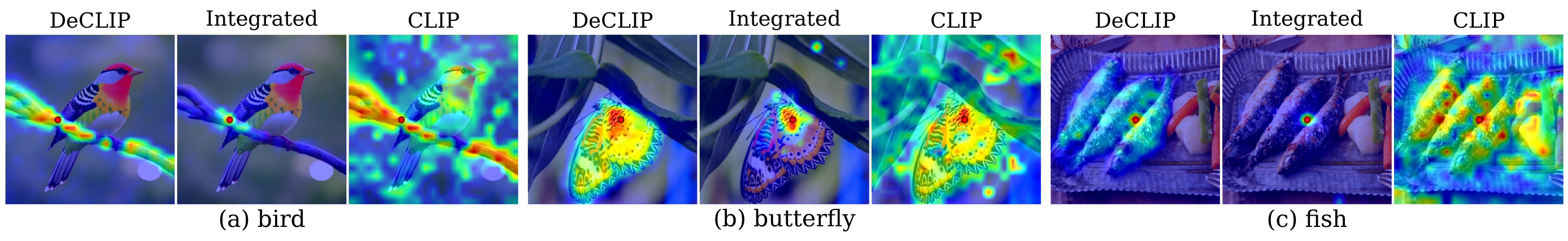}
    \caption{\revise{\textbf{Qualitative comparison of attention maps.} We compare DeCLIP, integrated learning, and CLIP on representative samples. DeCLIP produces more localized and object-consistent attention responses, while integrated learning retains more background activation and CLIP shows diffuse noisy responses.}}
    \label{fig:supp_attention_triplets}
\end{figure*}

\begin{table*}[tbh]
    \centering
    \caption{\textbf{Quantitative comparison of decoupled and integrated learning approaches for OV segmentation.} Results are reported in mIoU across multiple benchmarks. Decoupled learning consistently outperforms integrated learning on all datasets, leading to a substantial improvement in average mIoU.}
    \begin{tabular}{lccccccccc}
        \toprule
        \multirow{2}{*}{Method} & \multicolumn{3}{c}{\textit{With a background category}} & \multicolumn{5}{c}{\textit{Without background category}} & \multirow{2}{*}{Avg.} \\
        \cmidrule(lr){2-4} \cmidrule(lr){5-9}
        & VOC21 & Context60 & COCO-Obj & VOC20 & Cityscapes & Context59 & ADE & COCO-Stf & \\
        \midrule
        Integrated Learning & 58.5 &33.8 & 34.6 & 78.3 & 28.7 & 32.4 & 20.1 & 23.4  & 38.7 \\ 
        DeCLIP 
        & \textbf{64.1} & \textbf{37.9} & \textbf{38.7} & \textbf{85.3} & \textbf{35.7} & \textbf{41.6} & \textbf{23.1} & \textbf{26.8} & \textbf{44.1} \\
        \bottomrule
    \end{tabular}
    \label{tab:decouple_seg}
\end{table*}

\subsection{Attention and Feature Visualization}
\label{supp:subsec:feature_visualization}

\mypara{\revise{Attention Map Visualization.}} \revise{We compare CLIP, integrated learning, and DeCLIP at the attention-map level in Fig.~\ref{fig:supp_attention_triplets}. DeCLIP produces more localized and object-consistent responses than CLIP and integrated learning, supporting the benefit of content/context decoupling.}

\begin{figure*}[tbh]
    \centering
    \includegraphics[width=.86\linewidth]{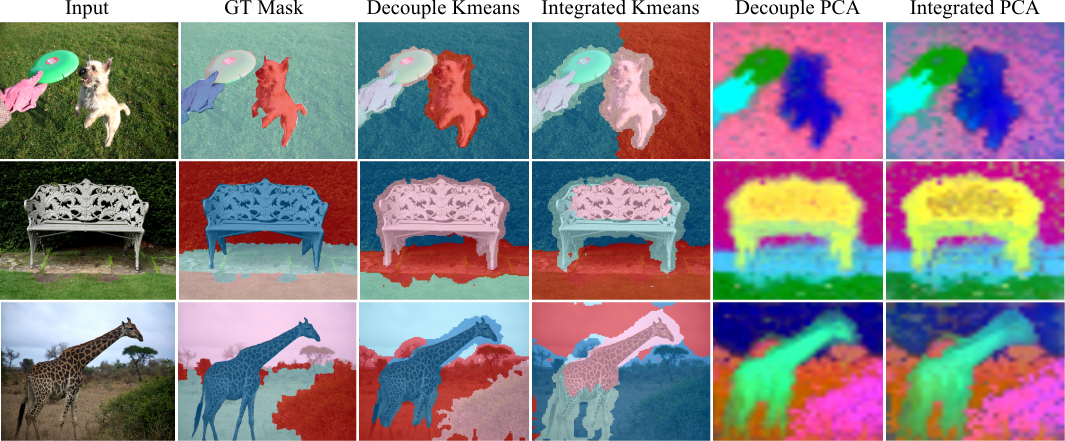}
    \caption{\textbf{Qualitative comparison of decoupled and integrated learning.}  From left to right are the input image, ground-truth mask, K-means clustering results, and PCA feature visualizations. Compared with integrated learning, the decoupled strategy produces cleaner segmentation masks with sharper object boundaries and closer alignment to ground truth. The PCA results further indicate that DeCLIP yields less noisy features than integrated learning.}
    \label{fig:feats_cluster_pca}
\end{figure*}

\mypara{\revise{Feature Visualization.}} \revise{We visualize model output features on the COCO-Panoptic dataset~\cite{coco_panoptic} using K-means clustering and Principal Component Analysis (PCA). As shown in Fig.~\ref{fig:feats_cluster_pca}, integrated learning produces noisy masks and less coherent PCA features, while DeCLIP aligns better with the ground truth and exhibits higher spatial consistency. These results show that decoupling yields cleaner and more discriminative dense representations.}

\begin{figure*}[tbh]
    \centering
    \includegraphics[width=0.82\linewidth]{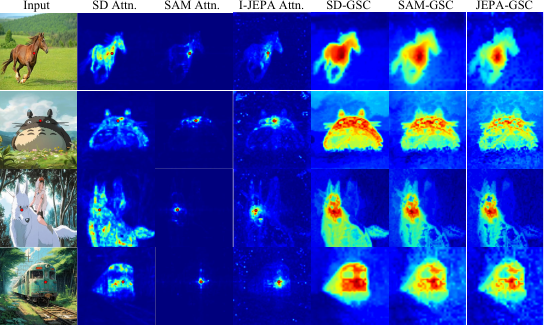}
   \caption{\textbf{Qualitative comparison of semantic completion strategies.} The red dot indicates the query token. From left to right, the columns display the input image, raw attention maps extracted from each model, and completed semantic affinity maps. In contrast to the boundary-focused SAM and fragmented I-JEPA, SD-GSC yields affinity maps with superior object integrity and spatial coherence.}
    \label{fig:gsc_vis}
\end{figure*}

\section{Analysis of Semantic Completion Sources}
\label{supp:sec:sd_gsc}
In this section, we provide a comprehensive evaluation of the proposed SD-GSC strategy against alternative semantic completion models.

\subsection{Ablation on Completion Sources}
\label{supp:subsec:sd_gsc_quan}
We conduct ablation studies to verify the effectiveness of the proposed SD-GSC strategy against alternative strategies. Specifically, we replace the Stable Diffusion model with SAM~\cite{sam} and I-JEPA~\cite{ijepa} to perform semantic completion. We select I-JEPA for the transformer-based feature inpainting ablation as it employs a Joint-Embedding Predictive Architecture to predict representations of masked target blocks conditioned on visible context, thereby serving as an inherent feature-level semantic inpainting baseline. Tab.~\ref{app:tab:sdgsc_seg} presents the quantitative results on OV semantic segmentation across eight benchmarks. Our SD-GSC method achieves the highest average mIoU of 44.2. This performance surpasses both the SAM-based variant (42.5) and the I-JEPA-based variant (42.2). Furthermore, to demonstrate the generalization ability of our strategy, we extend the evaluation to OV object detection in Tab.~\ref{app:tab:sdgsc_det}. Consistent with the segmentation results, SD-GSC outperforms the alternatives on base, novel, and overall AP metrics. Unlike discriminative or reconstruction-based approaches, SD leverages generative priors from its noise-to-image optimization process to encode global structural integrity, thereby facilitating superior semantic completion.

\begin{table*}[htb]
    \centering
    \caption{\textbf{Comparison of different semantic completion strategies.} We evaluate the effectiveness of our Stable Diffusion-based approach (SD-GSC) against alternative methods, including transformer-based feature inpainting using I-JEPA and boundary refinement using SAM.}
    \begin{tabular}{lccccccccc}
    \toprule
        \multirow{2.5}{*}{Method} & \multicolumn{3}{c}{\textit{With a background category}} & \multicolumn{5}{c}{\textit{Without background category}} & \multirow{2.5}{*}{Avg.} \\
        \cmidrule(lr){2-4} \cmidrule(lr){5-9}
        & VOC21 & Context60 & COCO-Obj & VOC20 & Cityscapes & Context59 & ADE & COCO-Stf & \\
    \midrule
        JEPA-GSC~\cite{ijepa} & 61.5 & 35.9 & 37.0 & 84.2 & 33.5 & 39.7 & 21.2 & 24.9 & 42.2 \\ 
        SAM-GSC~\cite{sam} & 61.7 & 36.1 & 37.2 & 83.9 & 34.1 & 39.9 & 21.9 & 25.1 & 42.5 \\ 
        SD-GSC & \textbf{64.1} & \textbf{37.9} & \textbf{38.7} & \textbf{85.3} & \textbf{35.7} & \textbf{41.6} & \textbf{23.1} & \textbf{26.8} & \textbf{44.2} \\
    \bottomrule
    \end{tabular}
    \label{app:tab:sdgsc_seg}
\end{table*}

\begin{table}[htb]
    \centering
    \caption{\textbf{Comparison of different semantic completion strategies on OV detection.} We report the Average Precision (AP) at IoU 0.5 for base categories, novel categories, and all categories on the OV-COCO benchmark.}
    \setlength{\tabcolsep}{3.5pt}
    \begin{tabular}{lccccc}
    \toprule
        Method & Backbone & Detector & $\text{AP}_{50}^{\text{Base}}$ & $\text{AP}_{50}^{\text{Novel}}$ & $\text{AP}_{50}^{\text{All}}$ \\
    \midrule
        JEPA-GSC~\cite{ijepa} &\multirow{3}{*}{ViT-B/16~\cite{evaclip}} & \multirow{3}{*}{F-ViT~\cite{wu2023clipself}} &54.3  &39.5  &50.4  \\
        SAM-GSC~\cite{sam} & & &54.2  &37.8  &49.9  \\
        SD-GSC & & & \textbf{56.7} & \textbf{43.3} & \textbf{53.2} \\
    \bottomrule
    \end{tabular}
    \label{app:tab:sdgsc_det}
\end{table}

\subsection{Visual Comparison of Semantic Completion}
\label{supp:subsec:sd_gsc_qual}
To further validate the efficacy of the proposed SD-GSC strategy, we have conducted qualitative comparisons against SAM-GSC and JEPA-GSC, as illustrated in Fig.~\ref{fig:gsc_vis}. The attention mechanism of SAM primarily focuses on discriminative boundaries, which leads to sparse affinity maps containing significant internal voids. Similarly, I-JEPA exhibits fragmented patterns stemming from its masked prediction objective, resulting in noise and insufficient spatial continuity within the completed maps. In contrast, Stable Diffusion utilizes generative priors to preserve global structural consistency by capturing long-range dependencies across the entire object region. Consequently, SD-GSC leverages these dense signals to generate complete and coherent semantic affinity maps. This structural advantage elucidates the quantitative superiority presented in Tab.~\ref{app:tab:sdgsc_seg} and Tab.~\ref{app:tab:sdgsc_det}.

\section{Robustness and Generalization Analysis}
\label{supp:sec:robustness}

This section provides a comprehensive evaluation of the robustness and generalization of DeCLIP under challenging real-world conditions. We systematically investigate robustness to degraded visual inputs, generalization under heavy occlusion, and cross-domain scenarios with significant domain shifts. 

\begin{table*}[tbh]
  \centering
    \caption{\textbf{Corruption Robustness for OV Segmentation.} Results are averaged over 5 severity levels for each of 15 corruption types and then aggregated over 8 segmentation benchmarks. \textbf{Clean}: average mIoU on clean images. For better readability, the results of the three noise corruptions (Gaussian/Shot/Impulse) and the four blur corruptions (Defocus/Glass/Motion/Zoom) are averaged and reported as \textbf{Noise (Avg.)} and \textbf{Blur (Avg.)}, respectively.}
  \label{tab:robust_seg1}
  \resizebox{\textwidth}{!}{%
    \setlength{\tabcolsep}{3pt} 
    \begin{tabular}{l c cc cc cc cc cc cc cc cc cc cc cc cc}
    \toprule
    \multirow{2.5}{*}{\textbf{Method}} & \textbf{Clean}
    & \multicolumn{2}{c}{\textbf{Noise (Avg.)}}
    & \multicolumn{2}{c}{\textbf{Blur (Avg.)}}
    & \multicolumn{2}{c}{\textbf{Snow}}
    & \multicolumn{2}{c}{\textbf{Frost}}
    & \multicolumn{2}{c}{\textbf{Fog}}
    & \multicolumn{2}{c}{\textbf{Brightness}}
    & \multicolumn{2}{c}{\textbf{Contrast}}
    & \multicolumn{2}{c}{\textbf{Elastic Trans.}}
    & \multicolumn{2}{c}{\textbf{Pixelate}}
    & \multicolumn{2}{c}{\textbf{JPEG Comp.}} \\
    \cmidrule(lr){3-4} \cmidrule(lr){5-6} \cmidrule(lr){7-8} \cmidrule(lr){9-10}
    \cmidrule(lr){11-12} \cmidrule(lr){13-14} \cmidrule(lr){15-16} \cmidrule(lr){17-18}
    \cmidrule(lr){19-20} \cmidrule(lr){21-22}
     & mIoU
     & mPC & rPC
     & mPC & rPC
     & mPC & rPC
     & mPC & rPC
     & mPC & rPC
     & mPC & rPC
     & mPC & rPC
     & mPC & rPC
     & mPC & rPC
     & mPC & rPC \\
    \midrule
    ClearCLIP~\cite{clearclip} & 37.5
    & 24.1 & 63.4
    & 26.1 & 68.5
    & 26.1 & 66.8
    & 25.7 & 66.1
    & 32.9 & 87.0
    & 35.6 & 93.9
    & 30.5 & 80.2
    & 27.6 & 73.7
    & 32.3 & 85.6
    & 31.2 & 82.4 \\
    
    SCLIP~\cite{sclip} & 38.2
    & 26.3 & 68.1
    & 28.9 & 75.0
    & 26.4 & 67.1
    & 26.3 & 67.3
    & 35.3 & 92.6
    & 36.9 & 97.3
    & 33.4 & 87.4
    & 28.6 & 76.3
    & 35.8 & 94.5
    & 33.3 & 87.2 \\
    
    DeCLIP & \textbf{44.2}
    & \textbf{33.6} & \textbf{75.8}
    & \textbf{34.8} & \textbf{77.7}
    & \textbf{33.4} & \textbf{73.2}
    & \textbf{33.2} & \textbf{72.8}
    & \textbf{41.3} & \textbf{93.2}
    & \textbf{43.3} & \textbf{97.8}
    & \textbf{40.4} & \textbf{91.0}
    & \textbf{36.7} & \textbf{83.2}
    & \textbf{41.9} & \textbf{94.9}
    & \textbf{40.4} & \textbf{90.9} \\
    \bottomrule
    \end{tabular}
  }
\end{table*}

\subsection{Robustness to Low-Quality Visual Inputs}
\label{supp:subsec:low_quality}

To evaluate robustness against degraded visual inputs, we follow the standard corruption benchmark protocol introduced in \cite{robust}, which includes 15 corruption types (e.g., noise, blur, weather, and compression artifacts), each evaluated at five severity levels, resulting in 75 corruption settings in total. Experiments are conducted on nine datasets spanning OV semantic segmentation and detection tasks, including VOC21, VOC20, Context60, Context59, COCO-Object, Cityscapes, ADE, COCO-Stuff, and OV-COCO. Performance is measured using mean Performance under Corruption (mPC) and relative Performance under Corruption (rPC), where rPC quantifies performance retention relative to clean inputs. We compare DeCLIP with strong CLIP-based baselines, including SCLIP, ClearCLIP, CLIP, and CLIPSelf. As summarized in Tab.~\ref{tab:robust_seg1} and Tab.~\ref{tab:robust-detection}, DeCLIP consistently achieves the highest mPC and rPC across nearly all corruption categories. Notably, compared with training-free OV segmentation methods such as SCLIP and ClearCLIP, DeCLIP exhibits significantly stronger robustness, indicating that decoupled distillation not only improves dense prediction accuracy but also enhances representation stability under visual degradation. For OV object detection on OV-COCO, results in Tab.~\ref{tab:robust-detection} further confirm this trend. DeCLIP substantially outperforms both vanilla CLIP and the distillation-based CLIPSelf baseline in terms of rPC, demonstrating superior resistance to severe image corruption in detection scenarios.

\begin{table}[htb]
    \centering
    \caption{\textbf{Corruption robustness for OV detection.}
    Experiments are conducted on OV-COCO.
    \textbf{Clean} is $\text{AP}_{50}$ on clean inputs.
    mPC and rPC are averaged over 15 corruption types and 5 severity levels.}
    
    \setlength{\tabcolsep}{2.4pt} 
    
    \begin{tabular}{lccccccccc}
    \toprule
        \multirow{2}{*}{Method} & 
        \multicolumn{3}{c}{$\text{AP}_{50}^{\text{Novel}}$} & 
        \multicolumn{3}{c}{$\text{AP}_{50}^{\text{Base}}$} & 
        \multicolumn{3}{c}{$\text{AP}_{50}^{\text{All}}$} \\
        
        \cmidrule(lr){2-4} \cmidrule(lr){5-7} \cmidrule(lr){8-10}
        
         & Clean & mPC & rPC & Clean & mPC & rPC & Clean & mPC & rPC \\
    \midrule
        CLIP~\cite{clip} & 26.7 & 13.8 & 51.8 & 44.0 & 26.2 & 59.6 & 39.5 & 23.0 & 58.2 \\ 
        CLIPSelf~\cite{wu2023clipself}  & 37.5 & 29.3 & 78.0 & 54.9 & 40.8 & 74.3 & 50.4 & 37.8 & 75.0 \\ 
        DeCLIP    & \textbf{43.7} & \textbf{35.3} & \textbf{80.7} & \textbf{56.7} & \textbf{44.6} & \textbf{78.7} & \textbf{53.3} & \textbf{41.7} & \textbf{78.2} \\
    \bottomrule
    \end{tabular}
    
    \label{tab:robust-detection}
\end{table}

\subsection{Performance under Occlusion Scenarios}
\label{supp:subsec:occlusion}
We further analyze robustness under occlusion using the OVIS benchmark~\cite{OVIS}, which explicitly categorizes instances into slightly, moderately, and heavily occluded subsets based on per-instance occlusion annotations ($\mathrm{AP}_{\mathrm{SO}}$, $\mathrm{AP}_{\mathrm{MO}}$, $\mathrm{AP}_{\mathrm{HO}}$). These levels correspond to instances with occlusion scores in the ranges $[0, 0.25]$, $(0.25, 0.5]$, and $(0.5, 0.75]$, respectively. The score derives from the average occlusion level of the most occluded 50\% of frames. Tab.~\ref{video_seg} in the main paper reports the overall AP to maintain conciseness and follow standard conventions~\cite{clip_vis}. Although this aggregate metric covers all difficulty levels, we provide the specific breakdown below to demonstrate robustness against heavy occlusion. Results in Tab.~\ref{tab:occuled_seg} demonstrate that replacing the ConvNeXt backbone with the proposed DeCLIP, while keeping other settings consistent with CLIP-VIS, significantly improves performance across all occlusion levels. Specifically, for the $\mathrm{AP}_{\mathrm{HO}}$ metric targeting heavy occlusion (occluded $\ge$ 50\% objects), DeCLIP outperforms CLIP-VIS by 2.3 and 5.9 points with ViT-B and ViT-L backbones, respectively. These results validate the effectiveness of DeCLIP in handling occluded scenarios.

\subsection{Cross-Domain Generalization}
\label{supp:subsec:cross_domain}
To further stress-test cross-domain transferability, we apply a COCO-distilled DeCLIP model directly to the remote sensing domain, following the SegEarth-OV~\cite{segearth} protocol. We evaluate on eight remote sensing datasets, including OpenEarthMap~\cite{openearthmap}, LoveDA~\cite{loveda}, iSAID~\cite{isaid}, Potsdam, Vaihingen\footnote{https://www.isprs.org/education/benchmarks/UrbanSemLab/semantic-labeling.aspx}, UAVid~\cite{uavid}, UDD5~\cite{udd5}, and VDD~\cite{vdd}. Results in Tab.~\ref{tab:segearth} show that DeCLIP achieves an average mIoU of 35.9, substantially outperforming CLIP (11.4) and surpassing strong OV segmentation methods such as ClearCLIP and GEM. Importantly, these gains are achieved without any exposure to remote sensing data during distillation, highlighting DeCLIP's strong cross-domain generalization capability.
\begin{table}[tb]
    \centering
    \caption{\textbf{Occlusion-level evaluation on OVIS for OV video instance segmentation.}
    We report overall $\mathrm{AP}$ and the breakdown by occlusion level, including $\mathrm{AP}_{\mathrm{SO}}$, $\mathrm{AP}_{\mathrm{MO}}$, and $\mathrm{AP}_{\mathrm{HO}}$. Higher is better.}
    \label{tab:ovis-occlude}
    \setlength{\tabcolsep}{4.5pt}
    \begin{tabular}{llcccc}
        \toprule
         Method & Backbone & $\mathrm{AP}$ & $\mathrm{AP}_{\mathrm{SO}}$ & $\mathrm{AP}_{\mathrm{MO}}$ & $\mathrm{AP}_{\mathrm{HO}}$ \\
         \midrule
         CLIP-VIS~\cite{clip_vis} & ResNet-50 & 14.1 & 24.6 & 14.1 & 3.1 \\
         \midrule
         CLIP-VIS & ConvNeXt-B & 18.5 & 31.8 & 20.3 & 4.4 \\
         CLIP-VIS+DeCLIP & ViT-B/16 & \textbf{22.2} & \textbf{37.2} & \textbf{26.3} & \textbf{6.7} \\
         \midrule
         CLIP-VIS & ConvNeXt-L & 20.0 & 36.6 & 22.5 & 5.5 \\
         CLIP-VIS+DeCLIP & ViT-L/14 & \textbf{29.3} & \textbf{47.0} & \textbf{33.9} & \textbf{11.4} \\
         \bottomrule
    \end{tabular}
    \label{tab:occuled_seg}
\end{table}

\begin{table*}[tbh]
    \centering
    \caption{\textbf{Cross-domain generalization to remote sensing applications.} We evaluated COCO-distilled DeCLIP on eight remote sensing segmentation datasets and report mIoU. Without exposure to any remote sensing images during decoupled distillation, DeCLIP achieved the best average performance, demonstrating robust transfer across a large domain gap.}
    \label{tab:segearth}
    \begin{tabular}{lccccccccc}
    \toprule
        Method & OpenEarthMap & LoveDA & iSAID & Potsdam & Vaihingen & UAVid & UDD5 & VDD & \textbf{Avg.} \\
        \midrule
        CLIP~\cite{clip} & 12.0  & 12.4  & 7.5  & 14.5  & 10.3  & 10.9  & 9.5  & 14.2  & 11.4  \\ 
        MaskCLIP~\cite{maskclip} & 25.1  & 27.8  & 14.5  & 31.7  & 24.7  & 28.6  & 32.4  & 32.9  & 27.2  \\ 
        SCLIP~\cite{sclip} & 29.3  & 30.4  & 16.1  & 36.6  & 28.4  & 31.4  & 38.7  & 37.9  & 31.1  \\ 
        GEM~\cite{gem} & 33.9  & 31.6  & 17.7  & 36.5  & 24.7  & 33.4  & 41.2  & 39.5  & 32.3  \\ 
        ClearCLIP~\cite{clearclip} & 31.0  & 32.4  & 18.2  & 40.9  & 27.3  & \textbf{36.2}  & 41.8  & 39.3  & 33.4  \\ 
        DeCLIP & \textbf{34.2}  & \textbf{38.5}  & \textbf{19.0}  & \textbf{41.4}  & \textbf{34.6}  & 35.4  & \textbf{44.1}  & \textbf{39.9}  & \textbf{35.9} \\
        \bottomrule
    \end{tabular}
\end{table*}

\section{\rev{Efficiency Analysis}}
\label{supp:sec:efficiency}
DeCLIP is a representation learning framework that strengthens CLIP dense features without changing the model design. The gains come from decoupled distillation during training, so the inference network remains identical to the original CLIP backbone. Consequently, DeCLIP introduces no extra parameters, FLOPs, or memory overhead over vanilla CLIP. We evaluate optimization compatibility, training and inference costs, backbone generality, and real-time use.

\subsection{Inference Cost Analysis}
\label{supp:subsec:inference_efficiency}
To quantitatively verify the cost-free property, we benchmark DeCLIP against vanilla CLIP on a single NVIDIA RTX 4090 GPU. The results are summarized in Tab.~\ref{tab:4090_speed}. Experiments are conducted with a batch size of 1 and an input resolution of $336 \times 336$. We report forward inference latency, parameter count, memory footprint, FLOPs, and MACs for both ViT-B/16 and ViT-L/14 backbones. The results show that DeCLIP matches or slightly improves inference efficiency compared to CLIP. Specifically, the forward inference time is reduced from 11.79 ms to 11.34 ms for ViT-B/16 and from 22.62 ms to 21.81 ms for ViT-L/14. The parameter count remains unchanged (86.26M for ViT-B/16 and 304.43M for ViT-L/14), confirming that DeCLIP introduces no extra learnable parameters. The slight reduction in inference latency and FLOPs is attributed to the removal of the MLP layer and residual connection in the final Transformer block, following the same simplification strategy adopted in ClearCLIP~\cite{clearclip}. Consequently, DeCLIP enhances dense perception without computational overhead, rendering it ideal for resource-constrained applications. This confirms that the performance gains stem solely from improved representation learning rather than architectural complexity.

\begin{table}[tb]
    \centering
    \caption{\textbf{Inference efficiency comparison of CLIP and DeCLIP.}
    DeCLIP preserves the same inference-time architecture and parameter count as the original CLIP, resulting in comparable latency, memory footprint, and computational complexity across both backbones.}
    \label{tab:4090_speed}
    \resizebox{\columnwidth}{!}{%
    \begin{tabular}{lccccc}
    \toprule
    Method & Size & Fwd. Time (ms) & Memory (MB) & FLOPs (G) & MACs (G) \\
    \midrule
    CLIP~\cite{clip} & ViT-B/16 & 11.79  & 658.45  & 57.18  & 28.59  \\ 
    CLIP & ViT-L/14 & 22.62  & 1842.28  & 262.92  & 131.46  \\ 
    \midrule
    DeCLIP & ViT-B/16 & 11.34  & 658.45  & 53.01  & 26.50  \\ 
    DeCLIP & ViT-L/14 & 21.81  & 1847.39  & 253.24  & 126.62 \\
    \bottomrule
    \end{tabular}%
    }
\end{table}

\subsection{Inference Efficiency in Downstream Tasks}
\label{supp:subsec:realtime-vis}
Building upon the intrinsic efficiency analysis in Sec.~\ref{supp:subsec:inference_efficiency}, we further investigate the generalizability of the cost-free property to downstream tasks. We utilize the OV video instance segmentation task to validate the efficiency of DeCLIP when integrated into downstream frameworks. As detailed in Sec.~\ref{OV_Video_Instance_Segmentation}, DeCLIP functions as a direct drop-in replacement for the feature extractor within the CLIP-VIS framework. Given that DeCLIP preserves the architectural topology of CLIP through distillation-based weight updates, this integration incurs no additional parametric or computational overhead. Quantitative results in Tab.~\ref{tab:video_fps} indicate that our method maintains a frame rate of 15.95 FPS on 512$\times$512 inputs. Notably, the inference latency exhibits parity with both CLIP and ConvNeXt baselines. These findings substantiate that DeCLIP enhances representation quality without compromising system-level latency, and the overall throughput remains determined by the downstream method efficiency.

\subsection{Generalization to Lightweight CLIP}
\label{supp:subsec:lightweight_backbone}

To evaluate the applicability of DeCLIP under computational and memory constraints, we further investigate its generalization capability on lightweight CLIP backbones. We adopt TinyCLIP-ViT-39M/16~\cite{tinyclip} as a representative compact model, which reduces the parameter count by approximately 50\% compared to the standard ViT-B/16. We apply DeCLIP to this backbone using our decoupled distillation strategy without architectural modifications. For comparison, we implement state-of-the-art training-free methods, SCLIP and ClearCLIP, on the same TinyCLIP architecture. As shown in Tab.~\ref{tab:tinyclip}, DeCLIP-Tiny consistently outperforms the competing methods across all eight benchmarks. Specifically, our method achieves an average mIoU of 39.6, significantly surpassing ClearCLIP-Tiny (35.6) and SCLIP-Tiny (19.1). These results demonstrate that our decoupled learning framework remains highly effective even with severely limited model capacity, validating its potential for OV perception in edge deployment.

\begin{table}[tb]
\centering
\setlength{\tabcolsep}{2.3pt} 
\caption{\textbf{Inference speed analysis.} Inference time is decomposed into preprocessing, backbone, segmentation head, and post-processing phases in milliseconds. Tests use an RTX 4090 (512×512 input, batch size 1).}
\label{tab:video_fps}
\begin{tabular}{lcccccc}
\toprule
\multirow{2.5}{*}{\textbf{Method}} & \multirow{2.5}{*}{\textbf{Backbone}} & \multirow{2.5}{*}{\textbf{FPS}} & \multicolumn{4}{c}{\textbf{Latency breakdown (ms)}} \\
\cmidrule(lr){4-7} 
 & & & Pre. & Backbone & Seg Head & Post. \\
\midrule
CLIP-VIS~\cite{clip_vis} & ConvNeXt-B & 16.24 & 0.54 & 12.37 & 22.43 & 26.21 \\
CLIP-VIS+CLIP& ViT-B/16 &15.84 & 0.56 & 14.88 & 22.08 & 25.58 \\
CLIP-VIS+DeCLIP & ViT-B/16 & 15.95 & 0.53 & 14.49 & 22.11 & 25.54 \\
\bottomrule
\end{tabular}
\end{table}

\begin{table*}[tb]
    \centering
    \caption{\textbf{Generalization analysis on the lightweight TinyCLIP-ViT-39M/16 backbone.} We implement state-of-the-art open-vocabulary segmentation methods on the capacity-constrained TinyCLIP model. DeCLIP consistently outperforms both SCLIP and ClearCLIP across eight diverse benchmarks, demonstrating the robustness of our decoupled learning approach in edge-deployment scenarios.}
    \label{tab:tinyclip}
    \begin{tabular}{lccccccccc}
        \toprule
        \multirow{2}{*}{Method} & \multicolumn{3}{c}{\textit{With a background category}} & \multicolumn{5}{c}{\textit{Without background category}} & \multirow{2}{*}{Avg.} \\
        \cmidrule(lr){2-4} \cmidrule(lr){5-9}
        & VOC21 & Context60 & COCO-Object & VOC20 & Cityscapes & Context59 & ADE & COCO-Stuff & \\
        \midrule
        SCLIP-Tiny~\cite{sclip} & 32.6 & 14.3 & 8.0 & 59.8 &  3.9 & 19.0 & 5.7 & 9.8  & 19.1 \\ 
        ClearCLIP-Tiny~\cite{clearclip} & 54.2 & 29.8 & 28.9 & 79.8 & 20.3 & 32.5 & 16.7 & 22.3 & 35.6 \\ 
        DeCLIP-Tiny & \textbf{59.3} & \textbf{32.2} & \textbf{32.2} & \textbf{82.8} & \textbf{31.9} & \textbf{35.8} & \textbf{18.2} & \textbf{24.1} & \textbf{39.6} \\
        \bottomrule
    \end{tabular}
\end{table*}

\subsection{Orthogonality to Inference Acceleration}
\label{supp:subsec:inference_optimization}

\begin{table*}[htb]
    \centering
    \caption{\textbf{Quantitative results of inference optimization.} Experiments on an NVIDIA H100 GPU (batch size 1, $560 \times 560$) demonstrate that DeCLIP is orthogonal to standard acceleration strategies. Existing optimization frameworks like TensorRT can significantly boost the inference speed of our method (e.g., reducing latency by 86\% with FP16).}
    \label{tab:inference_speed}
    \setlength{\tabcolsep}{5pt}
    \begin{tabular}{lccccccc}
        \toprule
        \textbf{Model} & \textbf{Backend} & \textbf{Precision} & \textbf{Latency (ms)} $\downarrow$ & \textbf{Throughput (FPS)} $\uparrow$ & \textbf{Memory (MB)} $\downarrow$ & \textbf{VOC21 (mIoU)}$\uparrow$  \\
        \midrule
        \multirow{4}{*}{CLIP~\cite{evaclip}} 
        & PyTorch & FP32 & 10.29 & 97.19 & 1397 & 43.4  \\
        & PyTorch & FP16 & 8.41 & 118.87 & 725 & 43.4  \\
        & TensorRT & FP16 & 1.46 & 683.05 & 679 & 43.4  \\
        & TensorRT & FP8 & 1.41 & 711.73 & 531 & 43.9  \\
        \midrule
        \multirow{4}{*}{DeCLIP} 
        & PyTorch & FP32 & 9.84 & 101.67 & 1397 & 64.1  \\
        & PyTorch & FP16 & 8.01 & 124.79 & 725 & 64.1  \\
        & TensorRT & FP16 & 1.37 & 730.63 & 679 & 64.1  \\
        & TensorRT & FP8 & 1.35 & 742.33 & 531 & 63.9  \\
        \bottomrule
    \end{tabular}
\end{table*}

Complementing the analysis of model capacity, we further investigate the orthogonality of DeCLIP with respect to standard inference optimization techniques. Specifically, we employ NVIDIA TensorRT\footnote{https://github.com/NVIDIA/TensorRT} to benchmark the inference latency and memory footprint across varying numerical precisions, including FP32, FP16, and FP8. We compare DeCLIP against the standard CLIP baseline, which utilizes the MaskCLIP~\cite{maskclip} strategy for dense prediction. As presented in Tab.~\ref{tab:inference_speed}, DeCLIP demonstrates acceleration characteristics consistent with the original CLIP architecture, confirming that our representation learning framework is structurally orthogonal to inference optimization techniques. Notably, DeCLIP maintains superior segmentation performance compared to the baseline while achieving significant efficiency gains through optimization. Furthermore, the method exhibits high numerical stability, preserving accuracy even under aggressive FP8 quantization. These results indicate that DeCLIP possesses the requisite robustness and efficiency for real-time OV perception in resource-constrained scenarios.

\subsection{\rev{Training Cost Analysis}}
\label{supp:subsec:training_cost}
\rev{Tab.~\ref{tab:training_cost} summarizes fine-tuning costs for CLIPSelf-B and DeCLIP with ViT-B and ViT-L, including training time, memory usage, and teacher computation. Compared with CLIPSelf-B, DeCLIP-B requires 3.6\% more training time and 1.9 GiB more memory, while substantially improving dense perception performance.}

\begin{table}[tb]
    \centering
    \caption{\rev{\textbf{Fine-tuning costs on eight NVIDIA A800 GPUs (batch size 8 per GPU).} Memory denotes peak allocation per GPU. SD incurs no computational overhead during fine-tuning, as its attention maps can be cached beforehand.}}
    \label{tab:training_cost}
    \setlength{\tabcolsep}{3pt}
    \resizebox{\columnwidth}{!}{%
    \begin{tabular}{lccccc}
    \toprule
    \multirow{2}{*}{Method} & Time & Memory & \multicolumn{3}{c}{Teacher computation (ms/step)} \\
    \cmidrule(lr){4-6}
    & (min/epoch) & (GiB/GPU) & CLIP & DINOv2 & Total \\
    \midrule
    CLIPSelf-B & 8.3 & 8.4 & 74.9 & -- & 74.9 \\
    DeCLIP-B & 8.6 & 10.3 & 74.5 & 16.7 & 91.2 \\
    DeCLIP-L & 37.0 & 33.5 & 574.1 & 22.8 & 596.9 \\
    \bottomrule
    \end{tabular}%
    }
\end{table}

\section{\revise{Additional Ablation Analysis}}
\label{supp:sec:ablation}

\begin{figure*}[tb]
    \centering
    \includegraphics[width=0.99\linewidth]{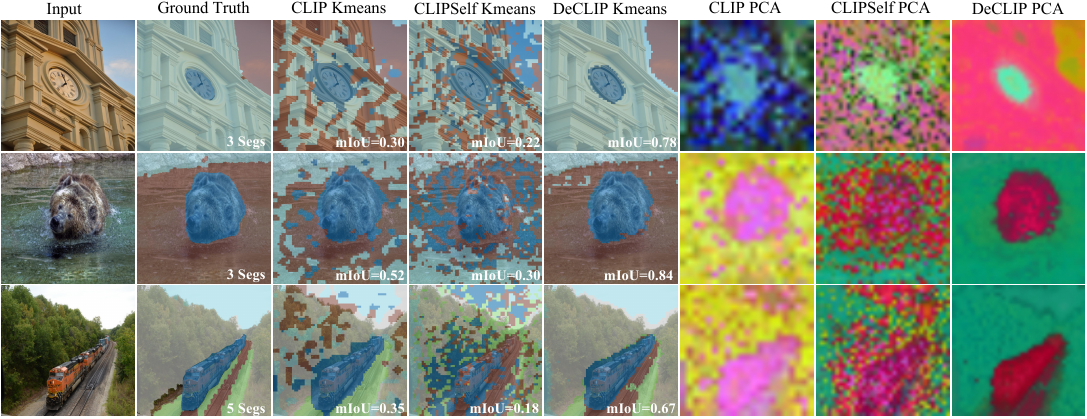}
   \caption{\textbf{Qualitative comparison of feature representations using K-means and PCA.} Columns 3 through 5 display token-level segmentation masks generated via K-means clustering. Columns 6 through 8 show the principal components of the features. Region-level alignment causes CLIPSelf to exhibit more severe feature fragmentation than the original CLIP. Conversely, DeCLIP generates integral object masks with robust activations that align closely with the ground truth.}
    \label{fig:rcc_vis}
\end{figure*}

\begin{figure*}[tb]
    \centering
    \includegraphics[width=0.95\linewidth]{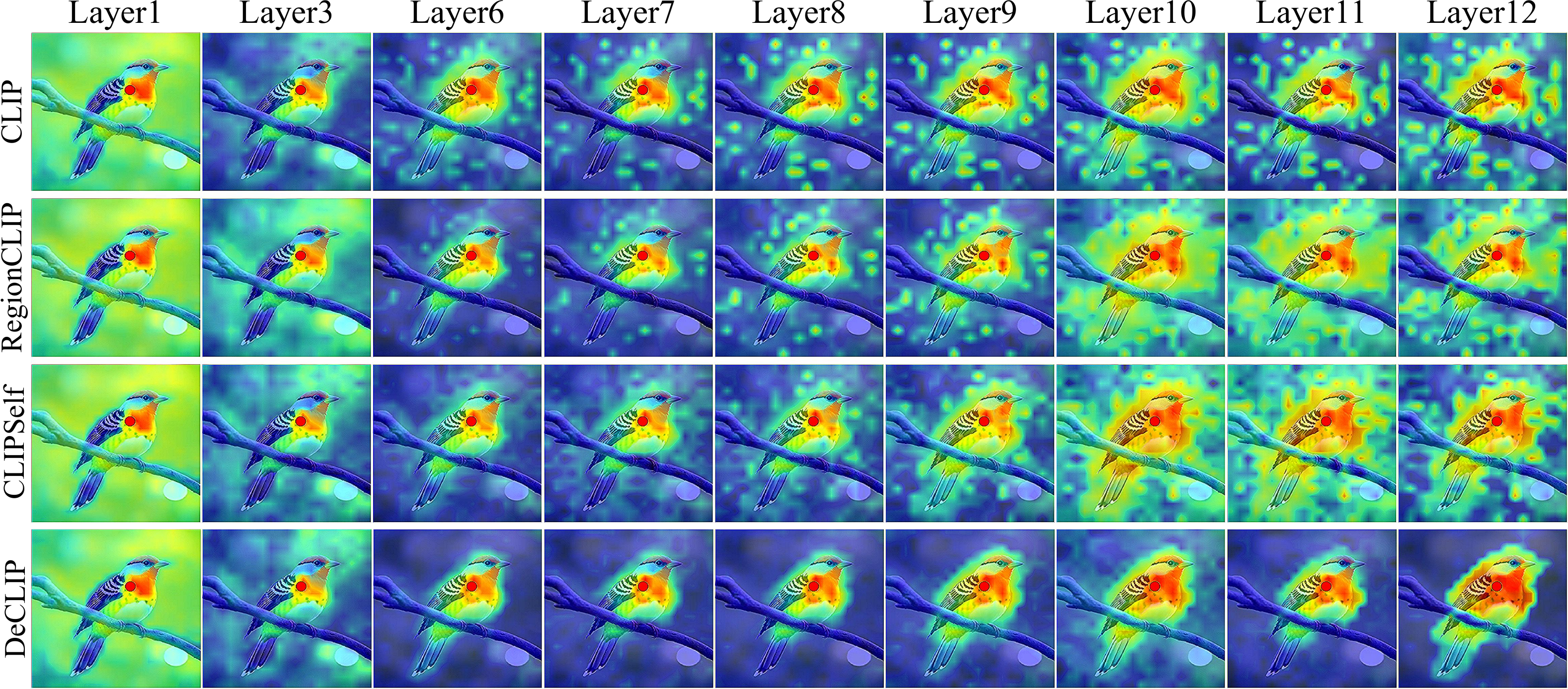}
   \caption{\textbf{Qualitative comparison of feature correlations between DeCLIP and existing methods~\cite{wu2023clipself,regionclip}.}  Specifically, the input image resolution is 336*336. We extract the output features from each attention block of CLIP, where each feature $\mathbf{F} \in \mathbb{R}^{441 \times D}$. Then, we compute the feature correlations $\mathbf{FC} \in \mathbb{R}^{441 \times 441}$ between the image tokens within $\mathbf{F}$ using cosine similarity. We randomly select a specific image token's feature correlation (indicated by the red dots) and upsample it to a resolution of 336*336 for visualization.} 
    \label{fig:rcc_vis2}
\end{figure*}

\subsection{Qualitative Analysis of RCC}
\label{supp:sec:rcc_vis}
This section presents a visualization analysis to substantiate the motivation of the proposed Region Correlation Constraint module. Region-based vision-language methods~\cite{regionclip, wu2023clipself} facilitate OV object detection by aligning region features from the CLIP visual encoder with corresponding region [CLS] tokens or text embeddings. However, empirical observations indicate that such region-level fine-tuning degrades dense token-level representations. Specifically, this process disrupts the intrinsic correlations among tokens within local regions.

\mypara{Analysis of Semantic Structure.} We analyze the impact of region-level fine-tuning on the global semantic structure via K-means clustering and PCA. Fig.~\ref{fig:rcc_vis} illustrates that these methods compromise intra-object coherence to prioritize region-text alignment. This process leads to fragmented semantic consistency within objects and results in noisy, discontinuous segmentation maps. Conversely, DeCLIP equipped with the RCC effectively preserves local spatial correlations. The K-means results demonstrate that our approach coherently aggregates tokens within semantic regions. Furthermore, PCA visualizations confirm that the RCC module maintains a distinct separation between foreground objects and the background.

\mypara{Feature Correlations Visualizations.} To complement the structural analysis, we investigate layer-wise feature correlations to examine local token affinities. Fig.~\ref{fig:rcc_vis2} compares CLIP, DeCLIP, and existing pre-finetuning methods by computing the cosine similarity between a specific anchor token within the primary object and all other image tokens. As shown in the first row, CLIP exhibits significant spurious correlations in the background, which undermines its dense perception capability. Existing region-based methods (Rows 2 and 3) exacerbate this issue, where high correlation scores diffuse extensively into the background and result in severe boundary leakage. In contrast, DeCLIP incorporates the RCC module to mitigate this degradation by enforcing local consistency. Consequently, our method effectively suppresses background noise and strictly confines high correlations within the semantic boundaries of the object. This validates the effectiveness of the RCC in preserving pixel-level precision during region alignment.

\subsection{Sanity Check on VLM Baseline}
\label{supp:subsec:sanity}

We conduct sanity checks to verify whether the performance improvement of DeCLIP in dense prediction stems from using EVA-CLIP. First, we use vanilla EVA-CLIP as the backbone in CAT-Seg~\cite{catseg} and compare its performance with DeCLIP on the OV semantic segmentation task (Tab.~\ref{sanity_check_ovss}). Next, we re-implement ClearCLIP\cite{clearclip} with EVA-CLIP (termed EVA-ClearCLIP) and compare EVA-CLIP, EVA-ClearCLIP, and DeCLIP in the training-free OV semantic segmentation task (Tab.~\ref{sanity_check_tfovss}). For detailed experimental settings and metric definitions of these two tasks, please refer to Sec.~\ref{sec:ov_seg} and Sec.~\ref{sec:training_free_ov_seg} in the main paper, respectively. Experimental results demonstrate that EVA-CLIP does not introduce any observable biases.

\begin{table}[tb]
    \centering
    \caption{Sanity check on OV semantic segmentation to eliminate potential bias introduced by EVA-CLIP. The evaluation metric is mIoU.}
    \setlength{\tabcolsep}{1.2pt}
    \begin{tabular}{llcccccc}
    \toprule
    Method & Size & ADE847 & Context459 & ADE150 & Context59 & VOC20 & VOC21\\
    \midrule
    CLIP & ViT-B & 12.0 & 19.0 & 31.8 & 57.5 & 94.6 & 77.3 \\
    CLIP & ViT-L & 16.0 & 23.8 & 37.9 & 63.3 & 97.0 & 82.5 \\    
    \midrule
    EVA-CLIP & ViT-B & 11.9 & 17.6 & 30.4 & 52.3 & 94.2 & 74.2 \\
    EVA-CLIP & ViT-L & 14.2 & 21.3 & 34.8 & 56.2 & 95.8 & 80.1 \\    
    \midrule
    DeCLIP & ViT-B & 15.3 & 21.4 & 36.3 & 60.6 & 96.6 & 81.3 \\
    DeCLIP & ViT-L & \textbf{17.6} & \textbf{25.9} & \textbf{40.7} & \textbf{63.9} & \textbf{97.7} & \textbf{83.9} \\
    \bottomrule
    \end{tabular}
\label{sanity_check_ovss}
\end{table}

\begin{table}[tb]
    \setlength{\tabcolsep}{1.5pt}
    \centering
    \caption{Sanity checks on training-free OV semantic segmentation to eliminate potential bias introduced by EVA-CLIP. The evaluation metric is mIoU.}
    \begin{tabular}{lcccccc}
        \toprule
        Method & VOC21 & Context60 & COCO-Obj & Cityscapes & ADE & COCO-Stf \\
        \midrule
        CLIP \cite{clip} & 18.8 & 9.9  & 8.1  & 6.5  & 3.1  & 5.7  \\
        EVA-CLIP  \cite{evaclip} & 23.4  & 12.8  & 15.3   & 12.8  & 7.7  & 9.7  \\
        \midrule
        ClearCLIP \cite{clearclip} & 51.8  & 32.6  & 33.0 & 30.0  & 16.7  & 23.9  \\
        EVA-ClearCLIP  & 47.0  & 29.7  & 30.2 & 26.3  & 16.7  & 20.4  \\
        \midrule
        DeCLIP & \textbf{64.1}  & \textbf{37.9}  & \textbf{38.7}  & \textbf{35.7}  & \textbf{23.1} & \textbf{26.8} \\
        \bottomrule
    \end{tabular}
\label{sanity_check_tfovss}
\end{table}

\begin{figure*}[tb]
    \centering
    \includegraphics[width=0.95\linewidth]{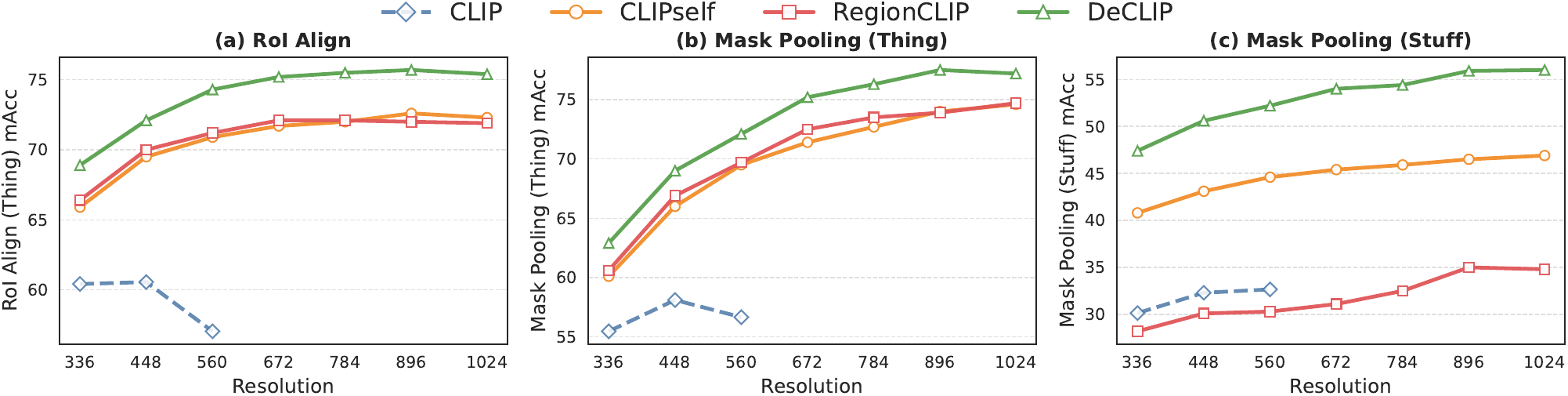}
    \caption{\textbf{Quantitative comparison of vision-language alignment across various resolutions.} Classification accuracy on COCO-Panoptic for \textbf{(a)} ``thing" categories using Ground Truth (GT) boxes, \textbf{(b)} ``thing" categories using GT masks, and \textbf{(c)} ``stuff" categories using GT masks. CLIP results are limited to 560 resolution due to the absence of high-resolution training.}
    \label{fig:region_cls}
\end{figure*}

\subsection{Vision-Language Alignment Analysis}
\label{supp:subsec:vl-align-analy}

Although CLIP demonstrates strong alignment capabilities with the [CLS] token via text-image contrastive learning, its dense features lack fine-grained vision-language alignment. DeCLIP is proposed specifically to address the fine-grained vision-language misalignment inherent in CLIP. To quantify the vision-language alignment of region features, we conducted OV region classification using the COCO Panoptic~\cite{coco_panoptic} dataset. This benchmark encompasses 133 semantic categories divided into two distinct groups. The ``thing" group consists of 80 classes of countable objects with defined geometry. The ``stuff" group comprises 53 classes of amorphous background materials or textures. Fig.~\ref{fig:region_cls} illustrates the Top-1 mean accuracy (mAcc) for OV region classification across various input resolutions. We observe that the classification performance generally improves as the resolution increases. However, the vanilla CLIP model exhibits performance degradation at higher resolutions attributed to a lack of high-resolution training. In contrast, DeCLIP consistently outperforms both the baseline and existing fine-tuning methods, such as RegionCLIP~\cite{regionclip} and CLIPSelf~\cite{wu2023clipself}, across all tested resolutions. These results demonstrate that the proposed decoupled learning strategy significantly enhances the fine-grained vision-language alignment of dense features.

\subsection{\revise{Object-Scale Detection Analysis}}
\label{supp:subsec:object_scale_detection}
\revise{To quantify the behavior on small objects, we evaluate OV-COCO detection results by object scale. As shown in Tab.~\ref{tab:object_scale_detection}, DeCLIP improves AP for small, medium, and large objects compared with CLIP and CLIPSelf.}

\begin{table}[tb]
    \centering
    \setlength{\tabcolsep}{3.5pt}
    \caption{\revise{\textbf{Object-scale detection analysis on OV-COCO.} We report novel $\mathrm{AP}_{50}$, bbox mAP, and AP for small, medium, and large objects.}}
    \begin{tabular}{lccccc}
    \toprule
        Method & Novel $\mathrm{AP}_{50}$ & BBox mAP & $\mathrm{AP}_{s}$ & $\mathrm{AP}_{m}$ & $\mathrm{AP}_{l}$ \\
    \midrule
        CLIP~\cite{evaclip} & 30.0 & 25.5 & 15.6 & 28.2 & 35.6 \\
        CLIPSelf~\cite{wu2023clipself} & 44.5 & 33.1 & 22.8 & 37.2 & 42.2 \\
        DeCLIP & \textbf{50.3} & \textbf{34.9} & \textbf{25.5} & \textbf{38.3} & \textbf{44.6} \\
    \bottomrule
    \end{tabular}
    \label{tab:object_scale_detection}
\end{table}

\subsection{\revise{Crop Sampling Range Sensitivity}}
\label{supp:subsec:crop_sampling_sensitivity}
\revise{To assess sensitivity to the number of sampled crops, we vary the sampling range of $m$ and $n$ in the content feature distillation branch while keeping other settings unchanged. Smaller ranges produce fewer and coarser crops, while larger ranges produce more and finer crops. As shown in Tab.~\ref{tab:crop_sampling_sensitivity}, DeCLIP remains stable across different crop ranges, with the average mIoU varying within 0.5.}

\begin{figure*}[t]
    \centering
    \includegraphics[width=0.97\linewidth]{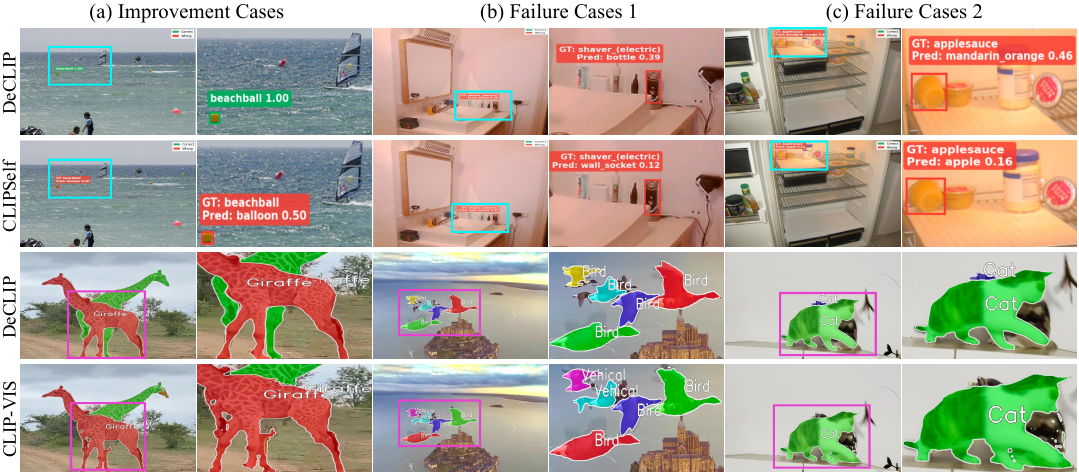}
   \caption{\textbf{Qualitative comparison of improvements and failure cases.} Rows 1-2 show OV detection on OV-LVIS, while Rows 3-4 display video segmentation on OVIS. Even columns present zoomed-in views of the predicted regions. \textbf{(a) Improvement Cases:} DeCLIP corrects baseline classification errors for small objects and refines segmentation boundaries for large, overlapping instances. \textbf{(b)-(c) Failure Cases:} The model still faces difficulties in detecting tiny objects from rare categories and segmenting instances under severe occlusion. }
    \label{fig:fail_case}
\end{figure*}

\begin{table*}[tb]
    \centering
    \setlength{\tabcolsep}{5pt}
    \caption{\revise{\textbf{Sensitivity to the crop sampling range.} Results are reported as mIoU on eight training-free OV semantic segmentation benchmarks.}}
    \begin{tabular}{lccccccccc}
    \toprule
        Crop range & VOC20 & VOC21 & Context59 & Context60 & COCO-Obj & COCO-Stuff & ADE & Cityscapes & Avg. \\
    \midrule
        $m,n \in [1,3]$ & 86.4 & 64.0 & 41.5 & 37.8 & 39.7 & 26.7 & 23.0 & 33.4 & 44.1 \\
        $m,n \in [1,6]$ & 85.3 & \textbf{65.1} & 42.2 & \textbf{38.5} & 39.0 & \textbf{27.1} & 23.2 & 35.9 & \textbf{44.5} \\
        $m,n \in [1,8]$ & 84.6 & 64.7 & \textbf{42.2} & 38.4 & 37.6 & 27.0 & \textbf{23.2} & \textbf{37.2} & 44.4 \\
    \bottomrule
    \end{tabular}
    \label{tab:crop_sampling_sensitivity}
\end{table*}

\subsection{\revise{Layer and Timestep Analysis}}
\label{supp:subsec:sd_layer_timestep}
\revise{We further analyze the Stable Diffusion teacher signal by varying the attention layer and diffusion timestep used in SD-GSC. We use t45\_up3 as the default setting and evaluate the resulting models under the same training-free OVSS protocol. For a fair comparison, all variants were retrained under the same setup using re-extracted Stable Diffusion attention caches, yielding a slightly higher default result than reported in the main table. As shown in Tab.~\ref{tab:sd_layer_timestep}, changing the attention layer causes a clear performance drop, while changing the timestep within the late diffusion stage leads to only minor variation. These results indicate that the high-resolution decoder layer is critical for SD-GSC, while the late diffusion timestep range is relatively stable.}

\begin{table*}[tb]
    \centering
    \setlength{\tabcolsep}{3pt}
    \caption{\revise{\textbf{Stable Diffusion layer and timestep ablation for SD-GSC on training-free OVSS.} We report mIoU on eight training-free OVSS benchmarks. The CLIP row is included as a reference baseline.}}
    \begin{tabular}{llccccccccc}
        \toprule
        \multirow{2}{*}{Model} & \multirow{2}{*}{SD setting} & \multicolumn{3}{c}{\textit{With a background category}} & \multicolumn{5}{c}{\textit{Without background category}} & \multirow{2}{*}{Avg.} \\
        \cmidrule(lr){3-5} \cmidrule(lr){6-10}
        & & VOC21 & Context60 & COCO-Obj & VOC20 & Cityscapes & Context59 & ADE & COCO-Stf & \\
        \midrule
        CLIP & -- & 18.8 & 9.9 & 8.1 & 49.4 & 6.5 & 11.1 & 3.1 & 5.7 & 14.1 \\
        \midrule
        DeCLIP & t45\_up3 (default) & \textbf{65.1} & \textbf{38.5} & \textbf{39.0} & \textbf{85.3} & 35.9 & \textbf{42.2} & \textbf{23.2} & \textbf{27.1} & \textbf{44.5} \\
        DeCLIP & t45\_up2 & 55.4 & 32.3 & 28.3 & 80.3 & 30.4 & 35.4 & 18.1 & 22.0 & 37.8 \\
        DeCLIP & t30\_up3 & 65.0 & 38.5 & 38.7 & 85.2 & \textbf{36.0} & 41.9 & 23.0 & 27.0 & 44.4 \\
        \bottomrule
    \end{tabular}
    \label{tab:sd_layer_timestep}
\end{table*}

\section{Evaluation Details and Failure Cases}
\label{supp:sec:failure_case}
\subsection{Spatiotemporal IoU for Video Consistency}
\label{supp:subsec:video_consistency}
To evaluate the temporal stability of the segmentation results, we adopt the standard mean AP based on the Spatiotemporal IoU metric (Tab.~\ref{video_seg} in the main paper). This metric serves as the standard evaluation protocol for the YTVIS, OVIS, and BURST benchmarks. The Spatiotemporal IoU metric inherently penalizes long-term drift. It is calculated across the entire video sequence of length $T$ as follows:
\begin{equation}
IoU(i, j) = \frac{\sum_{t=1}^T |m^i_t \cap\hat m^j_t|}{\sum_{t=1}^T |m^i_t \cup\hat m^j_t|},
\end{equation}
where $m^i_t$ and $\hat m^j_t$ denote the ground truth and predicted masks for object $i$ and hypothesis $j$ at frame $t$. This formulation requires consistent identification across the full duration. Any identity switch or segmentation drift significantly reduces the numerator and leads to a drop in the final score. Therefore, a high AP score under this metric serves as a direct proxy for temporal consistency.

\subsection{Data Preprocessing for 3D Tasks}
\label{supp:sec:3d_preprocessing}
In the 3D OV instance segmentation experiments on ScanNetv2~\cite{scannet200}, we integrated DeCLIP into the OpenMask3D~\cite{openmask3d} and Open3DIS~\cite{open3dis} frameworks. These baselines inherently leverage depth information for scene understanding. Specifically, these frameworks process the raw ScanNet data using the SensReader toolkit\footnote{https://github.com/ScanNet/ScanNet/blob/master/SensReader/README.md} to extract RGB images, depth maps, and camera poses. Subsequently, we utilize these components to reconstruct point clouds via back-projection. The baseline methods use this geometric information to align 2D features with 3D structures. Consequently, the evaluation operates within a multi-modal pipeline incorporating both RGB and depth modalities. This process ensures that depth information serves as a fundamental prerequisite for scene reconstruction rather than an auxiliary input.

\subsection{Failure Case Analysis}

\begin{figure*}[t]
    \centering
    \includegraphics[width=0.99\linewidth]{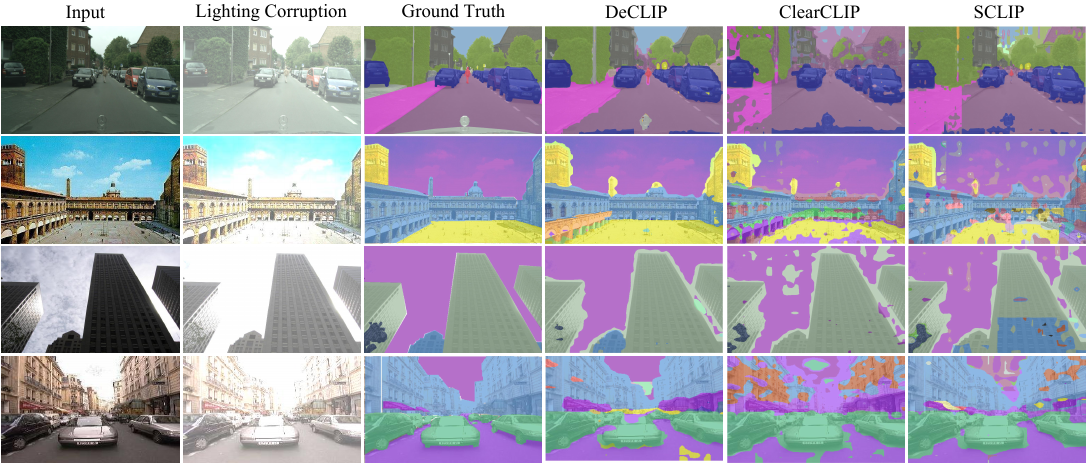}
      \caption{\textbf{Qualitative comparison under extreme lighting.} We present qualitative training-free OV semantic segmentation results on Cityscapes and ADE20K under high-brightness degradation. DeCLIP demonstrates superior semantic coherence compared to existing methods such as ClearCLIP and SCLIP. However, the third and fourth rows highlight a limitation where extreme overexposure leads to significant information loss, resulting in segmentation artifacts.}
    \label{fig:light_fail_case}
\end{figure*}

\mypara{Failure Cases in Occlusion and Novel Objects.} To understand the limitations of DeCLIP, we conducted a qualitative analysis on the OV-LVIS~\cite{lvis} and OVIS~\cite{OVIS} benchmarks. These datasets specifically target challenges such as small objects in long-tailed categories and heavy occlusions in videos. This analysis demonstrates scenarios where baseline methods (CLIPSelf~\cite{wu2023clipself}, CLIP-VIS~\cite{clip_vis}) are improved by employing DeCLIP as the backbone, while also identifying persistent challenges. As shown in Fig.~\ref{fig:fail_case}(a) (rows 1-2), DeCLIP successfully detects and aligns the small beachball with the correct textual prompt, whereas the baseline misclassifies it as a balloon. Additionally, in the video segmentation task (rows 3-4), DeCLIP achieves precise boundary delineation for large objects under heavy occlusion. However, Fig.~\ref{fig:fail_case}(b-c) reveals persistent limitations. In OV detection, identifying small objects in rare categories (e.g., electric shavers, applesauce) remains challenging. Similarly, video segmentation involving dense occlusions of small objects presents difficulties. Although DeCLIP avoids classification errors, it fails to segment certain object instances, resulting in false negatives.

\mypara{Failure Case in Extreme Lighting Conditions.} To investigate the impact of degraded image inputs (e.g., extreme lighting conditions) on DeCLIP, we performed a qualitative analysis in this scenario. Specifically, we adopted the benchmark protocol proposed by \cite{robust}. The qualitative comparison results of training-free OV semantic segmentation under the “Brightness” degradation at severity level 4 are shown in Fig.~\ref{fig:light_fail_case}. These results demonstrate that DeCLIP significantly outperforms existing methods in terms of segmentation accuracy and consistency under extreme lighting conditions (rows 1-2). This observation aligns with the quantitative robustness results in Tab.~\ref{tab:robust_seg1}. DeCLIP surpasses current methods in both mean Performance under Corruption (mPC) and relative Performance under Corruption (rPC) metrics, indicating the superior robustness of its learned representations. However, challenges remain in extreme conditions. In the third and fourth rows of Fig.~\ref{fig:light_fail_case}, although DeCLIP outperforms competitors, it struggles to preserve semantic details in severely overexposed regions.